\documentclass{article}
\usepackage{arxiv}
\usepackage{times}
\usepackage{latexsym}
\usepackage{url}
\usepackage{color,soul}
\usepackage{graphicx}
\usepackage{caption}
\usepackage{subcaption}
\usepackage[hidelinks]{hyperref}
\usepackage{longtable}
\usepackage{booktabs}
\usepackage{multirow}
\usepackage{amsmath}
\usepackage{rotating}
\usepackage{enumitem}
\usepackage{authblk}
\usepackage[titletoc,title]{appendix}

\usepackage{fontawesome5}
\usepackage{floatpag}
\usepackage[multiple]{footmisc}

\usepackage{floatpag}
\makeatletter
\def\namedlabel#1#2{\begingroup
    #2%
    \def\@currentlabel{#2}%
    \phantomsection\label{#1}\endgroup
}
\makeatother

\usepackage{newfloat}
\DeclareFloatingEnvironment[name={Supplementary Figure},fileext=lsf,listname={List of Supplementary Figures}]{suppfigure}

\usepackage{microtype}
\usepackage[sorting=none]{biblatex}
\providecommand{\keywords}[1]
{
  \small	
  \textbf{\textit{Keywords---}} #1
}

\title{Assessing the communication gap between AI models and healthcare professionals:
explainability, utility and trust in AI-driven clinical decision-making
}

\author[ ,1,2]{Oskar Wysocki\thanks{Corresponding author: oskar.wysocki@manchester.ac.uk}}
\author[3]{Jessica Katharine Davies}
\author[1]{Markel Vigo}
\author[3]{Anne Caroline Armstrong}
\author[2]{\\D\'onal Landers}
\author[3]{Rebecca Lee}
\author[1,2,4]{Andr\'e Freitas}
\affil[1]{Department of Computer Science, The University of Manchester}
\affil[2]{digital Experimental Cancer Medicine Team, Cancer Biomarker Centre,\authorcr CRUK Manchester Institute, University of Manchester}
\affil[3]{Faculty of Biology Medicine and Health, The University of Manchester}
\affil[4]{Idiap Research Institute}

\begin{document}

\date{}
\maketitle

\begin{abstract}

This paper contributes with a pragmatic evaluation framework for explainable Machine Learning (ML) models for clinical decision support. The study revealed a more nuanced role for ML explanation models, when these are pragmatically embedded in the clinical context. Despite the general positive attitude of healthcare professionals (HCPs) towards explanations as a safety and trust mechanism, for a significant set of participants there were negative effects associated with confirmation bias, accentuating model over-reliance and increased effort to interact with the model. Also, contradicting one of its main intended functions, standard explanatory models showed limited ability to support a critical understanding of the limitations of the model. However, we found new significant positive effects which repositions the role of explanations within a clinical context: these include reduction of automation bias, addressing ambiguous clinical cases (cases where HCPs were not certain about their decision) and support of less experienced HCPs in the acquisition of new domain knowledge. 

\end{abstract}

\keywords{explainable model, explainable AI, ML in healthcare, user study, clinical decision support, automation bias, confirmation bias, explanation's impact, clinical performance, usability}

\section{Introduction}

Clinical predictive models based on machine learning (ML) bring the promise of integrating real world evidence into clinical decision-making, balancing individual clinical experience with data-driven evidence. The dialogue between evidence, which is systematically collected and analysed, and in clinical practice, allows for a continuous evolution of the understanding of disease and treatment response. This is particularly important in the context of new diseases (e.g, COVID-19), new treatments or in understanding personalised responses (e.g. multi-morbidity or diverse populations).

Effective communication between healthcare professionals (HCPs) and ML models depends on the ability of the former to have a faithful mental representation of the latter and a critical ability to assess the strengths and limitations of ML models.

\textit{Explainability} \cite{bauerExplAiNed2021,liptonMythosModelInterpretability2017,gilpinExplainingExplanationsOverview2018,samekExplainableArtificialIntelligence2019,murdochDefinitionsMethodsApplications2019,thayaparan2020survey} emerged as a research area aiming to address interpretability bottlenecks in ML models resulting from the fact that many of them can be effective in predicting an outcome, but not in explaining their underlying reasoning, which limits their practical application in critical areas.
In recent work by \cite{holzingerInformationFusionIntegrative2022} \textit{Verification and Explainability Methods} were identified as one of the three main frontier topics in the area of medical AI (together with \textit{Complex Networks and their Inference} and \textit{Graph Causal models and Counterfactuals}) in order to satisfy the need for trustworthiness at multiple levels of the medical workflow.
Whilst recent explainable ML methods \cite{lundbergUnifiedApproachInterpreting2017,ribeiroAnchorsHighPrecision, haraMakingTreeEnsembles2018, vanlooverenInterpretableCounterfactualExplanations2020, dhurandharExplanationsBasedMissing2018, hanawaEvaluationSimilaritybasedExplanations2021, apleyVisualizingEffectsPredictor2019, jiangTrustNotTrust2018a} are instrumenting ML models to be more transparent, the perceived utility and suitability of explanation models has not been systematically investigated in the clinical context.

We aimed to address this research gap by conducting a user study on the effectiveness of current modalities of ML explainability amongst healthcare professionals for data-driven decision support. The study was performed using an explainable ML clinical decision support tool designed to help manage the admissions of patients with cancer and COVID-19: CORONET (COVID-19 Risk in ONcology Evaluation Tool \cite{coronet_paper}). In contrast to existing explainability studies performed over simplified tasks \cite{alufaisanDoesExplainableArtificial2020,cartonFeatureBasedExplanationsDon2020,lakkarajuFaithfulCustomizableExplanations2019,schafferCanBetterYour2019}, our experiment closely mimicked a real-world clinical scenario \cite{ExplainingBlackboxClassifiers}, where the decision is made based on multiple factors supported by a recommendation.

As part of the contributions, we propose a novel methodology to assess the pragmatic utility of explanations, which accounts for a systematic understanding of the interaction between a clinician’s mental model and the explainable ML model.
Additionally, we introduce a framework for evaluating the impact of ML explanations on perceived utility and usability in clinical context \cite{holzingerHumanAIInterfaces2021}. We present the results of an experiment conducted involving 23 HCPs, who were presented with 10 scenarios and had to decide whether to admit/discharge a patient with cancer and COVID-19 supported by a model firstly without an explanation (colour bar + score) and with explanation (+contributing features). 

\section{Methods}

\subsection{Framework for pragmatic evaluation of the model's explanation}

\begin{figure}[htbp]
\centering
\includegraphics[width= 1\textwidth]{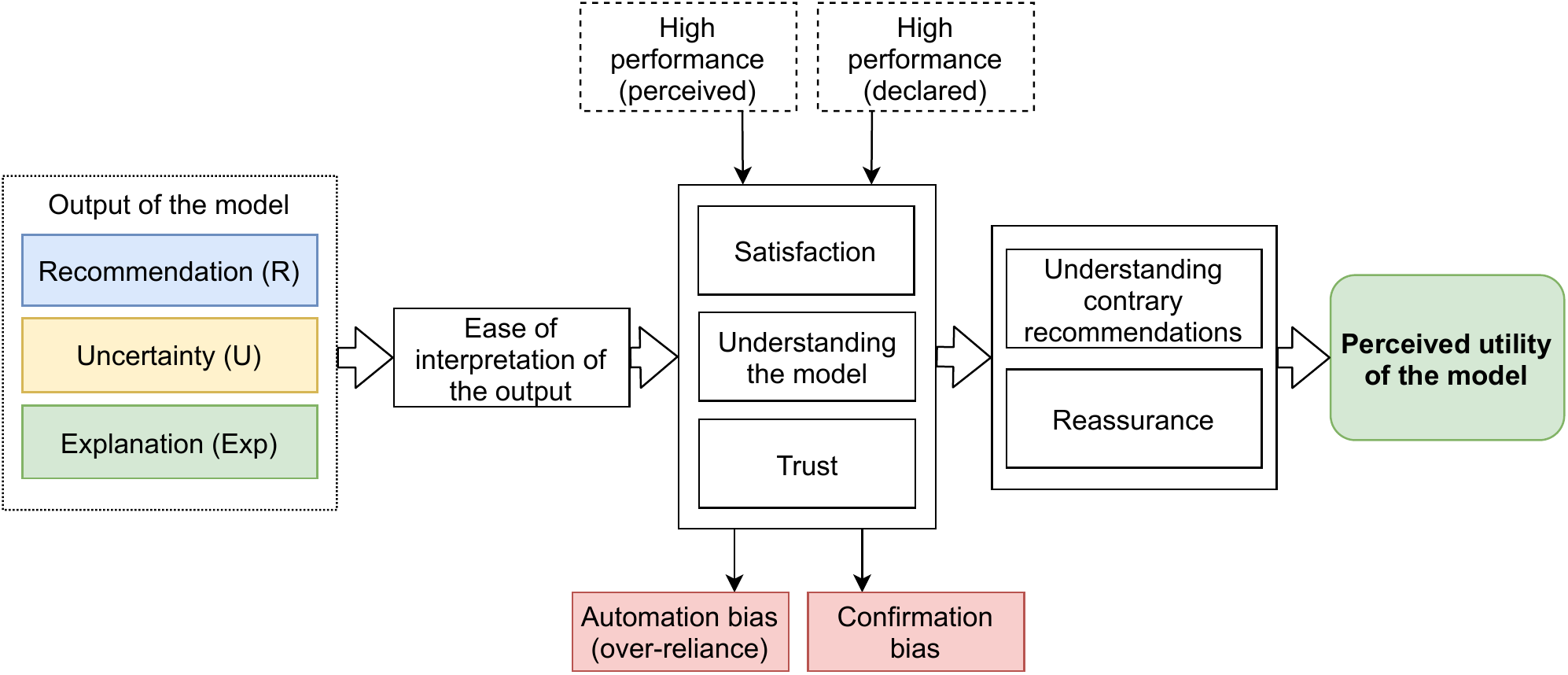}
\caption{Overview of the framework for pragmatic evaluation of model's explanation. It highlights key aspects of the user's mental model that affects the perceived utility of the tool a clinical setting. Pink boxes - negative effects.}
\label{fig:mental_model_diagram}
\end{figure}

\label{framework}
We introduce a framework for pragmatic evaluation of the model's explanation, outlining the key components from the model's output (including its explanatory components) to the model's usefulness in a clinical setting (see Fig.\ref{fig:mental_model_diagram}). It allows for investigating key aspects of the user's attitude that impact the perceived utility of the model in a clinical decision-making setting. Our end-to-end approach allows for the evaluation of the change in perception of the overall system's clinical suitability, i.e. evaluation of system-level effect, instead of item-level effect (single recommendation) \cite{ExplainingBlackboxClassifiers}.

In general \textit{explainability} is about transparency and traceability, highlighting decision-relevant parts of the representations within the algorithm, both for the dataset (population) or particular observation (individual patient). \textit{Explanation} is often used interchangeably with \textit{interpretation}, however the distinction must be made as the first is product by the second. \textit{Explanation} is a collection of features from the interpretable domain that contributes to the production of an abstract statement. \textit{Interpretation} is a process of mapping this statement into the domain the human expert can perceive, comprehend, and understand (for more detailed definition refer to \cite{holzingerHumanAIInterfaces2021}).
In our framework, first, the ease of interpretation of the tool's output is evaluated, which is a prerequisite for understanding, satisfaction and trust \cite{hudonExplainableArtificialIntelligence2021,bucincaTrustThinkCognitive2021}.
Note, that in the clinical decision-making process, the cognitive load due to the model output is one of many disjointed contextual factors and thus should not require the full working memory of the user \cite{musenClinicalDecisionSupportSystems2021}.

Then questions regarding trust, satisfaction and model understanding are systematically assessed. Trust in the model leads to its high persuasive power and \textit{reassurance} in cases where the user is less certain about the right decision \cite{mcgrathWhenDoesUncertainty2020}. Understanding the model's reasoning process supports the critical assessment of the model (i.e. understanding when the tool provide incorrect recommendations). Satisfaction measures how well the provided output corresponds the user's expectations.

Trust may be enhanced by providing the explanation \cite{holzingerInformationFusionIntegrative2022, ramaniExaminingPatternsUncertainty2020,doshi-velezConsiderationsEvaluationGeneralization2018,zhouEffectsUncertaintyCognitive2017}, uncertainty of the model's recommendation \cite{zhouEffectsUncertaintyCognitive2017,mcgrathWhenDoesUncertainty2020} and/or by ensuring the high performance of the model at a given task \cite{asanArtificialIntelligenceHuman2020}. The last aspect can be characterised by the declared performance by the model's developer in testing and validation, and the subjective performance perceived by the user during the experiment \cite{ExaminingEffectsPower,ExplainingBlackboxClassifiers}. 
However, excessive trust may have negative effects, i.e. \textit{confirmation bias} \cite{ghassemiFalseHopeCurrent2021} and \textit{automation bias} \cite{SKITKA1999991,goddardAutomationBiasHidden2011}. These are evaluated via the concordance between user's action, recommendation and correct action, and via time spent on decisions.

Finally, all aspects account for the perceived pragmatic utility of the model, evaluated via asking whether HCPs would use the model in their clinical practice.
The pragmatic utility aggregates model's \textit{usability}, defined as how effective, efficient and satisfying the model is in a specified context \cite{holzingerHumanAIInterfaces2021}, and \textit{clinical performance}, defined as the ability to deliver results that are correlated to a specific clinical condition and target population in experiments performed by medical experts \cite{mullerExplainabilityCausabilityArtificial2022}.

The output of the model may consist of three main components: i) recommendation (R, i.e. prediction, prognosis or classification); ii) uncertainty of the model's recommendation (U); iii) explanation (Exp); and can be presented to the user as: R, R+U, R+Exp and R+U+Exp. In our empirical analysis, as the CORONET model does not communicate its uncertainty, we put aside this aspect and performed a `R then R+Exp' within-subject scenario with one group of HCPs.
Although one of the visual components of the explanation (i.e. scatter plot, described in \ref{scatter_plot_explained}) makes the user aware of the model's imperfect performance (predicted scores for the whole training cohort, including incorrect recommendations), we did not evaluate this aspect directly. We did not provide explicitly any performance metric to the HCPs. Similarly, the perceived performance of the model was not measured. Thus, we did not evaluate the contribution of the model's performance to the HCPs' attitude towards the model. 
Individual characteristics of the user, such as user's background, knowledge and expectations that may impact their attitude \cite{devarajBarriersFacilitatorsClinical2014} were evaluated. 

Guided by the framework, we investigate the overall pragmatic impact of the explanations on the clinical decision by targeting the following research questions: 

\begin{enumerate}
\item[\namedlabel{itm:RQ1}{RQ1}] Does the explanation improve understanding of the model’s reasoning mechanisms?
\item[\namedlabel{itm:RQ2}{RQ2}] Does the explanation lead to an increase in model satisfaction?
\item[\namedlabel{itm:RQ3}{RQ3}] Does the explanation lead to an increase in trust?
\item[\namedlabel{itm:RQ4}{RQ4}] Do explanations improve the HCPs' critical understanding of the model?
\item[\namedlabel{itm:RQ5}{RQ5}] Do explanations enable improved clinical understanding?
\item[\namedlabel{itm:RQ6}{RQ6}] How do explanations impact the automation bias?
\item[\namedlabel{itm:RQ7}{RQ7}] Does the explanation increase the confirmation bias?

\end{enumerate}

%After determining the components contributing to the main objective of DSS, which is \textit{being useful in a clinical setting}, we hypothesized which components are affected by introducing an explanation to the model's output. To inspect the overall impact of the explanation on the user’s mental model, we formulated the following research questions regarding these specific components:

%how dimensions such as the initial attitude towards the model, expert knowledge, ease in interpreting visual explanations, user satisfaction, trust, reassurance and increasing confidence are correlated to the acceptability of the explanation model. 

This study indicates that there is a major unaddressed communication gap between explainable ML models and healthcare professionals and elicits priority areas for investigation.

\subsection{Clinical setting \& supporting model}

\begin{figure}[h!]
\centering
\includegraphics[width= .99\textwidth]{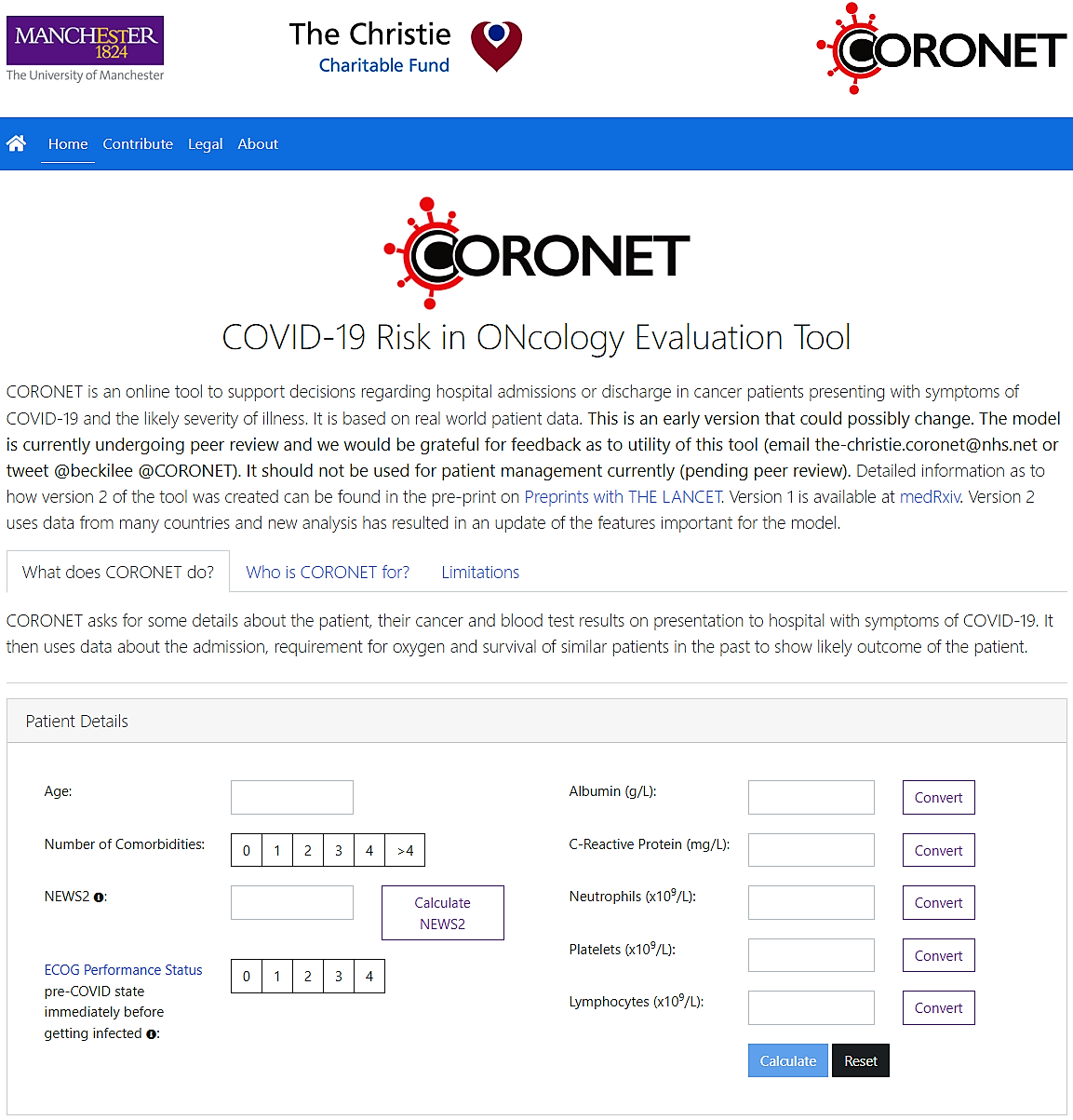}
\caption{User's interface of the CORONET available at \url{https://coronet.manchester.ac.uk}. The user inputs the values manually and is warned in case of values out of expected range (Supp Fig. \ref{fig:out_ouf_range_warning}).  `Calculate NEWS2' button leads to a pop-up window with a calculator (Supp Fig. \ref{fig:NEWS_pop_up}). `Convert' button refers to \url{http://unitslab.com/}. After pressing 'Calculate' the output is generated and presented to the user in the same window, below the 'Patient Details' field. Of note, during the experiment the participants did not directly interact with the interface. Static images with the tool's output were provided.}
\label{fig:user_interface}
\end{figure}

\noindent \textbf{CORONET model}: % (COVID-19 Risk in ONcology Evaluation Tool \cite{coronet_paper})
In this study we used CORONET (COVID-19 Risk in ONcology Evaluation Tool \cite{coronet_paper}) which was developed 
to help determine the need to admit patients to hospital on the bases of their likelihood of needing oxygen (as generally oxygen is only
given in hospital) and their severity of COVID-19, as indicated by predictions for required oxygen and/or death \cite{LEE2021100005,burkeBiomarkerIdentificationUsing2022,freemanWaveComparisonsClinical2022}. As a result, four key outcomes were established, arranged in a 0–3 point ordinal scale: discharged, admitted ($\geq24$ h inpatient), admitted+O2 requirement (including ventilator support), and admitted+O2+died (with the death directly attributable to COVID-19 disease, not to cancer). These four outcomes were used as measures of disease severity. Contrary to an analysis of binary outcomes (e.g., need for oxygen vs. no need for oxygen), this strategy improved the ability to provide a complete clinical picture that was important for overall decision-making regarding hospital admission. During development, the model was first validated on an external cohort \cite{coronet_paper}, and later on the most recent data from Omicron variant \cite{wysockiInternationalComparisonPresentation2022a}. 
The CORONET tool is available online\footnote{\url{https://coronet.manchester.ac.uk}} providing an interactive user interface with cross-device compatibility. The user interface is depicted in Fig. \ref{fig:user_interface} (for more details please see Fig. \ref{fig:out_ouf_range_warning}, \ref{fig:NEWS_pop_up} and Suppl Methods \ref{user_interface_development}).

\begin{figure}[htbp]
\centering
\begin{subfigure}{.75\textwidth}
  \centering
  \includegraphics[width=\linewidth]{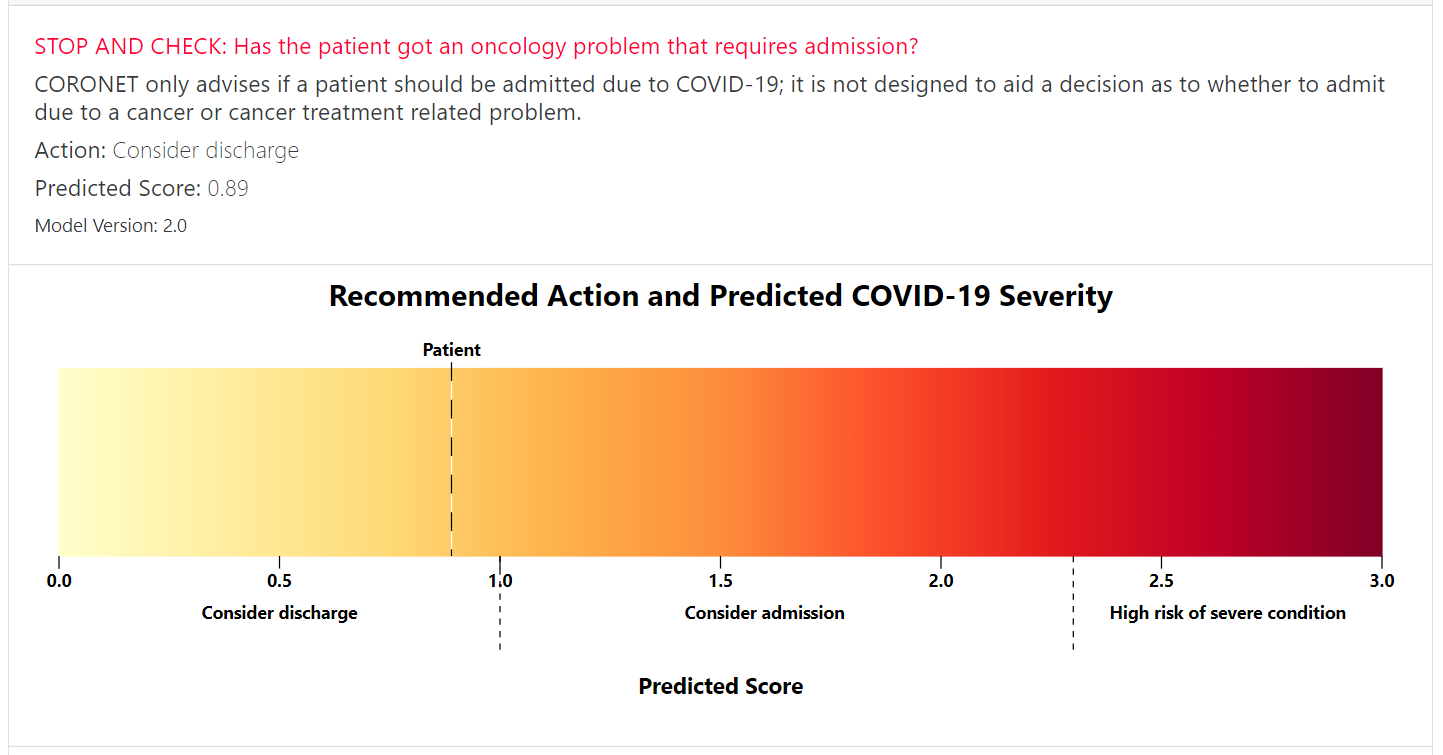}
  \caption{Recommended action and the CORONET score, which defines action. The score is presented on a colour bar, scaled 0-3 with decision thresholds}
  \label{fig:output_colorbar}
\end{subfigure}%
\\
\begin{subfigure}{.75\textwidth}
  \centering
  \includegraphics[width=\linewidth]{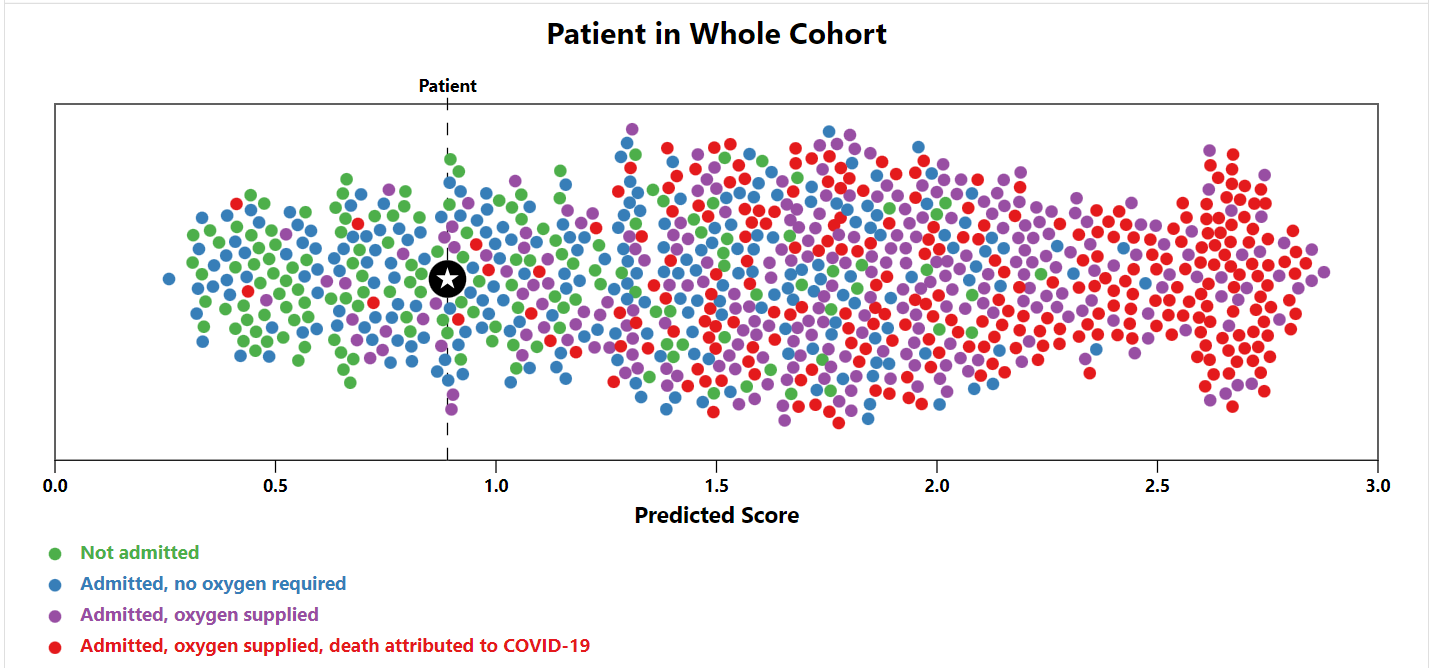}
  \caption{The plot shows all patients used for training of the CORONET model. Each dot represents an individual patient. The colour corresponds to their true outcome. The location on the X-axis is determined by the CORONET score based on the individual’s data.}
  \label{fig:output_scatterplot}
\end{subfigure}
\\
\begin{subfigure}{.75\textwidth}
  \centering
  \includegraphics[width=\linewidth]{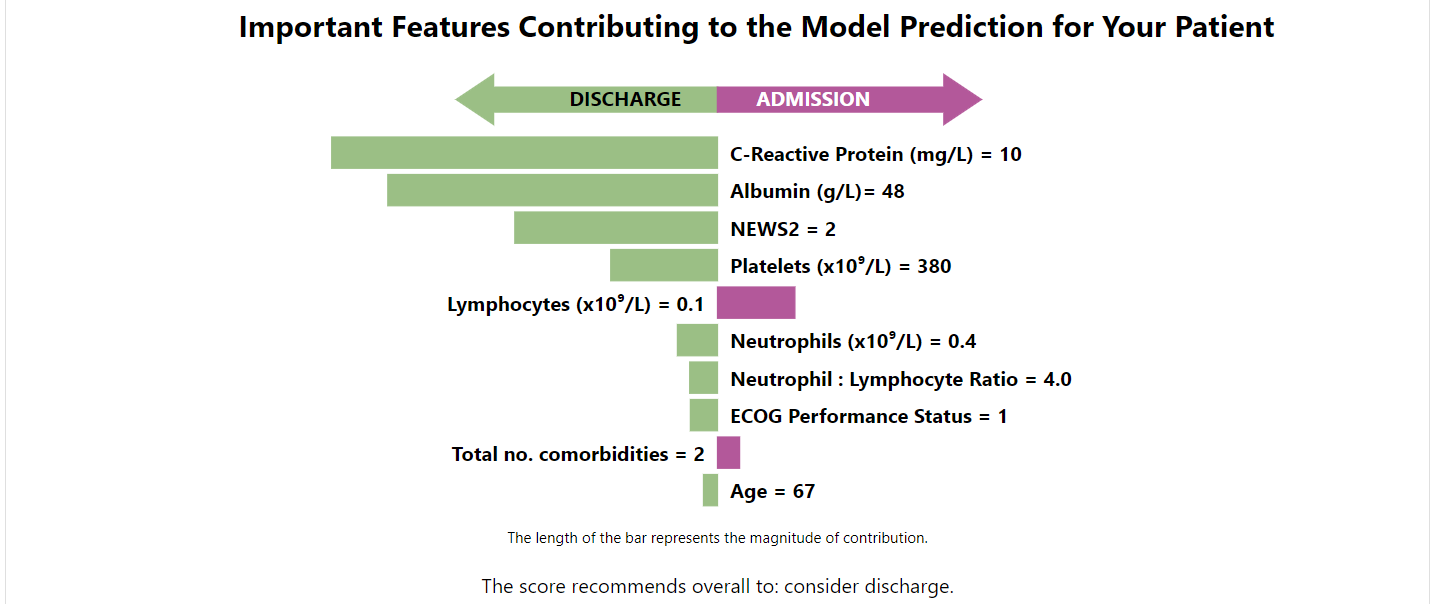}
  \caption{The contribution plot with features contributing to the individual prediction obtained using SHAP explanation.}
  \label{fig:output_barplot}
\end{subfigure}%

\caption{Output of the CORONET tool provided for each individual patient.}
\label{fig:output_plots}
\end{figure}

\noindent \textbf{Recommendation (R):} The basic output of the model is the recommended action and the CORONET score, which defines action. The score is presented on a colour bar, scaled 0-3 with decision thresholds (Fig.\ref{fig:output_colorbar}). The score below 1.0 recommends discharge, above 1.0 admission, and above 2.3 suggests high risk of severe COVID-19 illness. Critically, this is in a group of patients who may also present to hospital with cancer or treatment related problems and not just COVID-19. This therefore adds complexity to the decision-making for the HCP. However, CORONET is only built to aid decisions regarding COVID-19 severity and requirement for admission. The CORONET model is a regression random forest, which was trained using the ordinal 0-1-2-3 scale as a dependent variable. True outcomes used as \textit{y} are: 0-patient discharged; 1-patient admitted; 2-admitted and required supplemental oxygen; 3-admitted, required O2 and died. Thus, the model predicts the score in range 0-3 and the recommended action is delivered based on a pragmatically defined threshold \cite{coronet_paper,Lee2020.11.30.20239095,coronetabstract}. %CORONET's user interface is depicted in Fig.\ref{fig:user_interface}.

\noindent \textbf{Model explanation (Exp):} The explanation of the output consists of two visual components. Firstly, it shows a \textit{scatter plot} \label{scatter_plot_explained}(Fig.\ref{fig:output_scatterplot}) with the predicted score on the x axis for all patients used in model derivation (scores predicted in Leave-One-Out Cross Validation \cite{coronet_paper}). Each dot represents an individual patient and the colour corresponds to their true outcome. All patients are sorted from left to right according to their predicted score. The plot allows the user to locate the patient in question (marked by a star) in the whole cohort, considering both true outcomes and model’s recommendations. The point distribution elicits the model’s errors, i.e. some recommendations may be incorrect. For several ‘death outcomes’ the model predicted scores below admission threshold. Similarly, for some discharged patients the tool recommended admission. This reveals the performance of the model to the user.
%\textcolor{red}{The plot is intended to represent the \textit{Type A component} from European In Vitro Diagnostic Medical Devices Regulation (IVDR, Regulation (EU) 2017/746), which means it reveals the performance of the model to the user.}

Second, the explanation shows a bar plot with features contributing to the individual prediction (later referred to as \textit{contribution plot}, Fig.\ref{fig:output_barplot}). The length of the bar represents the magnitude obtained using SHAP explanation \cite{lundbergUnifiedApproachInterpreting2017}. The colour shows the direction of the contribution (towards discharge or admission). Features are ordered from top to bottom depending on the contribution. 
The figure serves as a local explanation of the model, where the bars change for each individual recommendation. The contribution plot is produced only for the patient in question, i.e. to see feature contribution for other cases the user needs to input new values. The contribution plot's layout was iteratively derived from discussions between user user experience designers and oncologists in order to produce the most intuitive visualisation. Among other considered layouts were: one side bar plot, waterfall plot \cite{lundbergUnifiedApproachInterpreting2017}, force plot \cite{lundbergUnifiedApproachInterpreting2017}, Partial Dependence plots \cite{friedman2001greedy} and Individual Conditional Expectation plots \cite{goldsteinPeekingBlackBox2015}.
%\textcolor{red}{The \textit{contribution plot} reveals relevant factors in the decision making of the model and connects them with the weighting, complexity and hierarchy of algorithmic features inherent in the algorithm, and according to IVDR represents \textit{Type B component}}

%%%%%%%%%%%%%%%%%%%%%%%%%%%%%%%%%%%%%%%%%%%%%%%%%%%%%%%%%%%%
% CORONET INTERFACE FIgURE
%%%%%%%%%%%%%%%%%%%%%%%%%%%%%%%%%%%%%%%%%%%%%%%%%%%%%%%%%%%%

Overall, the model’s output consists of three cognitive chunks \cite{doshi-velezConsiderationsEvaluationGeneralization2018} organized in a three-step hierarchy : i) colour bar with the score; ii) scatter plot; iii) contribution plot. The derivation of the model and its associated source code are available at \url{https://github.com/digital-ECMT/CORONET_tool}.

\subsection{Study design}

\begin{figure}[htbp]
\centering
\includegraphics[width= .5\textwidth]{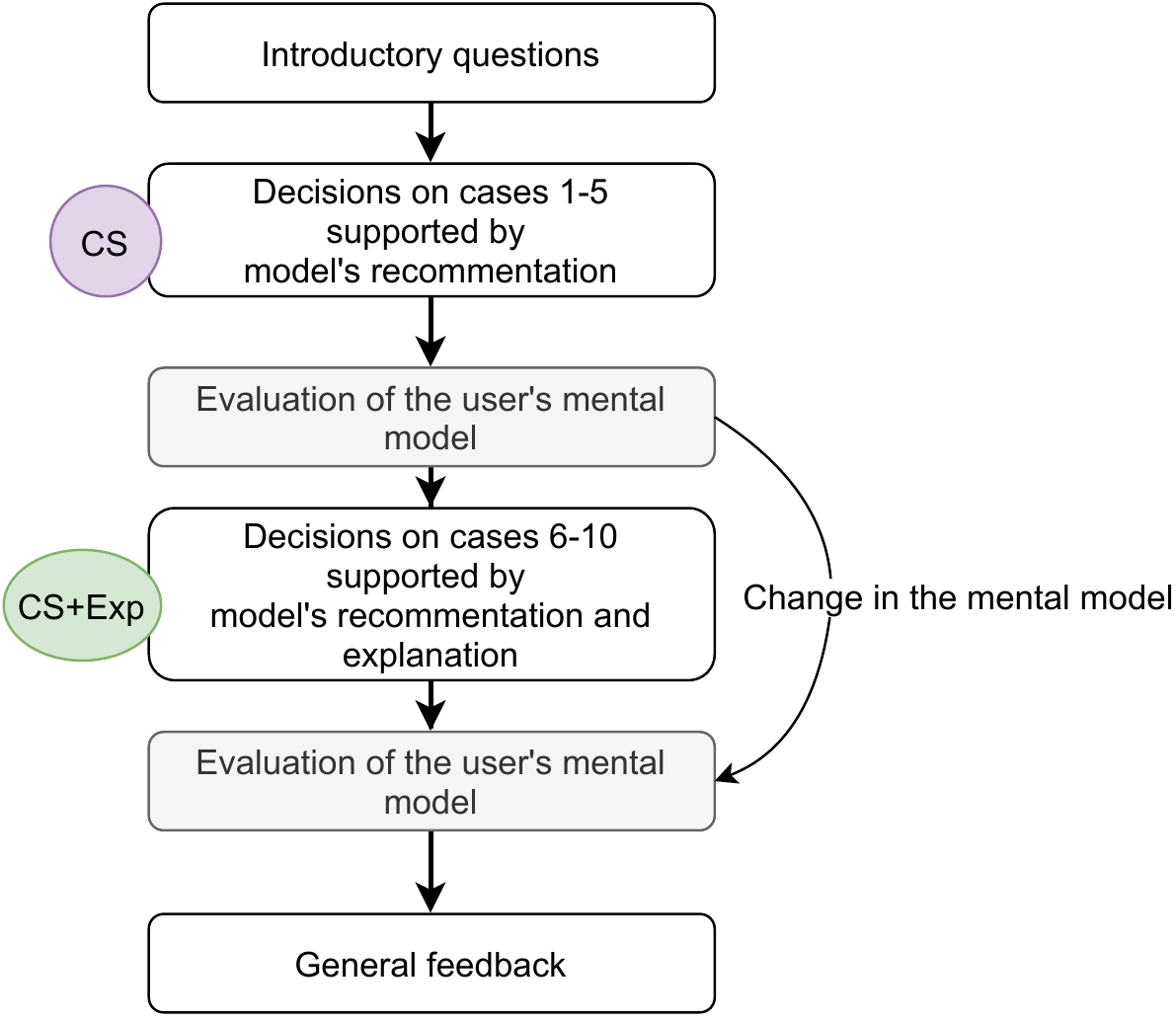}
\caption{Experimental design. The within-subject design enabled examination of the change in attitude
towards the model of each participant, as each of them was first supported by CORONET Score (CS) alone and then by CORONET Score and explanation (CS+Exp).
}
\label{fig:experiment_design}
\end{figure}

The study follows a within-subject design whereby each participant uses the tool in two conditions. Firstly, the participant is provided information in the form of a recommended action and a score on a scale of 0-3 (later referred to as \textbf{CS} - CORONET score), and told that the higher the patient’s score, the more severe the predicted outcome is. Secondly, the participant is then provided with the recommendation, the CORONET score and an explanation of how the tool has arrived at this score for the patient (later referred to as \textbf{CS+Exp} - CORONET score plus explanation). This approach minimises any random noise introduced by the variability on user's clinical and technological exposure and previous experience. The order of presented cases was the same for all participants. The study design is depicted in the Fig.\ref{fig:experiment_design}.

The online questionnaire consisted of six stages: i) introductory questions refering to the HCP's background and expectations  ii) five artificial patient cases with only the recommendation provided; iii) questions evaluating the model's usefulness; iv) another five artificial cases this time provided with the recommendation and explanation; v) questions evaluating the model's usefulness (same as in iii) vi) overall impressions of the tool. The questions were designed specifically for this study.

Ten artificial patient case scenarios were constructed, reviewed and approved by a senior oncology fellow and a consultant oncologist. The decision to admit or discharge was based on clinical guidelines and best practice. The scenarios were similarly structured and comprised an introduction to the patient (including demographics, presenting complaint, and relevant past medical history - all 10 cases are detailed in the supplementary material), the patient’s parameters from observations (vital signs), and blood test results. 

Two of the ten patient cases (Daniel in CS, and Christine in CS+Exp scenarios) were intentionally built in a way that the model would wrongly discharge the patient, as CORONET considers features of COVID-19 severity only \cite{LEE2021100005} and not oncological emergencies which might be concomitant e.g. neutropenic sepsis. In these cases, the healthcare professional should be able to apply their clinical judgement and recognise the patient would require admission for oncological reasons and subsequently override the model’s decision. In Daniel’s case, the admission is due to renal dysfunction and tumor lysis syndrome following chemotherapy received three days prior for his diagnosis of diffuse large B cell lymphoma. Therefore, this is a scenario where CORONET is thought to potentially ‘wrongly discharge’ the patient as it does not take into consideration oncological emergencies as part of it's recommendation.. In Christine’s case the admission is due to her recent chemotherapy course. Again, because CORONET does not take other types of infections and treatment related presentations, this is a situation where CORONET is not adequate in making an accurate decision.

These cases were included to assess for over-reliance on the model (automation bias). We analysed the concordance between clinicians’ decisions, and the model’s recommendations and the correct actions approved by our team of experts. 

We tracked the time spent on each part of the questionnaire ensuring that it was not quickly skipped when the user lost interest or was engaged with external tasks. It also allowed for tracking the time spent on decision of each patient case.

\subsection{Selection of participants}

This research project was ethically approved by the research ethics committee [REC reference: 20/WA/0269]. Participants were required to be clinically active healthcare professionals (determined as having patient contact in at least 10\% of their working hours) and currently working with patients with cancer presenting with COVID-19. This included senior doctors, junior doctors, physician associates, pharmacists, advanced practitioner nurses, and staff nurses. Any participants who failed to meet these inclusion criteria were excluded. Participants were recruited via email invitations to NHS Trusts, clinician networks, and social media groups. Although CORONET was already available online by the time of the experiment, to the best of our knowledge the participants were not using the tool prior to the experiment. The familiarity with the tool and/or with the preprint \cite{Lee2020.11.30.20239095} was not required. Knowledge related to ML, computational methods or statistics was not required.

%\newpage

\section{Results}

\subsection{Participants}
23 healthcare professionals participated in the experiment, with various level of experience and expertise (Table \ref{tab:participants_demo}). The median knowledge regarding managing patients with COVID-19 was five (in one to seven scale), evaluated via direct question. Most of the HCPs felt very comfortable using new technologies. 35\% of participants perform statistical analysis at work, and majority of participants do use online calculators or scoring systems, similar to CORONET (83\% and 65\%, more in Supp Table \ref{tab:tasks_performed}).

\begin{table}[htb!]
\centering
\caption{Characteristics of 23 healthcare professionals participated in the experiment.}
\label{tab:participants_demo}
\begin{tabular}{lc} 
\toprule
Age group                                                                                                                & \textbf{n}                 \\
21-30                                                                                                                    & 6                          \\
31-40                                                                                                                    & 10                         \\
41-50                                                                                                                    & 3                          \\
51-60                                                                                                                    & 4                          \\
                                                                                                                         &                            \\
                                                                                                                         & \textbf{n}                 \\
Advanced Nurse Practitioner or higher                                                                                    & 2                          \\
Consultant / attending physician                                                                                         & 5                          \\
Doctor (within 2-4 years of graduating /
  specialty trainee (ST) / fellow                                               & 2                          \\
Doctor within first year of graduating
  (FY1) / intern                                                                  & 1                          \\
General practitioner / GPST / community
  doctor                                                                         & 3                          \\
Pharmacist                                                                                                & 1                          \\
Registered Nurse                                                                                                         & 1                          \\
Specialist Nurse                                                                                                         & 1                          \\
Specialist registrar / ST 3+ / senior
  resident/senior fellow                                                           & 7                          \\
                                                                                                                         &                            \\
                                                                                                                         & \textbf{Median [min,max]}  \\
How comfortable you feel when using new
  technology?                                                                    & 6 [3,7]                    \\
\begin{tabular}[c]{@{}l@{}}Knowledge on the management of patients with cancer\\who have developed COVID-19\end{tabular} & 5 [2,7]                    \\
\bottomrule
\end{tabular}
\end{table}

\subsection{HCPs want to know both contributing features and uncertainty. Uncertainty was considered to be more important than contributing features}
\begin{figure}[htb!]
\centering
\includegraphics[width= .9\textwidth]{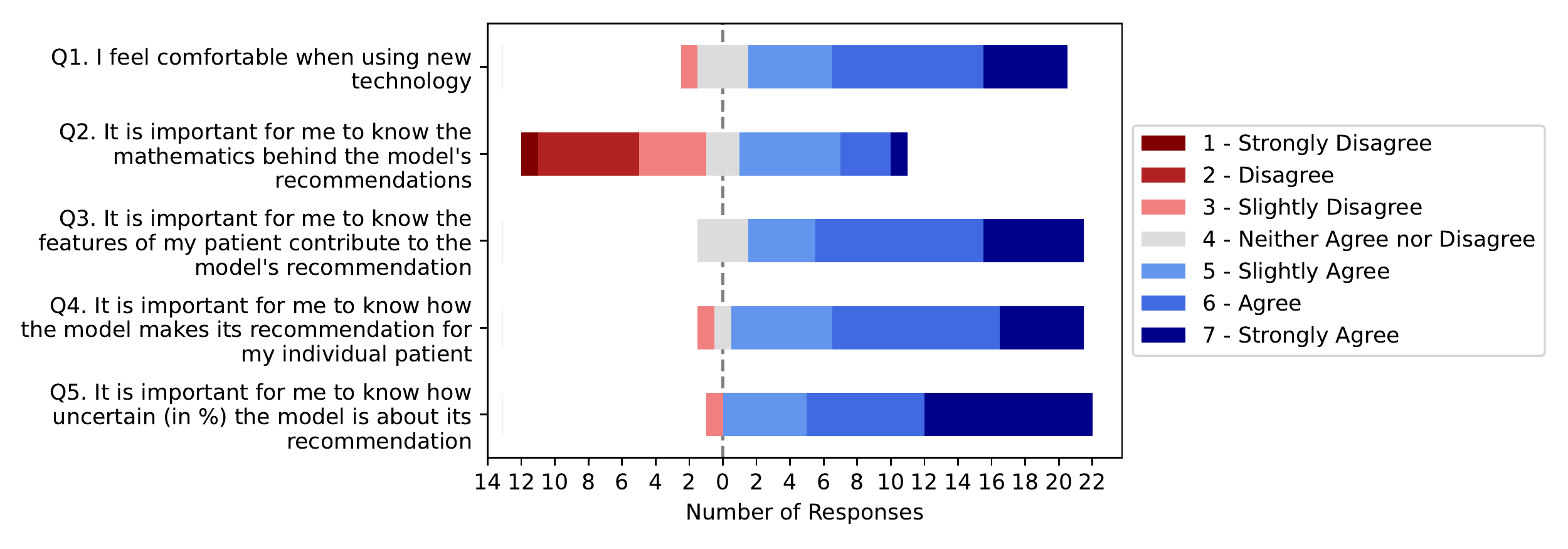}
\caption{Responses to questions related to HCP’s expectations regarding a ML model. Q3-Q5 relate directly to the expectation regarding model transparency and explanation. Q3 investigates the need for global explanation, and Q4 asks about local explanation for the “individual patient”.}
\label{fig:Q1Q5_likert}
\end{figure}

The responses regarding the expectations for the ML-based DSS are depicted in Fig.\ref{fig:Q1Q5_likert}. 87\% (20/23) of HCPs were interested in knowing the features contributing to the model's recommendation, three HCPs had neutral opinions. 91\% (21/23) considered the model’s explanation of an individual recommendation as important. Apart from knowing \textit{why} the model recommends such action, the uncertainty behind the recommendation was even more essential for the HCPs: 96\% (22/23) of them at least slightly agreed, and 43\% (10/23) strongly agreed, which is the highest `strongly agree' proportion in the study.

Knowing the mathematical framework behind the model is significantly less important than the local/global explanation and the model's uncertainty ($p<0.001$, Kruskal-Wallis test, Supp Table \ref{tab:expectation_stats}). Intriguingly, 73\% of HCPs who were not interested in the mathematics behind the model would like to see an associated model explanation and uncertainty. This points into largely unaddressed research questions on the dialogue between explanations and safety properties (such as a representation for uncertainty) as proxies for risk assessment. 

\begin{figure}[htb!]
\centering
\includegraphics[width= .9\textwidth]{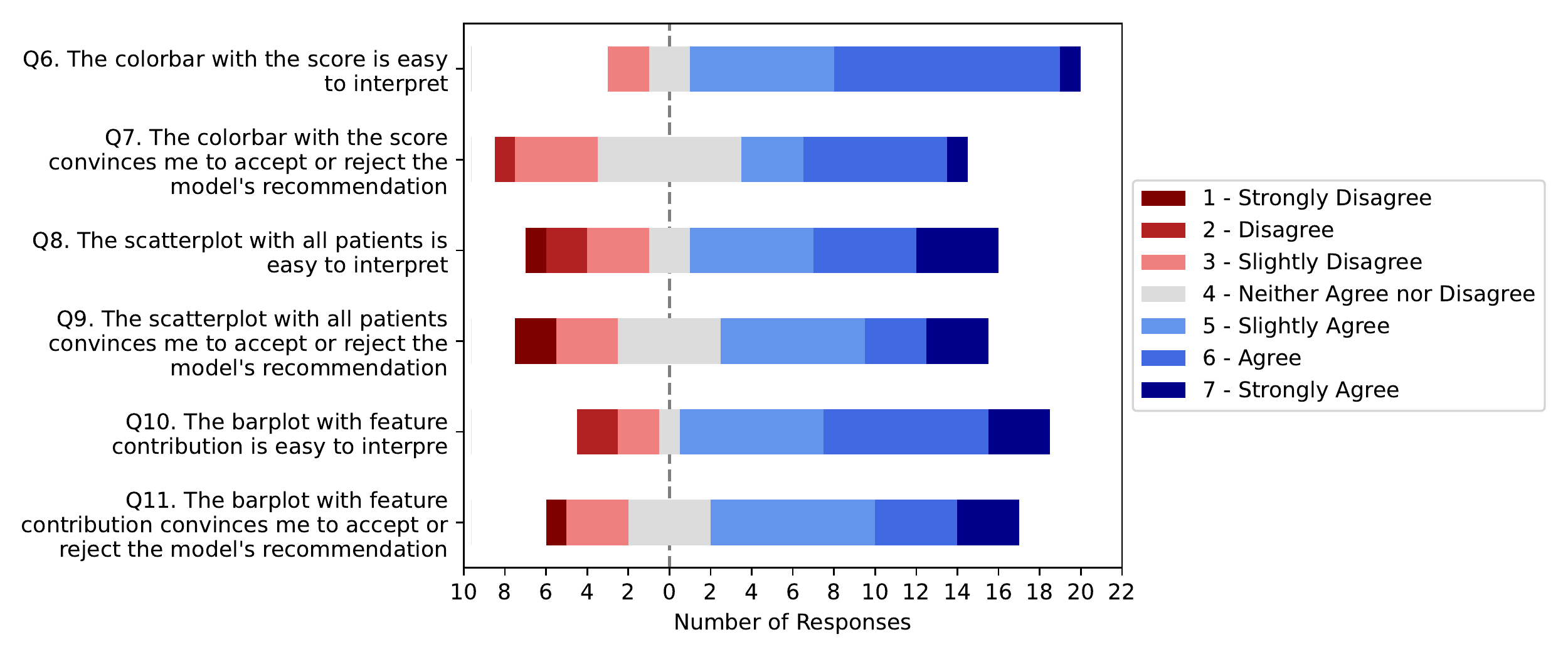}
\caption{Responses related to HCP’s easiness of interpretation of visual output of CORONET and whether it convinces the user to accept the model’s recommendation.}
\label{fig:Q6Q11_likert}
\end{figure}

\subsection{Visual explanations were easy to interpret}

Explanation visualisations were easy to interpret and contributed to convincing HCPs to accept or reject the model's recommendation (Fig. \ref{fig:Q6Q11_likert}, Supp Table \ref{tab:visual_stats}). This aspect was measured via direct questions for each visualisation: i) whether it was easy to interpret and ii) it was convincing to accept of reject the recommendation.
For the majority of HCPs, the colour bar was easy to interpret (83\%; 19/23). Interestingly, the CS (score without further explanation) presented on 0-3 colour scale convinces 48\% (11/23) HCPs to accept/reject the recommendation. Among these 11 convinced by CS, the average response to questions regarding expectations are high ($>$5.7).

We found no significant difference between the ease of interpretation of the colour bar, scatterplot or contribution plot, as well as no difference in their persuasive power ($p>0.05$, Supp Table \ref{tab:visual_stats_pairwise}). We did not find any significant correlation between questions related to the visual output and either knowledge on the management of patients with cancer who have developed COVID-19, nor to expectations for ML-based DSS (Supp Table \ref{tab:visual_stats_corrs}). Thus, the perception of the visual output was not affected by expectations, nor the competence in the task.

%Despite the general high level of comfort with new technology, there is a polarity in responses to the question about interpreting the dot plot (figure with ‘your patient’ in the whole cohort, Q8).fourresponders strongly agree and three disagree or strongly disagree with easiness of interpretation. This may be caused by differences in patience and eagerness to understand the plot, which requires certain cognitive effort. This again highlights the tradeoff between a quick and easy interpretation of the figure and the amount of contained  information. 

\subsection{Explanations did not significantly impact on the decision-making}

\begin{figure}[htb!]
\centering
\includegraphics[width= .8\textwidth]{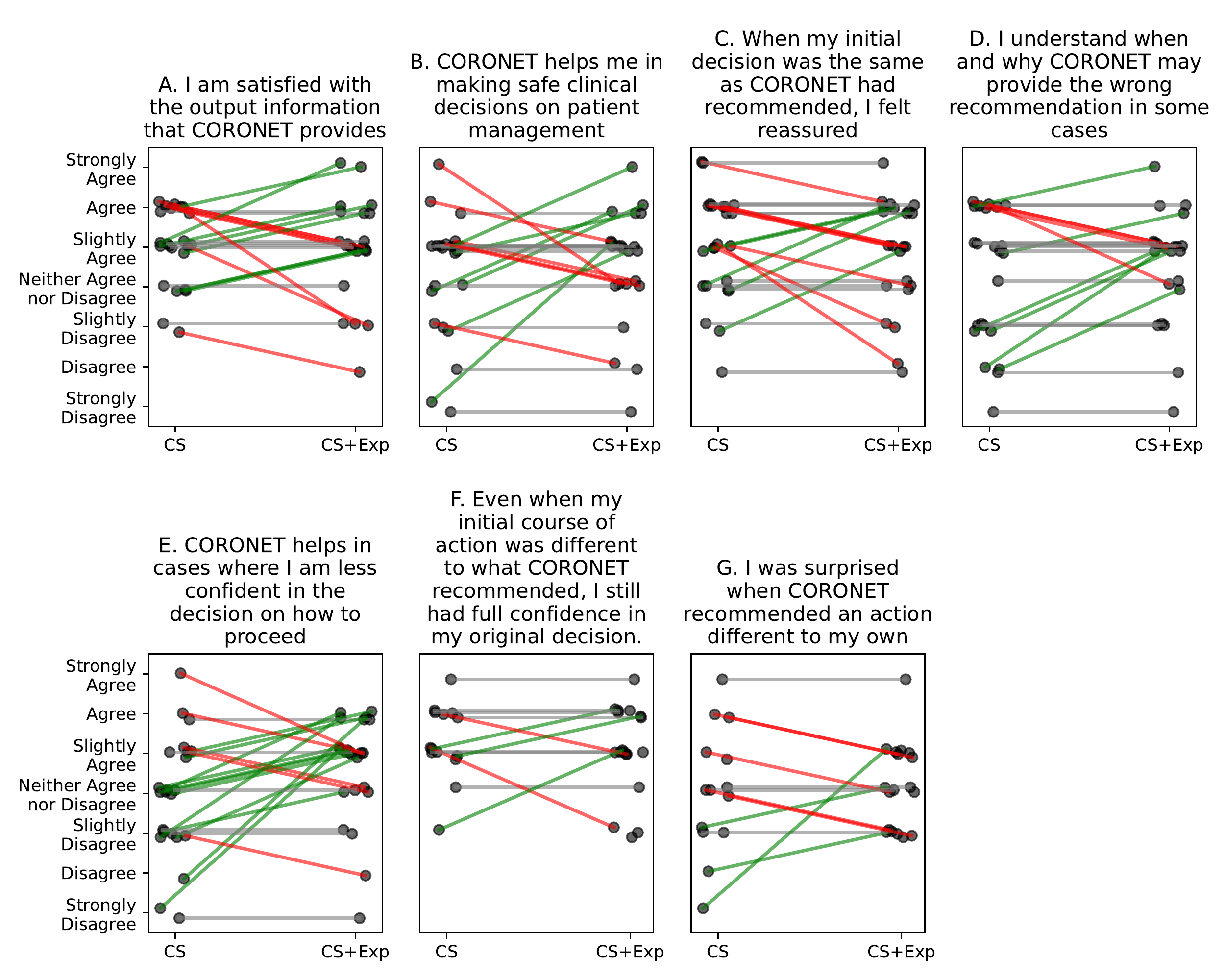}
\caption{Paired answers for the same questions among 23 Healthcare Professionals. Each dot is an individual answer. Lines: gray - no change; green - positive change up in the Likert scale; red - negative change. No statistically significant change observed in pairwise comparison.}
\label{fig:explanation_impact}
\end{figure}

In the pairwise comparison of responses between CS and CS+Exp scenarios, we did not find a statistically significant change in the HCPs' attitude towards the model (i.a. satisfaction, trust, understanding, reassurance). The results are summarized in Fig. \ref{fig:explanation_impact}, Fig.\ref{fig:likert_scale_change_in_feedback}, Supp Table \ref{tab:AG_medians}.

The explanation did not improve satisfaction (\ref{itm:RQ2}, Fig.\ref{fig:explanation_impact} question A), trust (\ref{itm:RQ3}, Fig.\ref{fig:explanation_impact} questions B,C,E) nor the understanding of how the model produced the recommendation (\ref{itm:RQ1}, Fig.\ref{fig:explanation_impact} questions D,F,G).
Of note, we observe a slight but not significant positive change ($p=0.056$) in `help in cases where I am less confident in the decision on how to proceed'. 

The majority of HCPs replied positively (57-74\% answers at least `Slightly agree' in quest. A-C) regarding the CS output alone. We established the following associations between positive responses and HCPs' expertise and expectations:

\begin{itemize}
    
    \item The lower the expertise, the more helpful the tool appeared to be, even when no explanation is provided ($r=-0.482$, $p=0.02$, Fig.\ref{fig:expertise_vs_feedback_CS}).

    \item The higher the need for knowing the contributing features, the less helpful the CS output is likely to be: in making safe decisions ($r=-0.653$; $p=0.001$, Fig.\ref{fig:knowing_features_vs_helpful_CS}) and in cases where HCPs are less confident ($r=-0.553$, $p=0.006$, Fig.\ref{fig:knowing_features_vs_helpful_CS})

\end{itemize}

Additionally, being convinced by the colour bar was correlated with satisfaction and reassurance in the CS only scenario ($r=0.484$, $p=0.019$ and $r=0.527$, $p=0.01$, Fig.\ref{fig:colorbar_vs_satisfaction}).

\subsection{Model explanations may have an adverse effect on HCPs}
CS-Exp output of the model led to both positive and negative changes in satisfaction, helpfulness of the model, reassurance and understanding. 
% satisfaction
34\% HCPs were more satisfied with CS+Exp than CS, but in 23\% their satisfaction decreased (\ref{itm:RQ2}). 
Higher satisfaction was weakly correlated with ease in the interpretation of the contribution plot ($r=0.436$, $p=0.038$), but not correlated with being convinced by it ($p=0.24$). 
HCPs who were convinced by the contribution plot in the CS+Exp scenario (65\%; 15/23), showed high levels of satisfaction both without (average 5.4) and an with explanation (average 5.6). This is also true regarding the scatter plot.

% help in making safe decisions
26\% of HCPs said that the tool with CS+Exp was more helpful in making safe clinical decisions, with 26\% stating the opposite.
There was a positive correlation $(r=0.63$, $p=0.001$, $p_{adj}=0.135$) between change in satisfaction and change in help in making safe decisions.
% reassurance
22\% of HCPs felt more reassured in the CS+Exp scenario, while 35\% felt the opposite.

% understanding
The provided explanations did not change the level of understanding of when and why the model produces wrong recommendations for 57\% (13/23) of HCPs (\ref{itm:RQ1}).
For 17\% the explanation even led to lower understanding. This cannot be attributed to the time spent on the patients' analyses, nor ease in interpreting the diagrams ($p>0.05$). 

The change in understanding wrong recommendations was associated with the change in the reassuring utility of the tool ($r=0.762$, $p<0.001$, $p_{adj}=0.003$), which is the strongest correlation found between changes (Supp Table \ref{tab:correlations_changes}).

% Help in cases with less confidence
The highest percentage - 52\% of positive changes was observed to help in cases where the HCP is less confident. Among them, there were three HCPs who rated their knowledge of managing patients with COVID-19 as 2 (\ref{itm:RQ5}). However, 22\% found CS+Exp less helpful.

%TO DISCUSSION
%We hypothesise that high \% of dissatisfied HCPs is related to information overload (too much cognitive effort) or simply not meeting the expectations regarding the explanation

%Same clinicians that felt more satisfied when the explanation was delivered, also found CORONET more helpful then

%a negative change in attitude among eight clinicians may suggest that explanation introduced even more uncertainty in clinical judgment, as it may have differed with human reasoning in particular cases.

%We found that an increase in understanding why the model recommends the wrong decision leads to an increase in reassurance when both user’s and the model’s decisions align. Additionally, the reassurance comes together with satisfaction. 

\subsection{Over-reliance on the model's recommendation leads to wrong decisions}

\begin{figure}[h!]
\centering
\includegraphics[width= .7\textwidth]{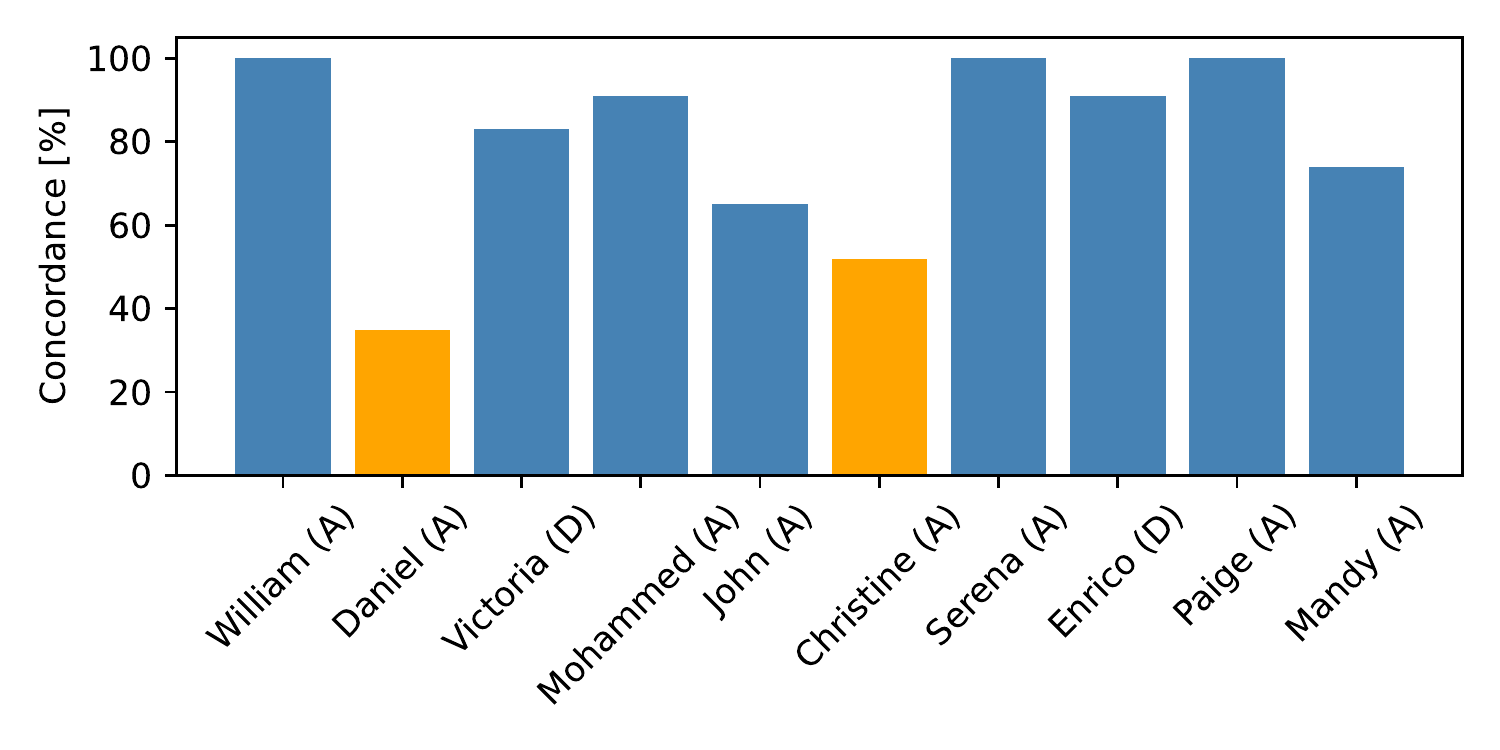}
\caption{Concordance between decisions made by respondents and correct action approved by our team. Higher is better. (A) - correct decision ‘admit’, (D) - ‘discharge’. Daniel and Christine (orange bars) were cases where CORONET recommends incorrect discharge.}
\label{fig:overreliance}
\end{figure}

There are only three cases where all the HCPs made the correct decision, and six cases where the concordance with the correct decisions was more than 80\% (Fig.\ref{fig:overreliance}). Of note, the concordance does not depend on whether a patient should be admitted or not when the tool recommends correctly (on average 88\% vs 87\%, p=0.73).

However, the lowest concordance with the correct decision was observed for cases where CORONET recommended the wrong action (\ref{itm:RQ6}, Fig.\ref{fig:overreliance}). In Daniel’s case 65\% (15/23) HCPs decided to discharge, as CORONET recommended, while for Christine this accounted for 48\% (11/23). As the level of complexity and difficulty in decisions across the cases 1-10 was similar, we argue that the lowest concordance for Daniel and Christine is caused by the wrong recommendation provided by the model. The results strongly suggest that for both types of output (CS and CS+Ex) HCPs over-relied in the recommendation provided by the model.

\subsection{Quicker decisions due to over-reliance on explanations}
Decisions which were in agreement with the model recommendation required less time (\ref{itm:RQ6}, Table \ref{tab:time_spent}, Fig.\ref{fig:time_spent_on_decisions}). For cases 1-5, where the tool provides only the CS, the median time spent on one case was 48s when the user agreed with the tool, and 61s otherwise. For cases 6-10, where the CS+Exp was provided, we observed a statistically significant difference: decisions required 38s for decisions aligned with and 84s (median values) for decisions opposed to the tool’s recommendation ($p<0.001$, $CLES=0.754$, MWU test). 

As described in the previous section, some respondents needed less time when deciding on cases where the explanation was delivered. Further investigation of the agreement with the model reveals that when the HCP agrees with the recommendation (\ref{itm:RQ7}), the combined CS+Exp output leads to significantly quicker decisions than CS output: 38s for CS+Exp and 48s for CS ($p=0.04$, $CLES=0.59$). Moreover, when the explanation is provided, it takes more time to decide when the model contradicts the user's decision, but the difference to CS is not significant. Median time for CS output was 61s, compared to 84s for CS+Exp ($p=0.17$, MWU test). 

% \usepackage{multirow}
\begin{table}[]
\center
\caption{Time spent on individual decisions for cases 1-5 (CS) and 6-10 (CS+Ex). p values from Mann Whitney-U Test; significance: ns - non significant, $p>0.05$, * - $p<0.05$, ** - $p<0.01$, *** - $p<0.001$}
\label{tab:time_spent}
\begin{tabular}{lcccc} 
\toprule
\begin{tabular}[c]{@{}l@{}}Time [s]\\ median [Q1,Q3]\end{tabular} & \multicolumn{1}{l}{\begin{tabular}[c]{@{}l@{}}Action the same\\ as the recommendation\end{tabular}} & \multicolumn{1}{l}{\begin{tabular}[c]{@{}l@{}}Action against\\ the recommendation\end{tabular}} & \multicolumn{1}{l}{p} & \multicolumn{1}{l}{Significance}  \\
Cases 1-5, output: CS                                             & 48 [34,74]                                                                                          & 61 [39,85]                                                                                      & 0.097                 & ns                                \\
Cases 6-10, output: CS+Exp                                         & 38 [22,64]                                                                                          & 84 [50,116]                                                                                     & 0.000                 & ***                               \\
p                                                                 & 0.042                                                                                               & 0.170                                                                                           & \multicolumn{2}{l}{\multirow{2}{*}{}}                     \\
Significance                                                      & *                                                                                                   & ns                                                                                              & \multicolumn{2}{l}{}                                      \\
\bottomrule
\end{tabular}
\end{table}

\subsection{Positive feedback, but no attribution to the explanatory model}

\begin{figure}[h!]
\centering
\includegraphics[width= .9\textwidth]{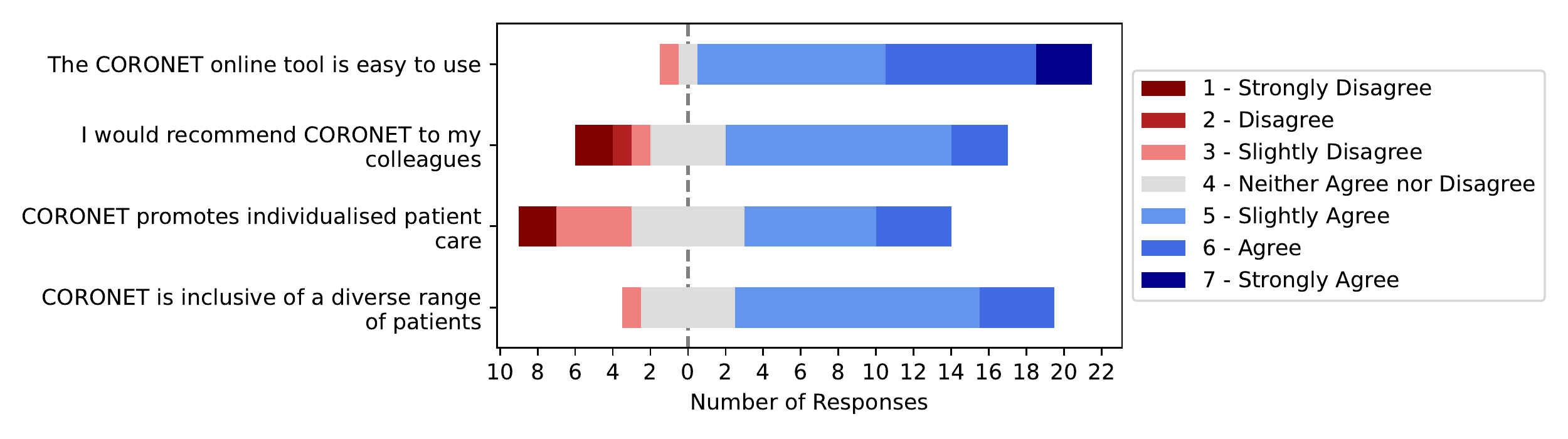}
\caption{Positive feedback regarding the CORONET evaluated at the end of the experiment.}
\label{fig:general_feedback}
\end{figure}

General feedback regarding the CORONET model is positive (Fig.\ref{fig:general_feedback}). The model and supporting interface is considered easy to use, and the majority of respondents would recommend the tool to their colleagues.

\begin{figure}[h!]
\centering
\includegraphics[width= .3\textwidth]{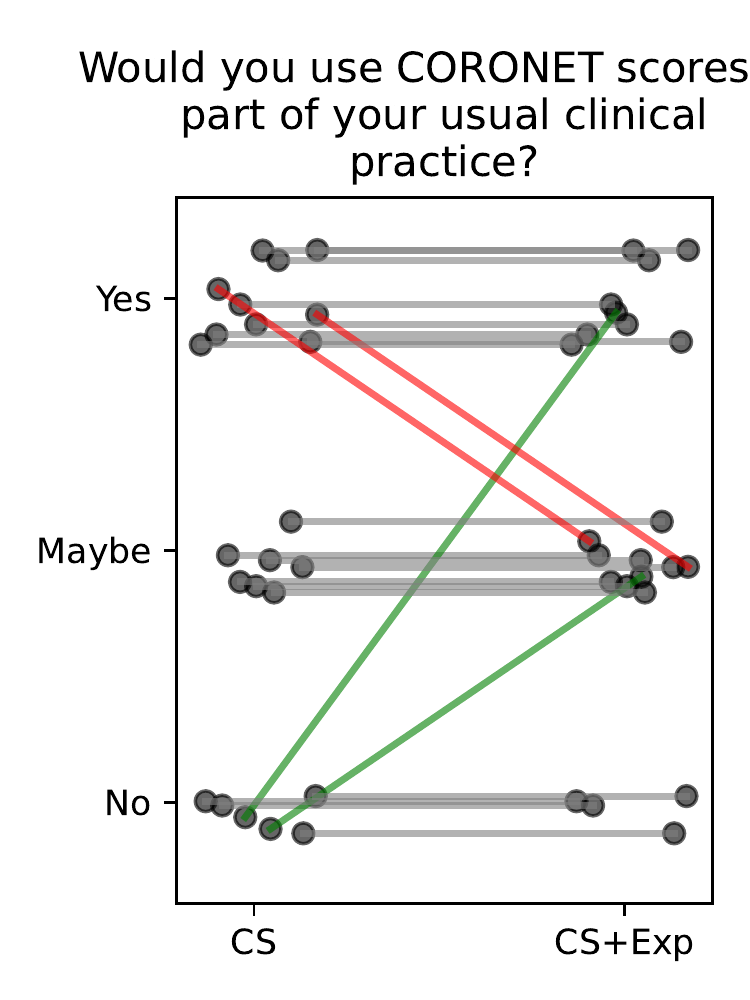}
\caption{Evaluation of clinical utility of the tool. Same question was asked twice, for CS and CS+Ex. No significant change ($p>0.05$).
}
\label{fig:would_you_use}
\end{figure}

We found no impact of the explanatory component on the clinical utility of the model (Fig.\ref{fig:would_you_use}), as the explanation did not improve the attitude towards whether to use the CORONET in clinical practice or not (\ref{itm:RQ5}, $p>0.05$, Wilcoxon signed-rank test).

\subsection{Respondents' engagement}
\begin{figure}[h!]
\centering
\includegraphics[width= .7\textwidth]{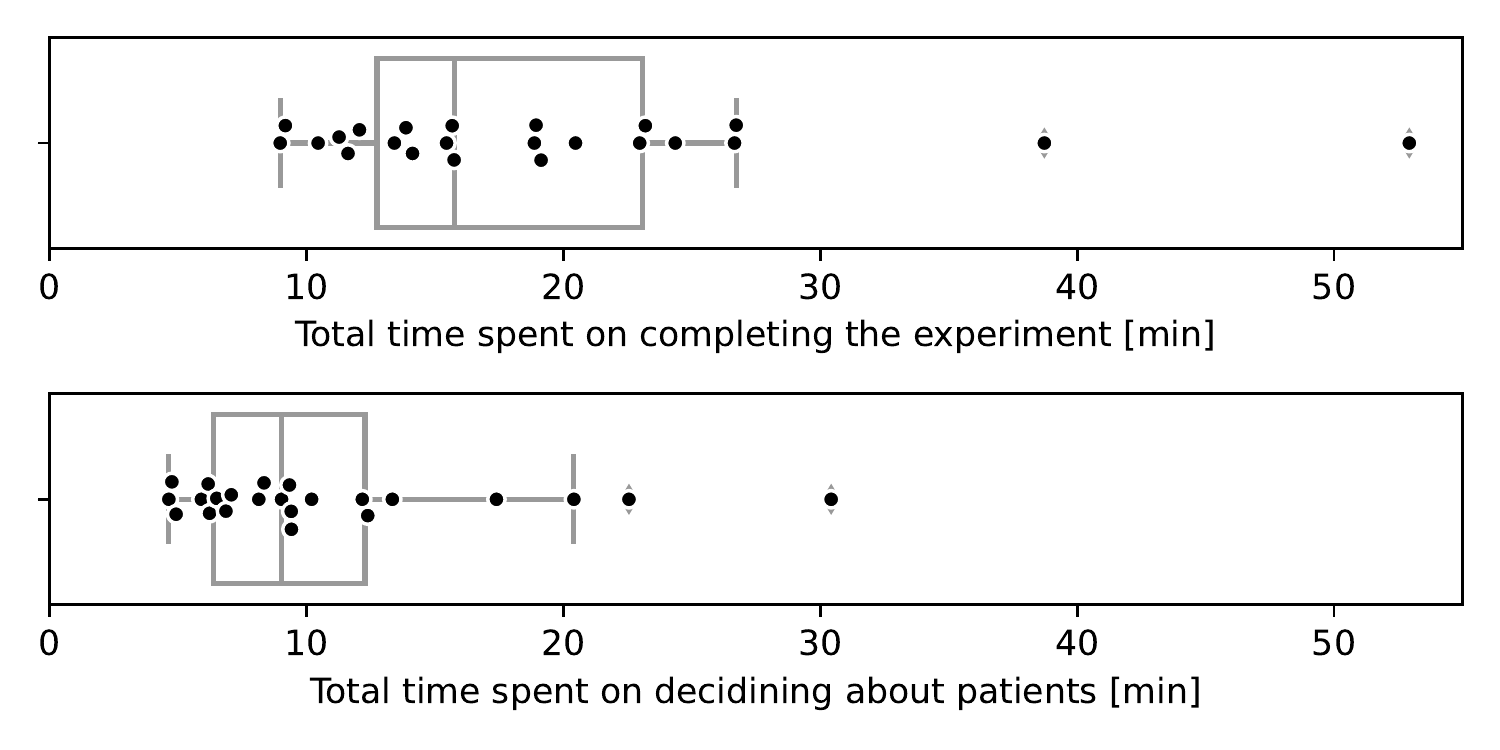}
\caption{Total time spent by 23 Healthcare Professionals.}
\label{fig:time_spent_on_questionaire}
\end{figure}

Our experiment is supported by an online questionnaire, which is to be completed without direct supervision and time limits. Although the questionnaire is estimated to require $\approx30$min, the actual average completion time was $\approx19$ min (median 16 min, Fig.\ref{fig:time_spent_on_questionaire}). Of note, nine HCPs completed it in less than 15 min, spending less than six min on deciding about 10 patients. We also identified several anomalies in the time spent in particular sections which we manually curated (see details in supplementary material). Most likely they were caused by occasional interruptions, which are inherent in the experiment conducted in the clinical setting, where the respondents may be distracted by more urgent matters.  

Overall, we did not find a significant difference between total time spent on deciding on patient cases between CS and CS+Exp ($p=0.846$, Wilcoxon signed-rank test, Fig.\ref{fig:time_pairedplot}). However, 56\% (13/23) HCPs spent less time when CS+Exp was provided. We did not find any correlation between this decrease and other aspects investigated in previous paragraphs.

% ----------------------------------------------------------------------------------

\section{Discussion: Pragmatic clinical embedding of explainable ML models}

\textbf{CORONET explanation among the existing taxonomies}
\newline The empirical analysis conducted in this study fits into a \textit{domain expert experiment with the exact application task}, which is an application-grounded evaluation, with real humans and real tasks (according to the taxonomy introduced in \cite{doshi-velezConsiderationsEvaluationGeneralization2018}).

In designing the CORONET output, we followed the \textit{perceptive interpretability} framework \cite{tjoaSurveyExplainableArtificial2021}, emphasising an immediate interpretation: \textit{you see, and you know}. We argue that the contribution plot follows this assumption.

According to taxonomies established in \cite{murdochDefinitionsMethodsApplications2019} and \cite{lakkarajuFaithfulCustomizableExplanations2019}, the output of the tool investigated in this study delivers:
\begin{itemize}
\item \textbf{modularity:} each interpretable element of the output (cognitive chunk) appears as a separate visual component. In the contribution plot, each bar delivers a meaningful portion of information and can be interpreted independently (e.g. low  contribution of albumin level to recommendation of `admission')
\item \textbf{relevance:} it provides insight for a particular audience, in this case clinicians seeking a patient's characteristics critical for the decision (contribution plot), and also how the patient in question can be located in the whole cohort (scatter plot). Additionally, the colour bar and score describe how severe the patient's outcome is expected to be.
\end{itemize}

However, CORONET's output does not deliver:
\begin{itemize}
\item \textbf{simulatability:} a model is simulatable when the user can reason and simulate how the model produces the output for an arbitrary input. It could be possible (assuming limited complexity of the model) for algorithms such decision tree or a set of rules. However, as CORONET uses a Random Forest regression model,  with multiple ($>$100) decision trees used to arrive at the prediction, the user cannot precisely elicit the formal decision process. 
\item \textbf{unambiguity:} CORONET provides local explanations, focusing on individualised patient care. Based on that, it is not possible to reveal how the model behaves in various parts of the feature space. Such behavior was investigated during model derivation \cite{coronet_paper} with the support of dependency plots. However, these are not presented to the user in the context of this study.
\end{itemize}

Analogously to unambiguity, the user is not able to evaluate the \textit{descriptive accuracy} \cite{murdochDefinitionsMethodsApplications2019}, which measures how well the relationship learned by the model is reflected in the explanation. Such evaluation was performed during model development, and then intentionally excluded from the cognitive chunks presented to the user. We argue that additional cognitive load would hamper the benefit of such information \cite{lyellAutomationBiasVerification2017,bucincaTrustThinkCognitive2021}.

%EXPECTATIONS of THE HCPs
\textbf{Characterising HCPs' attitude on the explanations}

More than 87\% of participants stated that knowing the contribution of features is important, both for the overall model and individual recommendations. The most important aspect, with the highest number of `strongly agree' answers, is to know the model’s uncertainty. This could suggest that HCPs, are self-reportedly more comfortable in assessing risks and uncertainty, rather than interpreting the feature contribution to the output of the ML model. Hence, delivering a model's uncertainty may contribute more to building trust in comparison to a break-down explanation of the features' contributions. However, this conclusion is solely based on the proportion of `strongly Agree', as we did not find significant differences between expectations regarding uncertainty, feature contribution and individual recommendation (p$>$0.05, Kruskal-Wallis test,  Supp Table \ref{tab:expectation_stats}). An intuition on the mathematical principles behind the model is perceived as significantly less important (p$<$0.05). 

Showing uncertainty increases trust and the likelihood of following the model prediction \cite{mcgrathWhenDoesUncertainty2020}, under a low cognitive load setting \cite{zhouEffectsUncertaintyCognitive2017}. Limited cognitive investment  in the model interpretation would result in a reduced understanding of the output. While the model used in our study does not explicitly provide uncertainty of its prediction, it recommends a binary action based on a numeric score and a threshold, which are shown to the HCPs. We argue it serves for the user as a proxy of the model's recommendation confidence.

%re need for showing uncertainty

% visuals easy
\textbf{Unmet need to build pragmatically grounded explanatory models}

Although the majority of HCPs (78\%; 18/23) found the explanation component accounting for the features' contributions easy to interpret, still 17\% of HCPs found the explanation of the modelmodel difficult (\ref{itm:RQ2}). This points to opportunities for the design and optimisation of explanatory models from the point of view of end-user interaction. Contemporary explanation methods such as SHAP or LIME focus on the calculation of a faithful explanation for the model, abstracting away the pragmatic aspects of this explanation (e.g. their interaction with domain experts). Recent work binding representation design and cognitive models \cite{raggiDissectingRepresentations2020,chengCognitivePropertiesRepresentations2021} may provide a formal avenue for designing more pragmatically efficient explanatory models.

%There is no consensus regarding the optimal format of the explanation (e.g. figure vs. text), type (e.g. feature contribution vs. set of rules vs. counterfactuals), amount of information provided and level of details [to add ref]. Furthermore, there has been significant debate as to  the main function of the explanation in the context of ML models% \cite{montaniArtificialIntelligenceClinical2019}. 

Explanations in the form of feature importance may act as a safety feature, drawing the user's attention to features omitted by initial judgment but marked as highly relevant by the model \cite{weertsHumanGroundedEvaluationSHAP2019}. According to \cite{lakkarajuHowFoolYou2020} the feature-based explanatory models can be in one of three groups: desired (expected by included in the model), ambivalent (user is indifferent about whether they are included or not) or prohibited features (expected to be omitted, following ethical ML principles). Displaying contributing features allows the user to verify whether the model utilises expected features instead of inconsistent or irrelevant features, based on supporting domain knowledge. In our study, we asked the HCPs explicitly whether the model included all desired features. Additionally, no prohibited features were recognized in the model. One of the downsides of feature importance is that it may lead to confirmation bias, both in the model derivation phase \cite{kumarProblemsShapleyvaluebasedExplanations2020} and in the post-hoc interpretation of the output. Practically, it is caused by noticing only the features that confirm previous assumptions and overlooking other, potentially incorrect features. 

% no impact of explanation

% explanation is not so perfect
\textbf{Ambiguity on the utility and perceived value of explanations}

A considerable number of HCPs did not change or even worsen their attitude towards the model after the explanation was provided (\ref{itm:RQ1}).
57\% of them reported the same level of understanding, and 17\% were understanding even less why the model provides recommendations different to their own.  
Satisfaction remained the same for 43\%, while 23\% were less satisfied with the explanation. %Perhaps due to disappointment as they awaited the opening of the black-box which hadn't occurred, because the models stills appeared opaque.
A possible explanation is an information overload (cognitive effort) effect or simply not meeting the expectations regarding the explanation model.

For 48\% and 26\% of HCPs the model with and without explanation was considered the same in terms of support for making safe clinical decisions and helping in ambiguous cases, respectively. For 26\% and 22\% the explanation even decreased the helpfulness. As high as 35\% felt less reassured when model provided the same recommendation as initial decision. We hypothesise, that the explanation could introduce further uncertainty in the clinical judgment, as it may have differed from the clinical assessment in particular cases.
This highlights previous adoption barriers towards ML-based decision support tools, i.e. lack of agreement with the model, lack of knowledge of the model, HCPs' attitude toward the model \cite{devarajBarriersFacilitatorsClinical2014,suttonOverviewClinicalDecision2020, evansExplainabilityParadoxChallenges2022}. Despite the widespread notion on the need for safe and explainable models, we found in our study that a step towards such models remained unnoticed or negatively perceived among almost half of the participants (\ref{itm:RQ3}). As suggested in \cite{evansExplainabilityParadoxChallenges2022}, initial skepticism regarding ML-based models may have caused such a rejection.

This implies that explainable ML researchers should further move from an algorithm-centric perspective \cite{tjoaSurveyExplainableArtificial2021}, to a pragmatic perspective, i.e. prioritising their pragmatic embedding within the experts' decision-making process. Feature contribution, well-received in the data science community, is not necessarily acknowledged among HCPs. Intensifying the dialog and creating feedback loops between explainable AI researchers and domain experts will be an essential part of the development of pragmatically relevant explainable ML models.

Although explainable ML is advocated as the methodological silver bullet for addressing transparency issues \cite{bauerExplAiNed2021}, some user studies, similarly to this paper, show evidence of a limited or even no impact of the explanation on performance in a task \cite{VANDERWAA2021103404,cartonFeatureBasedExplanationsDon2020,10.1145/3359152,poursabzi-sangdehManipulatingMeasuringModel2021,zhangEffectConfidenceExplanation2020}. This highlights the direction of further research in what explanation is expected depending on the context, e.g. clinical setting. Our study shows that we should be careful about our preconceived notions about what needs to be explained.

% less understanding
\textbf{For many HCPs, explanations did not deliver a critical understanding of the model}

Explanations did not improve the critical understanding on why the model recommends a different action to the HCPs for 57\% of the participants (\ref{itm:RQ4}). One possible reason may be a misunderstanding of the difference between local and global explanations. The barplot used to visualise feature importance shows the magnitude of the contribution of each feature to the recommendation for an individual patient. For each analysed case the bars can be reordered. This may lead to an extra complexity in interpretation and shows the simplistic design of standard explanation visualisation devices. However, when promoting more individualised patient care one should also expect higher variation in local explanations of the model. Second, the tool is oversimplifying the complexity of the patient and does not refer to the underlying deep biological processes which may explain a biomarker and its relation to the recommended action. The tool is focused on one aspect of the patient (in this scenario COVID-19) and a finite set of predictors. This is often overlooked by the HCPs, and the limitations of the model remained unidentified even with the explanation provided.

% most positive changes
\textbf{Explainable models help HCPs to address ambiguous cases}

The highest number of positive changes between CS and CS+Exp was observed to support clinical cases where HCPs were less confident in their decisions (\ref{itm:RQ5}). Interestingly, less experienced HCPs tended to trust the model more. This confirms the findings of \cite{goddardAutomationBiasEmpirical2014} and \cite{dowdingNursesUseComputerised2009a} where less experienced clinicians are more likely to rely on the recommendation, changing their initial decision. 
This suggests that the main function of a ML-based (data-driven) clinical decision support system, such as CORONET, is to be helpful when the user is uncertain, pragmatically positioning the model within the clinical workflow \cite{yangUnremarkableAIFitting2019}. This is aligned with the direct feedback received from HCPs (see Supp \ref{feedback_from_HCPs}). The recommended action, together with the explanation, may deliver new evidence and justifications for more ambiguous cases. %In this sense, it adds an additional element to the entire picture of a patient, i.e. important features worth taking into account and a recommendation that proves to be correct for the vast majority of patients in the validation cohort (assuming sufficient performance). 

\textbf{Explainable models help HCPs to acquire new domain knowledge}

When the user still lacks a consolidated domain prior knowledge, the explanation may act as a guideline or a source of new evidence-based knowledge, pointing into a second function of explainable ML models (\ref{itm:RQ5}). In \cite{schafferCanBetterYour2019} explanations only had an impact on HCPs who felt they had insufficient knowledge to achieve their given task, which may signify that the knowledge gap was addressed by the model's output. 

% LESS TIME SPENT FOR SET2
\textbf{Quicker decisions for explanatory models in contrast to black-box models}

57\% (13/23) of HCPs spent less time on decisions for cases 6-10 (the cases with explanations), when more information was delivered to end-users (\ref{itm:RQ6}). Potential reasons are: i) lost interest in the experiment; ii) gained fluency in interpreting the provided information. After the firstfivecases, HCPs familiarized with the layout of the decryption were able to decide on upcoming cases quicker; iii) provided explanations expedited the decision due to over-reliance on the tool. 
The questionnaire was not designed to verify points i) and ii) leaving this point of ambiguity. Cases 6-10 reflect the same level of difficulty from cases 1-5. 

% REDUCTION IN AUTOMATION BIAS
\textbf{Models and explanations can increase confirmation bias}

The results suggest that when the recommendation agrees with their initial decision, HCPs decide quicker, when compared to a model disagreement setting (\ref{itm:RQ6}). Furthermore, when explanations are provided, the decision is comparatively quicker. When the tool contradicts the clinician, explanations lead to a longer reflection on the decision. This highlights the risk of confirmation bias, possibly caused by the explanation that increases the reassurance on the decision (\ref{itm:RQ7}). This aligns to the results in \cite{bauerExplAiNed2021} where users tend to use the explanation to support the justification of their prior decision. The authors found that users reinforced pre-existing beliefs when the explanation supported them but did not abandon them when it was the opposite. Of note, model accuracy also affects the likelihood of adjusting prior beliefs. The risk of confirmation bias is particularly high when the explanation closely matches HCP's expectations \cite{evansExplainabilityParadoxChallenges2022}. In some cases, the explanation may appear reassuring even if it does not explain the model \cite{adebayoSanityChecksSaliency2020}. 
%Thus, a key aspect to improving how users interact with ML models should be better understanding as to how to increase trust in their outputs.

\textbf{Evidence of explanations reducing automation bias}
 
On the other hand, explanations drove users to reflect longer on the decision when disagreeing with the model outcome (\ref{itm:RQ6}). More time may indicate cognitive forcing, which is reported to reduce over-reliance on the model \cite{bucincaTrustThinkCognitive2021}.
At the same time, to reduce the automation bias, the explanation must reduce the cognitive load \cite{lyellAutomationBiasVerification2017,goddardAutomationBiasHidden2011}. Working memory constraints, together with increased cognitive load, lead to higher uncertainty in decision-making processes, thus increasing automation bias. This is particularly true in clinical settings where multiple contextual factors affect the reasoning process of HCPs \cite{ramaniExaminingPatternsUncertainty2020}.

% OVER-RELIANCE
\textbf{Over-reliance on the model's recommendation}

Some evidence in the literature suggests models that use explainability techniques can hamper the user's ability to detect when a serious error is made \cite{poursabzi-sangdehManipulatingMeasuringModel2021}. The explanation may excessively increase users' confidence in an algorithmic decision, communicating a false impression of correctness and rigour and resulting in decreased vigilance and auditing of the output \cite{ghassemiClinicalVisSupportingClinical2018, eibandImpactPlacebicExplanations2019}. In addition, \cite{SKITKA1999991} and  \cite{roviraEffectsImperfectAutomation2007} reported higher error rates when decisions relied too heavily on automation, resulting in bias towards a wrong action despite, evidence available which should have resulted in the user overriding the model. Automation bias is also linked to higher cognitive load and to consistently higher accuracy of the model \cite{lyellAutomationBiasVerification2017}, and familiarity with the task \cite{schafferCanBetterYour2019}. In our study, we identified over-reliance on the model for the cases where CORONET was intentionally wrong (\ref{itm:RQ6}). For these incorrect recommendations, the HCP should have overridden the decision and admitted the patient. Only 35\% (CS scenario) and 52\% (CS+Ex) HCPs took that critical perspective over the model.

\subsection{Assumptions \& limitations of this study}

\textbf{General concordance among clinical judgment.} 
In the assessment of AI models for decision support in healthcare, a well-known challenge is the definition of a ground truth \cite{chenEvaluationArtificialIntelligence2021}, as decisions may vary across HCPs. Thus, a panel of HCPs is recommended rather than a small number of annotators. In our study, the `correct' actions were defined by a panel of two experienced oncologists.

%Small Sample Size (from Jessica’s thesis)
%Due to the current pandemic, it is a particularly busy time for healthcare professionals and, understandably, it may be difficult for them to find free time to volunteer as a participant in the study. However, the responses we have received so far are of valuable detail and quality for improving future versions of CORONET.

\textbf{Lack of controlled environment.} 
An ideal environment for HCPs to participate would be at a face-to-face setting. Due to the pandemic and social distancing measures, the study was developed in an online setup. In this context, devices used by individual HCPs could not be controlled, which may have impacted on the consistency and the quantity of the usability feedback on each type of device. Despite this, the developers of CORONET designed its interface for cross-device compatibility (e.g.\ phones, tablets, and computers), and therefore by undertaking this method of design, we were able to capture feedback on CORONET’s usability based on a mixture of device types.
Due to remote setting and lack of control, potential additional factors may have impacted the results. However, the experimental design assumes performing a task that HCPs are familiarized with: admit or discharge a patient based on a detailed clinical description. Thus we argue that all participants had very good understanding of the task.

\textbf{Rush in completing the questionnaire.}
We argue that attempting to understand the model and therefore building trust in the tool require a substantial time commitment for some HCPs. Designing a study which is supervised, better controlled, and which better incentiveses HCPs for a longer time commitment would improve the current study design. 

%Considering the decision time time spent solely on deciding about the patients, we noticed that in some cases the decision was very quick, i.e. less than 25 seconds per patient, which we believe is a minimum time required to read all the information presented on the page. This rush may have affected the overall feedback regarding the explanation.

\textbf{Use of simulated patient cases.} 
To preserve the confidentiality of the patients, artificial patient case scenarios were used instead of real-life data. The cases were constructed in a clinically consistent manner by two domain experts. However, this does not take away the fact that in real life, a HCP would undertake clinical observations of the patient and account for any visible signs of distress and illness, such as any skin changes, unusual responses, and patient examination findings, which is not possible to reproduce in an artificial setting. 

\section{Conclusions}

In addition to the acceptability of the predictive model \cite{ghassemiFalseHopeCurrent2021,londonArtificialIntelligenceBlackBox2019}, a pragmatic evaluation of the user-model interaction is key to a successful deployment of ML-based recommendation tools in healthcare. 
The deployment depends not only on a explanation-interfaces \cite{holzingerHumanAIInterfaces2021} designed by software developers but to a greater extend, as shown in this study, on clinical performance evaluated by biomedical experts, initial attitudes and biases.
This paper contributes a pragmatic evaluation framework for explainable ML models for clinical decision support. The study used a within-subject design involving 23 healthcare professionals and compared an explainable ML model with a black-box model. 
Such a relatively small number of participants and their heterogeneity (i.e. various backgrounds and levels of expertise) may hinder drawing statistically robust conclusions. However, we argue that the study points into the direction of a more nuanced role of ML explanation models when these are pragmatically embedded in the clinical context and the results settle a solid base for further research. HCPs acknowledged the role of explanations as a safety and trust mechanism. Communicating the uncertainty behind the model emerged as a stronger requirement, when compared to the explanations.

Despite the general positive attitude towards explanations, for a significant set of participants (17-35\%), an undesirable effect for the explanations was observed, possibly due to an increase in cognitive effort. Moreover, explanations did not improve the critical understanding of the model for 57\% of the participants (i.e. their ability to detect an error in the model). We also found that explanations can increase confirmation bias, possibly accentuating over-reliance in the model. On the other hand, explanations drove HCPs to reflect longer on the decision when disagreeing with the model outcome, evidencing explanations as a possible mechanism for reducing automation bias. There was strong evidence that explainable models better supported HCPs to address ambiguous clinical cases (cases where HCPs were not certain about their decision). Also, explainable models helped less experienced HCPs to acquire new domain knowledge.

This work points to still open research questions in the area, including the need for further pragmatic evaluation of explainable models in complex clinical workflows, the co-development of models of explanations with domain experts and a better understanding of the dialogue between other model safety mechanisms and explanations.

\section*{Acknowledgements}
We would like to express our great gratitude to the healthcare professionals who participated in the experiment.
We appreciate that they voluntary devoted their time, so limited in the time of pandemic, to complete the experiment.
This paper and the research behind it would not have been possible without the significant feedback they delivered.

\section*{Funding}
%\begin{figure}[htb!]
%\centering
%\includegraphics[width= .9\textwidth]{figures/cce_dart_acknowledgements.png}
%\captionsetup{labelformat=empty}
%\caption{}
%\label{fig:cce_dart}
%\end{figure}

This project has received funding from the European Union's Horizon 2020 research and innovation programme under grant agreement No 965397.
Funding for developing the CORONET online tool has been provided by The Christie Charitable fund (1049751).
Dr Rebecca Lee is supported by the National Institute for Health Research.

%\newpage 

%\bibliography{eacl2021}
%\bibliographystyle{acl_natbib}

\printbibliography %Prints bibliography

@online{adebayoSanityChecksSaliency2020,
  title = {Sanity {{Checks}} for {{Saliency Maps}}},
  author = {Adebayo, Julius and Gilmer, Justin and Muelly, Michael and Goodfellow, Ian and Hardt, Moritz and Kim, Been},
  date = {2020-11-06},
  eprint = {1810.03292},
  eprinttype = {arxiv},
  primaryclass = {cs, stat},
  url = {http://arxiv.org/abs/1810.03292},
  %%urldate = {2021-11-16},
  abstract = {Saliency methods have emerged as a popular tool to highlight features in an input deemed relevant for the prediction of a learned model. Several saliency methods have been proposed, often guided by visual appeal on image data. In this work, we propose an actionable methodology to evaluate what kinds of explanations a given method can and cannot provide. We find that reliance, solely, on visual assessment can be misleading. Through extensive experiments we show that some existing saliency methods are independent both of the model and of the data generating process. Consequently, methods that fail the proposed tests are inadequate for tasks that are sensitive to either data or model, such as, finding outliers in the data, explaining the relationship between inputs and outputs that the model learned, and debugging the model. We interpret our findings through an analogy with edge detection in images, a technique that requires neither training data nor model. Theory in the case of a linear model and a single-layer convolutional neural network supports our experimental findings.},
  archiveprefix = {arXiv},
  keywords = {⛔ No DOI found,archived,Computer Science - Computer Vision and Pattern Recognition,Computer Science - Machine Learning,Statistics - Machine Learning},
  file = {C\:\\Users\\Lenovo\\Zotero\\storage\\VRLL9RTW\\Adebayo et al_2020_Sanity Checks for Saliency Maps.pdf;C\:\\Users\\Lenovo\\Zotero\\storage\\4GT2DIQP\\1810.html}
}

@report{alufaisanDoesExplainableArtificial2020,
  type = {preprint},
  title = {Does {{Explainable Artificial Intelligence Improve Human Decision-Making}}?},
  author = {Alufaisan, Yasmeen and Marusich, Laura Ranee and Bakdash, Jonathan Z and Zhou, Yan and Kantarcioglu, Murat},
  date = {2020-06-18},
  institution = {{PsyArXiv}},
  doi = {10.31234/osf.io/d4r9t},
  url = {https://osf.io/d4r9t},
  %%urldate = {2021-11-15},
  abstract = {Explainable AI provides insights to users into the why formodel predictions, offering potential for users to better un-derstand and trust a model, and to recognize and correct AIpredictions that are incorrect. Prior research on human andexplainable AI interactions has typically focused on measuressuch as interpretability, trust, and usability of the explanation.There are mixed findings whether explainable AI can improveactual human decision-making and the ability to identify theproblems with the underlying model. Using real datasets, wecompare objective human decision accuracy without AI (con-trol), with an AI prediction (no explanation), and AI predic-tion with explanation. We find providing any kind of AI pre-diction tends to improve user decision accuracy, but no con-clusive evidence that explainable AI has a meaningful impact.Moreover, we observed the strongest predictor for human de-cision accuracy was AI accuracy and that users were some-what able to detect when the AI was correct vs. incorrect, butthis was not significantly affected by including an explana-tion. Our results indicate that, at least in some situations, thewhy information provided in explainable AI may not enhanceuser decision-making, and further research may be needed tounderstand how to integrate explainable AI into real systems.},
  langid = {english},
  keywords = {read},
  file = {C\:\\Users\\d07321ow\\Documents\\alufaisanDoesExplainableArtificial2020 - Extracted Annotations (15112021, 015308)three different settings where a user needs to make decision 1) No AI predi.md;C\:\\Users\\Lenovo\\Zotero\\storage\\NJCDXYIQ\\Alufaisan et al. - 2020 - Does Explainable Artificial Intelligence Improve H.pdf}
}

@article{asanArtificialIntelligenceHuman2020,
  title = {Artificial {{Intelligence}} and {{Human Trust}} in {{Healthcare}}: {{Focus}} on {{Clinicians}}},
  shorttitle = {Artificial {{Intelligence}} and {{Human Trust}} in {{Healthcare}}},
  author = {Asan, Onur and Bayrak, Alparslan Emrah and Choudhury, Avishek},
  date = {2020-06-19},
  journaltitle = {Journal of Medical Internet Research},
  shortjournal = {J Med Internet Res},
  volume = {22},
  number = {6},
  pages = {e15154},
  issn = {1438-8871},
  doi = {10/ghxwzf},
  url = {http://www.jmir.org/2020/6/e15154/},
  %%urldate = {2021-11-17},
  abstract = {Artificial intelligence (AI) can transform health care practices with its increasing ability to translate the uncertainty and complexity in data into actionable—though imperfect—clinical decisions or suggestions. In the evolving relationship between humans and AI, trust is the one mechanism that shapes clinicians’ use and adoption of AI. Trust is a psychological mechanism to deal with the uncertainty between what is known and unknown. Several research studies have highlighted the need for improving AI-based systems and enhancing their capabilities to help clinicians. However, assessing the magnitude and impact of human trust on AI technology demands substantial attention. Will a clinician trust an AI-based system? What are the factors that influence human trust in AI? Can trust in AI be optimized to improve decision-making processes? In this paper, we focus on clinicians as the primary users of AI systems in health care and present factors shaping trust between clinicians and AI. We highlight critical challenges related to trust that should be considered during the development of any AI system for clinical use.},
  langid = {english},
  file = {C\:\\Users\\Lenovo\\Zotero\\storage\\7GFFWIMN\\Asan et al. - 2020 - Artificial Intelligence and Human Trust in Healthc.pdf}
}

@article{bauerExplAiNed2021,
  title = {Expl({{Ai}}){{Ned}}: {{The Impact}} of {{Explainable Artificial Intelligence}} on {{Cognitive Processes}}},
  shorttitle = {Expl({{Ai}}){{Ned}}},
  author = {Bauer, Kevin and von Zahn, Moritz and Hinz, Oliver},
  options = {useprefix=true},
  date = {2021},
  journaltitle = {SSRN Electronic Journal},
  shortjournal = {SSRN Journal},
  issn = {1556-5068},
  doi = {10/gm9786},
  url = {https://www.ssrn.com/abstract=3872711},
  %%urldate = {2021-11-02},
  abstract = {This paper explores the interplay of feature-based explainable AI (XAI) techniques, information processing, and human beliefs. Using a novel experimental protocol, we study the impact of providing users with explanations about how an AI system weighs inputted information to produce individual predictions (LIME) on users’ weighting of information and beliefs about the task-relevance of information. On the one hand, we find that feature-based explanations cause users to alter their mental weighting of available information according to observed explanations. On the other hand, explanations lead to asymmetric belief adjustments that we interpret as a manifestation of the confirmation bias. Trust in the prediction accuracy plays an important moderating role for XAI-enabled belief adjustments. Our results show that feature-based XAI does not only superficially influence decisions but really change internal cognitive processes, bearing the potential to manipulate human beliefs and reinforce stereotypes. Hence, the current regulatory efforts that aim at enhancing algorithmic transparency may benefit from going hand in hand with measures ensuring the exclusion of sensitive personal information in XAI systems. Overall, our findings put assertions that XAI is the silver bullet solving all of AI systems’ (black box) problems into perspective.},
  langid = {english},
  keywords = {read},
  file = {C\:\\Users\\d07321ow\\Documents\\bauerExplAiNed2021 - Extracted Annotations (03112021, 231158)Overall, our ndings put assertions that XAI is the silver bullet solving al.md;C\:\\Users\\Lenovo\\Zotero\\storage\\9MKULVII\\Bauer et al. - 2021 - Expl(Ai)Ned The Impact of Explainable Artificial .pdf}
}

@article{cartonFeatureBasedExplanationsDon2020,
  title = {Feature-{{Based Explanations Don}}'t {{Help People Detect Misclassifications}} of {{Online Toxicity}}},
  author = {Carton, Samuel and Mei, Qiaozhu and Resnick, Paul},
  date = {2020-05-26},
  journaltitle = {Proceedings of the International AAAI Conference on Web and Social Media},
  volume = {14},
  pages = {95--106},
  issn = {2334-0770},
  url = {https://ojs.aaai.org/index.php/ICWSM/article/view/7282},
  %%urldate = {2021-11-17},
  abstract = {We present an experimental assessment of the impact of feature attribution-style explanations on human performance in predicting the consensus toxicity of social media posts with advice from an unreliable machine learning model. By doing so we add to a small but growing body of literature inspecting the utility of interpretable machine learning in terms of human outcomes. We also evaluate interpretable machine learning for the first time in the important domain of online toxicity, where fully-automated methods have faced criticism as being inadequate as a measure of toxic behavior.We find that, contrary to expectations, explanations have no significant impact on accuracy or agreement with model predictions, through they do change the distribution of subject error somewhat while reducing the cognitive burden of the task for subjects. Our results contribute to the recognition of an intriguing expectation gap in the field of interpretable machine learning between the general excitement the field has engendered and the ambiguous results of recent experimental work, including this study.},
  langid = {english},
  keywords = {⛔ No DOI found,read},
  file = {C\:\\Users\\d07321ow\\Documents\\cartonFeatureBasedExplanationsDon2020 - Extracted Annotations (23112021, 012336)present an experimental assessment of the impact of feature attribution-sty.md;C\:\\Users\\Lenovo\\Zotero\\storage\\D2G28MFF\\Carton et al_2020_Feature-Based Explanations Don't Help People Detect Misclassifications of.pdf}
}

@article{chenEvaluationArtificialIntelligence2021,
  title = {Evaluation of Artificial Intelligence on a Reference Standard Based on Subjective Interpretation},
  author = {Chen, Po-Hsuan Cameron and Mermel, Craig H and Liu, Yun},
  date = {2021-11},
  journaltitle = {The Lancet Digital Health},
  shortjournal = {The Lancet Digital Health},
  volume = {3},
  number = {11},
  pages = {e693-e695},
  issn = {25897500},
  doi = {10/gmv7xj},
  url = {https://linkinghub.elsevier.com/retrieve/pii/S2589750021002168},
  %%urldate = {2021-11-11},
  langid = {english},
  keywords = {read},
  file = {C\:\\Users\\d07321ow\\Documents\\chenEvaluationArtificialIntelligence2021 - Extracted Annotations (12112021, 024052)The first method is straightforward graders should have relevant specialis.md;C\:\\Users\\Lenovo\\Zotero\\storage\\CRGIUULP\\Chen et al. - 2021 - Evaluation of artificial intelligence on a referen.pdf}
}

@incollection{doshi-velezConsiderationsEvaluationGeneralization2018,
  title = {Considerations for {{Evaluation}} and {{Generalization}} in {{Interpretable Machine Learning}}},
  booktitle = {Explainable and {{Interpretable Models}} in {{Computer Vision}} and {{Machine Learning}}},
  author = {Doshi-Velez, Finale and Kim, Been},
  editor = {Escalante, Hugo Jair and Escalera, Sergio and Guyon, Isabelle and Baró, Xavier and Güçlütürk, Yağmur and Güçlü, Umut and van Gerven, Marcel},
  options = {useprefix=true},
  date = {2018},
  series = {The {{Springer Series}} on {{Challenges}} in {{Machine Learning}}},
  pages = {3--17},
  publisher = {{Springer International Publishing}},
  location = {{Cham}},
  doi = {10.1007/978-3-319-98131-4_1},
  url = {http://link.springer.com/10.1007/978-3-319-98131-4_1},
  %%urldate = {2021-11-04},
  isbn = {978-3-319-98130-7 978-3-319-98131-4},
  langid = {english},
  keywords = {read},
  file = {C\:\\Users\\Lenovo\\Documents\\doshi-velezConsiderationsEvaluationGeneralization2018 - Extracted Annotations (14112021, 000847)Application-grounded Evaluation Real humans, real tasks (Doshi-Velez and.md;C\:\\Users\\Lenovo\\Zotero\\storage\\XTBUZ48Q\\Doshi-Velez and Kim - 2018 - Considerations for Evaluation and Generalization i.pdf}
}

@inproceedings{eibandImpactPlacebicExplanations2019,
  title = {The {{Impact}} of {{Placebic Explanations}} on {{Trust}} in {{Intelligent Systems}}},
  booktitle = {Extended {{Abstracts}} of the 2019 {{CHI Conference}} on {{Human Factors}} in {{Computing Systems}}},
  author = {Eiband, Malin and Buschek, Daniel and Kremer, Alexander and Hussmann, Heinrich},
  date = {2019-05-02},
  series = {{{CHI EA}} '19},
  pages = {1--6},
  publisher = {{Association for Computing Machinery}},
  location = {{New York, NY, USA}},
  doi = {10/gmv78z},
  url = {https://doi.org/10.1145/3290607.3312787},
  %%urldate = {2021-11-30},
  abstract = {Work in social psychology on interpersonal interaction has demonstrated that people are more likely to comply to a request if they are presented with a justification - even if this justification conveys no information. In the light of the many calls for explaining reasoning of interactive intelligent systems to users, we investigate whether this effect holds true for human-computer interaction. Using a prototype of a nutrition recommender, we conducted a lab study (N=30) between three groups (no explanation, placebic explanation, and real explanation). Our results indicate that placebic explanations for algorithmic decision-making may indeed invoke perceived levels of trust similar to real explanations. We discuss how placebic explanations could be considered in future work.},
  isbn = {978-1-4503-5971-9},
  keywords = {archived,explainability,explanations,intelligent systems,transparency,XAI},
  annotation = {https://web.archive.org/web/20211124052700/https://dl.acm.org/doi/10.1145/3290607.3312787},
  file = {C\:\\Users\\Lenovo\\Zotero\\storage\\ARDJWSI6\\Eiband et al_2019_The Impact of Placebic Explanations on Trust in Intelligent Systems.pdf}
}

@article{ExaminingEffectsPower,
author = {Taehyun Ha and Young June Sah and Yuri Park and Sangwon Lee},
title = {Examining the effects of power status of an explainable artificial intelligence system on users’ perceptions},
journal = {Behaviour \& Information Technology},
volume = {0},
number = {0},
pages = {1-13},
year  = {2020},
publisher = {Taylor & Francis},
doi = {10.1080/0144929X.2020.1846789},

URL = { 
        https://doi.org/10.1080/0144929X.2020.1846789
    
},
eprint = { 
        https://doi.org/10.1080/0144929X.2020.1846789
    
}

}

@article{ExplainingBlackboxClassifiers,
  title = {Explaining Black-Box Classifiers Using Post-Hoc Explanations-by-Example: {{The}} Effect of Explanations and Error-Rates in {{XAI}} User Studies},
  author = {Kenny, Eoin M. and Ford, Courtney and Quinn, Molly and Keane, Mark T.},
  year = {2021},
  journal = {Artificial Intelligence},
  volume = {294},
  pages = {103459},
  issn = {0004-3702},
  doi = {10.1016/j.artint.2021.103459}
}

@online{ghassemiClinicalVisSupportingClinical2018,
  title = {{{ClinicalVis}}: {{Supporting Clinical Task-Focused Design Evaluation}}},
  shorttitle = {{{ClinicalVis}}},
  author = {Ghassemi, Marzyeh and Pushkarna, Mahima and Wexler, James and Johnson, Jesse and Varghese, Paul},
  date = {2018-10-13},
  eprint = {1810.05798},
  eprinttype = {arxiv},
  primaryclass = {cs},
  url = {http://arxiv.org/abs/1810.05798},
  %%urldate = {2021-11-30},
  abstract = {Making decisions about what clinical tasks to prepare for is multi-factored, and especially challenging in intensive care environments where resources must be balanced with patient needs. Electronic health records (EHRs) are a rich data source, but are task-agnostic and can be difficult to use as summarizations of patient needs for a specific task, such as "could this patient need a ventilator tomorrow?" In this paper, we introduce ClinicalVis, an open-source EHR visualization-based prototype system for task-focused design evaluation of interactions between healthcare providers (HCPs) and EHRs. We situate ClinicalVis in a task-focused proof-of-concept design study targeting these interactions with real patient data. We conduct an empirical study of 14 HCPs, and discuss our findings on usability, accuracy, preference, and confidence in treatment decisions. We also present design implications that our findings suggest for future EHR interfaces, the presentation of clinical data for task-based planning, and evaluating task-focused HCP/EHR interactions in practice.},
  archiveprefix = {arXiv},
  keywords = {⛔ No DOI found,archived,Computer Science - Human-Computer Interaction,read},
  file = {C\:\\Users\\d07321ow\\Documents\\ghassemiClinicalVisSupportingClinical2018 - Extracted Annotations (01122021, 190927).md;C\:\\Users\\Lenovo\\Zotero\\storage\\C5R9QW94\\Ghassemi et al_2018_ClinicalVis.pdf;C\:\\Users\\Lenovo\\Zotero\\storage\\LEMF9BEQ\\1810.html}
}

@article{ghassemiFalseHopeCurrent2021,
  title = {The False Hope of Current Approaches to Explainable Artificial Intelligence in Health Care},
  author = {Ghassemi, Marzyeh and Oakden-Rayner, Luke and Beam, Andrew L.},
  date = {2021-11-01},
  journaltitle = {The Lancet Digital Health},
  shortjournal = {The Lancet Digital Health},
  volume = {3},
  number = {11},
  eprint = {34711379},
  eprinttype = {pmid},
  pages = {e745-e750},
  publisher = {{Elsevier}},
  issn = {2589-7500},
  doi = {10/gm8bdh},
  url = {https://www.thelancet.com/journals/landig/article/PIIS2589-7500(21)00208-9/fulltext},
  %%urldate = {2021-11-16},
  langid = {english},
  keywords = {read},
  file = {C\:\\Users\\d07321ow\\Documents\\ghassemiFalseHopeCurrent2021 - Extracted Annotations (16112021, 153747)false hope that individual users or those affected by AI will be able to j.md;C\:\\Users\\Lenovo\\Zotero\\storage\\ELHR835D\\Ghassemi et al_2021_The false hope of current approaches to explainable artificial intelligence in.pdf;C\:\\Users\\Lenovo\\Zotero\\storage\\8BSMVLEM\\fulltext.html}
}

@inproceedings{gilpinExplainingExplanationsOverview2018,
  title = {Explaining {{Explanations}}: {{An Overview}} of {{Interpretability}} of {{Machine Learning}}},
  shorttitle = {Explaining {{Explanations}}},
  booktitle = {2018 {{IEEE}} 5th {{International Conference}} on {{Data Science}} and {{Advanced Analytics}} ({{DSAA}})},
  author = {Gilpin, Leilani H. and Bau, David and Yuan, Ben Z. and Bajwa, Ayesha and Specter, Michael and Kagal, Lalana},
  date = {2018-10},
  pages = {80--89},
  doi = {10/ggccrr},
  eventtitle = {2018 {{IEEE}} 5th {{International Conference}} on {{Data Science}} and {{Advanced Analytics}} ({{DSAA}})}
}

@article{goddardAutomationBiasEmpirical2014,
  title = {Automation Bias: {{Empirical}} Results Assessing Influencing Factors},
  author = {Goddard, Kate and Roudsari, Abdul and Wyatt, Jeremy C.},
  date = {2014-05-01},
  journaltitle = {International Journal of Medical Informatics},
  shortjournal = {International Journal of Medical Informatics},
  volume = {83},
  number = {5},
  pages = {368--375},
  issn = {1386-5056},
  doi = {10/f2wz3k},
  url = {https://www.sciencedirect.com/science/article/pii/S1386505614000148}
}

@inproceedings{kumarProblemsShapleyvaluebasedExplanations2020,
  title = {Problems with {{Shapley-value-based}} Explanations as Feature Importance Measures},
  booktitle = {Proceedings of the 37th {{International Conference}} on {{Machine Learning}}},
  author = {Kumar, I. Elizabeth and Venkatasubramanian, Suresh and Scheidegger, Carlos and Friedler, Sorelle},
  date = {2020-11-21},
  pages = {5491--5500},
  publisher = {{PMLR}},
  issn = {2640-3498; https://web.archive.org/web/20211116001008/https://proceedings.mlr.press/v119/kumar20e.html},
  url = {https://proceedings.mlr.press/v119/kumar20e.html},
  %%urldate = {2021-11-16},
  abstract = {Game-theoretic formulations of feature importance have become popular as a way to "explain" machine learning models. These methods define a cooperative game between the features of a model and distribute influence among these input elements using some form of the game’s unique Shapley values. Justification for these methods rests on two pillars: their desirable mathematical properties, and their applicability to specific motivations for explanations. We show that mathematical problems arise when Shapley values are used for feature importance and that the solutions to mitigate these necessarily induce further complexity, such as the need for causal reasoning. We also draw on additional literature to argue that Shapley values do not provide explanations which suit human-centric goals of explainability.},
  eventtitle = {International {{Conference}} on {{Machine Learning}}},
  langid = {english},
  keywords = {archived,read},
  file = {C\:\\Users\\d07321ow\\Documents\\kumarProblemsShapleyvaluebasedExplanations2020 - Extracted Annotations (16112021, 014416)Secondly, taking a humancentric perspective, we evaluate Shapley-value-base.md;C\:\\Users\\Lenovo\\Zotero\\storage\\Z77G2TD7\\Kumar et al_2020_Problems with Shapley-value-based explanations as feature importance measures.pdf}
}

@inproceedings{lakkarajuFaithfulCustomizableExplanations2019,
  title = {Faithful and {{Customizable Explanations}} of {{Black Box Models}}},
  booktitle = {Proceedings of the 2019 {{AAAI}}/{{ACM Conference}} on {{AI}}, {{Ethics}}, and {{Society}}},
  author = {Lakkaraju, Himabindu and Kamar, Ece and Caruana, Rich and Leskovec, Jure},
  date = {2019-01-27},
  pages = {131--138},
  publisher = {{ACM}},
  location = {{Honolulu HI USA}},
  doi = {10/ggvzms},
  url = {https://dl.acm.org/doi/10.1145/3306618.3314229},
  %%urldate = {2021-11-01},
  abstract = {As predictive models increasingly assist human experts (e.g., doctors) in day-to-day decision making, it is crucial for experts to be able to explore and understand how such models behave in different feature subspaces in order to know if and when to trust them. To this end, we propose Model Understanding through Subspace Explanations (MUSE), a novel model agnostic framework which facilitates understanding of a given black box model by explaining how it behaves in subspaces characterized by certain features of interest. Our framework provides end users (e.g., doctors) with the flexibility of customizing the model explanations by allowing them to input the features of interest. The construction of explanations is guided by a novel objective function that we propose to simultaneously optimize for fidelity to the original model, unambiguity and interpretability of the explanation. More specifically, our objective allows us to learn, with optimality guarantees, a small number of compact decision sets each of which captures the behavior of a given black box model in unambiguous, well-defined regions of the feature space. Experimental evaluation with real-world datasets and user studies demonstrate that our approach can generate customizable, highly compact, easy-to-understand, yet accurate explanations of various kinds of predictive models compared to state-of-the-art baselines.},
  eventtitle = {{{AIES}} '19: {{AAAI}}/{{ACM Conference}} on {{AI}}, {{Ethics}}, and {{Society}}},
  isbn = {978-1-4503-6324-2},
  langid = {english},
  keywords = {read},
  file = {C\:\\Users\\d07321ow\\Documents\\lakkarajuFaithfulCustomizableExplanations2019 - Extracted Annotations (04112021, 193027)Model Understanding through Subspace Explanations (MUSE), which is designed.md;C\:\\Users\\Lenovo\\Zotero\\storage\\Q4DF88RE\\Lakkaraju et al. - 2019 - Faithful and Customizable Explanations of Black Bo.pdf}
}

@inproceedings{lakkarajuHowFoolYou2020,
  title = {"{{How}} Do {{I}} Fool You?": {{Manipulating User Trust}} via {{Misleading Black Box Explanations}}},
  shorttitle = {"{{How}} Do {{I}} Fool You?},
  booktitle = {Proceedings of the {{AAAI}}/{{ACM Conference}} on {{AI}}, {{Ethics}}, and {{Society}}},
  author = {Lakkaraju, Himabindu and Bastani, Osbert},
  date = {2020-02-07},
  pages = {79--85},
  publisher = {{ACM}},
  location = {{New York NY USA}},
  doi = {10/ghj6hb},
  url = {https://dl.acm.org/doi/10.1145/3375627.3375833},
  %%urldate = {2021-10-31},
  abstract = {As machine learning black boxes are increasingly being deployed in critical domains such as healthcare and criminal justice, there has been a growing emphasis on developing techniques for explaining these black boxes in a human interpretable manner. There has been recent concern that a high-fidelity explanation of a black box ML model may not accurately reflect the biases in the black box. As a consequence, explanations have the potential to mislead human users into trusting a problematic black box. In this work, we rigorously explore the notion of misleading explanations and how they influence user trust in black box models. Specifically, we propose a novel theoretical framework for understanding and generating misleading explanations, and carry out a user study with domain experts to demonstrate how these explanations can be used to mislead users. Our work is the first to empirically establish how user trust in black box models can be manipulated via misleading explanations.},
  eventtitle = {{{AIES}} '20: {{AAAI}}/{{ACM Conference}} on {{AI}}, {{Ethics}}, and {{Society}}},
  isbn = {978-1-4503-7110-0},
  langid = {english},
  keywords = {read},
  file = {C\:\\Users\\d07321ow\\Documents\\lakkarajuHowFoolYou2020 - Extracted Annotations (04112021, 175825)Poursabzi-Sangdeh et. al. (2018) show that longer explanations are harder f.md;C\:\\Users\\Lenovo\\Zotero\\storage\\XUA36S5C\\Lakkaraju and Bastani - 2020 - How do I fool you Manipulating User Trust via .pdf}
}

@online{liptonMythosModelInterpretability2017,
  title = {The {{Mythos}} of {{Model Interpretability}}},
  author = {Lipton, Zachary C.},
  date = {2017-03-06},
  eprint = {1606.03490},
  eprinttype = {arxiv},
  primaryclass = {cs, stat},
  url = {http://arxiv.org/abs/1606.03490},
  %%urldate = {2021-11-15},
  abstract = {Supervised machine learning models boast remarkable predictive capabilities. But can you trust your model? Will it work in deployment? What else can it tell you about the world? We want models to be not only good, but interpretable. And yet the task of interpretation appears underspecified. Papers provide diverse and sometimes non-overlapping motivations for interpretability, and offer myriad notions of what attributes render models interpretable. Despite this ambiguity, many papers proclaim interpretability axiomatically, absent further explanation. In this paper, we seek to refine the discourse on interpretability. First, we examine the motivations underlying interest in interpretability, finding them to be diverse and occasionally discordant. Then, we address model properties and techniques thought to confer interpretability, identifying transparency to humans and post-hoc explanations as competing notions. Throughout, we discuss the feasibility and desirability of different notions, and question the oft-made assertions that linear models are interpretable and that deep neural networks are not.},
  archiveprefix = {arXiv},
  keywords = {⛔ No DOI found,archived,Computer Science - Artificial Intelligence,Computer Science - Computer Vision and Pattern Recognition,Computer Science - Machine Learning,Computer Science - Neural and Evolutionary Computing,Statistics - Machine Learning},
  file = {C\:\\Users\\Lenovo\\Zotero\\storage\\FFUGF266\\Lipton_2017_The Mythos of Model Interpretability.pdf;C\:\\Users\\Lenovo\\Zotero\\storage\\DY3N3G9T\\1606.html}
}

@article{londonArtificialIntelligenceBlackBox2019,
  title = {Artificial {{Intelligence}} and {{Black-Box Medical Decisions}}: {{Accuracy}} versus {{Explainability}}},
  shorttitle = {Artificial {{Intelligence}} and {{Black-Box Medical Decisions}}},
  author = {London, Alex John},
  date = {2019},
  journaltitle = {Hastings Center Report},
  volume = {49},
  number = {1},
  pages = {15--21},
  issn = {1552-146X},
  doi = {10/ghvzdb},
  url = {https://onlinelibrary.wiley.com/doi/abs/10.1002/hast.973},
  %%urldate = {2021-12-01},
  abstract = {Although decision-making algorithms are not new to medicine, the availability of vast stores of medical data, gains in computing power, and breakthroughs in machine learning are accelerating the pace of their development, expanding the range of questions they can address, and increasing their predictive power. In many cases, however, the most powerful machine learning techniques purchase diagnostic or predictive accuracy at the expense of our ability to access “the knowledge within the machine.” Without an explanation in terms of reasons or a rationale for particular decisions in individual cases, some commentators regard ceding medical decision-making to black box systems as contravening the profound moral responsibilities of clinicians. I argue, however, that opaque decisions are more common in medicine than critics realize. Moreover, as Aristotle noted over two millennia ago, when our knowledge of causal systems is incomplete and precarious—as it often is in medicine—the ability to explain how results are produced can be less important than the ability to produce such results and empirically verify their accuracy.},
  langid = {english},
  keywords = {archived,read},
  annotation = {\_eprint: https://onlinelibrary.wiley.com/doi/pdf/10.1002/hast.973; https://web.archive.org/web/20211201174132/https://onlinelibrary.wiley.com/doi/abs/10.1002/hast.973},
  file = {C\:\\Users\\Lenovo\\Zotero\\storage\\7RF7QL6B\\London_2019_Artificial Intelligence and Black-Box Medical Decisions.pdf;C\:\\Users\\Lenovo\\Zotero\\storage\\3332YRV9\\hast.html}
}

@inproceedings{lundbergUnifiedApproachInterpreting2017,
  title = {A {{Unified Approach}} to {{Interpreting Model Predictions}}},
  booktitle = {Advances in {{Neural Information Processing Systems}}},
  author = {Lundberg, Scott M and Lee, Su-In},
  date = {2017},
  volume = {30},
  publisher = {{Curran Associates, Inc.}},
  url = {https://proceedings.neurips.cc/paper/2017/hash/8a20a8621978632d76c43dfd28b67767-Abstract.html},
  %%urldate = {2021-12-02},
  keywords = {⛔ No DOI found},
  file = {C\:\\Users\\Lenovo\\Zotero\\storage\\L9PCUJ5X\\Lundberg and Lee - 2017 - A Unified Approach to Interpreting Model Predictio.pdf}
}

@article{lyellAutomationBiasVerification2017,
  title = {Automation Bias and Verification Complexity: A Systematic Review},
  shorttitle = {Automation Bias and Verification Complexity},
  author = {Lyell, David and Coiera, Enrico},
  date = {2017-03-01},
  journaltitle = {Journal of the American Medical Informatics Association: JAMIA},
  shortjournal = {J Am Med Inform Assoc},
  volume = {24},
  number = {2},
  eprint = {27516495},
  eprinttype = {pmid},
  pages = {423--431},
  issn = {1527-974X},
  doi = {10/gng5ps},
  langid = {english},
  pmcid = {PMC7651899}
}

@online{mcgrathWhenDoesUncertainty2020,
  title = {When {{Does Uncertainty Matter}}?: {{Understanding}} the {{Impact}} of {{Predictive Uncertainty}} in {{ML Assisted Decision Making}}},
  shorttitle = {When {{Does Uncertainty Matter}}?},
  author = {McGrath, Sean and Mehta, Parth and Zytek, Alexandra and Lage, Isaac and Lakkaraju, Himabindu},
  date = {2020-11-13},
  eprint = {2011.06167},
  eprinttype = {arxiv},
  primaryclass = {cs},
  url = {http://arxiv.org/abs/2011.06167},
  %%urldate = {2021-10-31},
  abstract = {As machine learning (ML) models are increasingly being employed to assist human decision makers, it becomes critical to provide these decision makers with relevant inputs which can help them decide if and how to incorporate model predictions into their decision making. For instance, communicating the uncertainty associated with model predictions could potentially be helpful in this regard. However, there is little to no research that systematically explores if and how conveying predictive uncertainty impacts decision making. In this work, we carry out user studies to systematically assess how people respond to different types of predictive uncertainty i.e., posterior predictive distributions with different shapes and variances, in the context of ML assisted decision making. To the best of our knowledge, this work marks one of the first attempts at studying this question. Our results demonstrate that people are more likely to agree with a model prediction when they observe the corresponding uncertainty associated with the prediction. This finding holds regardless of the properties (shape or variance) of predictive uncertainty (posterior predictive distribution), suggesting that uncertainty is an effective tool for persuading humans to agree with model predictions. Furthermore, we also find that other factors such as domain expertise and familiarity with ML also play a role in determining how someone interprets and incorporates predictive uncertainty into their decision making.},
  archiveprefix = {arXiv},
  keywords = {⛔ No DOI found,archived,Computer Science - Machine Learning,read},
  file = {C\:\\Users\\d07321ow\\Documents\\mcgrathWhenDoesUncertainty2020 - Extracted Annotations (02112021, 225531).md;C\:\\Users\\Lenovo\\Zotero\\storage\\XHTVCUHK\\McGrath et al_2020_When Does Uncertainty Matter.pdf;C\:\\Users\\Lenovo\\Zotero\\storage\\HLTPCXGN\\2011.html}
}

@article{murdochDefinitionsMethodsApplications2019,
  title = {Definitions, Methods, and Applications in Interpretable Machine Learning},
  author = {Murdoch, W. James and Singh, Chandan and Kumbier, Karl and Abbasi-Asl, Reza and Yu, Bin},
  date = {2019-10-29},
  journaltitle = {Proceedings of the National Academy of Sciences},
  shortjournal = {Proc Natl Acad Sci USA},
  volume = {116},
  number = {44},
  pages = {22071--22080},
  issn = {0027-8424, 1091-6490},
  doi = {10/ggbhmq},
  url = {http://www.pnas.org/lookup/doi/10.1073/pnas.1900654116},
  %%urldate = {2021-11-04},
  abstract = {Machine-learning models have demonstrated great success in learning complex patterns that enable them to make predictions about unobserved data. In addition to using models for prediction, the ability to interpret what a model has learned is receiving an increasing amount of attention. However, this increased focus has led to considerable confusion about the notion of interpretability. In particular, it is unclear how the wide array of proposed interpretation methods are related and what common concepts can be used to evaluate them. We aim to address these concerns by defining interpretability in the context of machine learning and introducing the predictive, descriptive, relevant (PDR) framework for discussing interpretations. The PDR framework provides 3 overarching desiderata for evaluation: predictive accuracy, descriptive accuracy, and relevancy, with relevancy judged relative to a human audience. Moreover, to help manage the deluge of interpretation methods, we introduce a categorization of existing techniques into model-based and post hoc categories, with subgroups including sparsity, modularity, and simulatability. To demonstrate how practitioners can use the PDR framework to evaluate and understand interpretations, we provide numerous real-world examples. These examples highlight the often underappreciated role played by human audiences in discussions of interpretability. Finally, based on our framework, we discuss limitations of existing methods and directions for future work. We hope that this work will provide a common vocabulary that will make it easier for both practitioners and researchers to discuss and choose from the full range of interpretation methods.},
  langid = {english},
  keywords = {read},
  file = {C\:\\Users\\d07321ow\\Documents\\murdochDefinitionsMethodsApplications2019 - Extracted Annotations (14112021, 225632)We then introduce the predictive, descriptive, relevant (PDR) framework, co.md;C\:\\Users\\Lenovo\\Zotero\\storage\\QJGQWCS6\\Murdoch et al. - 2019 - Definitions, methods, and applications in interpre.pdf}
}

@online{poursabzi-sangdehManipulatingMeasuringModel2021,
  title = {Manipulating and {{Measuring Model Interpretability}}},
  author = {Poursabzi-Sangdeh, Forough and Goldstein, Daniel G. and Hofman, Jake M. and Vaughan, Jennifer Wortman and Wallach, Hanna},
  date = {2021-08-15},
  eprint = {1802.07810},
  eprinttype = {arxiv},
  primaryclass = {cs},
  url = {http://arxiv.org/abs/1802.07810},
  %%urldate = {2021-11-02},
  abstract = {With machine learning models being increasingly used to aid decision making even in high-stakes domains, there has been a growing interest in developing interpretable models. Although many supposedly interpretable models have been proposed, there have been relatively few experimental studies investigating whether these models achieve their intended effects, such as making people more closely follow a model's predictions when it is beneficial for them to do so or enabling them to detect when a model has made a mistake. We present a sequence of pre-registered experiments (N=3,800) in which we showed participants functionally identical models that varied only in two factors commonly thought to make machine learning models more or less interpretable: the number of features and the transparency of the model (i.e., whether the model internals are clear or black box). Predictably, participants who saw a clear model with few features could better simulate the model's predictions. However, we did not find that participants more closely followed its predictions. Furthermore, showing participants a clear model meant that they were less able to detect and correct for the model's sizable mistakes, seemingly due to information overload. These counterintuitive findings emphasize the importance of testing over intuition when developing interpretable models.},
  archiveprefix = {arXiv},
  keywords = {⛔ No DOI found,archived,Computer Science - Artificial Intelligence,Computer Science - Computers and Society,I.2},
  file = {C\:\\Users\\Lenovo\\Zotero\\storage\\8NN9JJHB\\Poursabzi-Sangdeh et al_2021_Manipulating and Measuring Model Interpretability.pdf;C\:\\Users\\Lenovo\\Zotero\\storage\\DSM68T2H\\1802.html}
}

@article{ramaniExaminingPatternsUncertainty2020,
  title = {Examining the Patterns of Uncertainty across Clinical Reasoning Tasks: Effects of Contextual Factors on the Clinical Reasoning Process},
  shorttitle = {Examining the Patterns of Uncertainty across Clinical Reasoning Tasks},
  author = {Ramani, Divya and Soh, Michael and Merkebu, Jerusalem and Durning, Steven J. and Battista, Alexis and McBee, Elexis and Ratcliffe, Temple and Konopasky, Abigail},
  date = {2020-08-27},
  journaltitle = {Diagnosis},
  volume = {7},
  number = {3},
  pages = {299--305},
  issn = {2194-802X, 2194-8011},
  doi = {10/gm9785},
  url = {https://www.degruyter.com/document/doi/10.1515/dx-2020-0019/html},
  %%urldate = {2021-11-02},
  abstract = {Objectives: Uncertainty is common in clinical reasoning given the dynamic processes required to come to a diagnosis. Though some uncertainty is expected during clinical encounters, it can have detrimental effects on clinical reasoning. Likewise, evidence has established the potentially detrimental effects of the presence of distracting contextual factors (i.e., factors other than case content needed to establish a diagnosis) in a clinical encounter on clinical reasoning. The purpose of this study was to examine how linguistic markers of uncertainty overlap with different clinical reasoning tasks and how distracting contextual factors might affect physicians’ clinical reasoning process.},
  langid = {english},
  keywords = {archived,read},
  annotation = {https://web.archive.org/web/20211102184100/https://www.degruyter.com/document/doi/10.1515/dx-2020-0019/html},
  file = {C\:\\Users\\d07321ow\\Documents\\ramaniExaminingPatternsUncertainty2020 - Extracted Annotations (03112021, 185834).md;C\:\\Users\\Lenovo\\Zotero\\storage\\S5XCL76F\\Ramani et al. - 2020 - Examining the patterns of uncertainty across clini.pdf}
}

@inproceedings{ribeiroAnchorsHighPrecision,
  title={Anchors: High-precision model-agnostic explanations},
  author={Ribeiro, Marco Tulio and Singh, Sameer and Guestrin, Carlos},
  booktitle={Proceedings of the AAAI conference on artificial intelligence},
  volume={32},
  number={1},
  year={2018}
}

@article{roviraEffectsImperfectAutomation2007,
  title = {Effects of {{Imperfect Automation}} on {{Decision Making}} in a {{Simulated Command}} and {{Control Task}}},
  author = {Rovira, Ericka and McGarry, Kathleen and Parasuraman, Raja},
  date = {2007-02-01},
  journaltitle = {Human Factors},
  shortjournal = {Hum Factors},
  volume = {49},
  number = {1},
  pages = {76--87},
  publisher = {{SAGE Publications Inc}},
  issn = {0018-7208},
  doi = {10/fmh7q6},
  url = {https://doi.org/10.1518/001872007779598082},
  %%urldate = {2021-11-28},
  abstract = {Objective: Effects of four types of automation support and two levels of automation reliability were examined. The objective was to examine the differential impact of information and decision automation and to investigate the costs of automation unreliability.  Background: Research has shown that imperfect automation can lead to differential effects of stages and levels of automation on human performance. Method: Eighteen participants performed a “sensor to shooter” targeting simulation of command and control. Dependent variables included accuracy and response time of target engagement decisions, secondary task performance, and subjective ratings of mental workload, trust, and self-confidence.  Results: Compared with manual performance, reliable automation significantly reduced decision times. Unreliable automation led to greater cost in decision-making accuracy under the higher automation reliability condition for three different forms of decision automation relative to information automation. At low automation reliability, however, there was a cost in performance for both information and decision automation. Conclusion: The results are consistent with a model of human-automation interaction that requires evaluation of the different stages of information processing to which automation support can be applied.  Application: If fully reliable decision automation cannot be guaranteed, designers should provide users with information automation support or other tools that allow for inspection and analysis of raw data.},
  langid = {english},
  keywords = {read},
  file = {C\:\\Users\\Lenovo\\Documents\\roviraEffectsImperfectAutomation2007 - Extracted Annotations (28112021, 152250)Compared with manual performance, reliable automation significantly reduced.md;C\:\\Users\\Lenovo\\Zotero\\storage\\JC7FW74E\\Rovira et al_2007_Effects of Imperfect Automation on Decision Making in a Simulated Command and.pdf}
}

@incollection{samekExplainableArtificialIntelligence2019,
  title = {Towards {{Explainable Artificial Intelligence}}},
  booktitle = {Explainable {{AI}}: {{Interpreting}}, {{Explaining}} and {{Visualizing Deep Learning}}},
  author = {Samek, Wojciech and Müller, Klaus-Robert},
  editor = {Samek, Wojciech and Montavon, Grégoire and Vedaldi, Andrea and Hansen, Lars Kai and Müller, Klaus-Robert},
  date = {2019},
  series = {Lecture {{Notes}} in {{Computer Science}}},
  volume = {11700},
  pages = {5--22},
  publisher = {{Springer International Publishing}},
  location = {{Cham}},
  doi = {10.1007/978-3-030-28954-6_1},
  url = {http://link.springer.com/10.1007/978-3-030-28954-6_1},
  %%urldate = {2021-11-16},
  abstract = {In recent years, machine learning (ML) has become a key enabling technology for the sciences and industry. Especially through improvements in methodology, the availability of large databases and increased computational power, today’s ML algorithms are able to achieve excellent performance (at times even exceeding the human level) on an increasing number of complex tasks. Deep learning models are at the forefront of this development. However, due to their nested nonlinear structure, these powerful models have been generally considered “black boxes”, not providing any information about what exactly makes them arrive at their predictions. Since in many applications, e.g., in the medical domain, such lack of transparency may be not acceptable, the development of methods for visualizing, explaining and interpreting deep learning models has recently attracted increasing attention. This introductory paper presents recent developments and applications in this field and makes a plea for a wider use of explainable learning algorithms in practice.},
  isbn = {978-3-030-28953-9 978-3-030-28954-6},
  langid = {english},
  file = {C\:\\Users\\Lenovo\\Zotero\\storage\\FFTYUIUS\\Samek and Müller - 2019 - Towards Explainable Artificial Intelligence.pdf}
}

@inproceedings{schafferCanBetterYour2019,
  title = {I Can Do Better than Your {{AI}}: Expertise and Explanations},
  shorttitle = {I Can Do Better than Your {{AI}}},
  booktitle = {Proceedings of the 24th {{International Conference}} on {{Intelligent User Interfaces}}},
  author = {Schaffer, James and O'Donovan, John and Michaelis, James and Raglin, Adrienne and Höllerer, Tobias},
  date = {2019-03-17},
  series = {{{IUI}} '19},
  pages = {240--251},
  publisher = {{Association for Computing Machinery}},
  location = {{New York, NY, USA}},
  doi = {10/ggkkjj},
  url = {https://doi.org/10.1145/3301275.3302308},
  %%urldate = {2021-11-28},
  abstract = {Intelligent assistants, such as navigation, recommender, and expert systems, are most helpful in situations where users lack domain knowledge. Despite this, recent research in cognitive psychology has revealed that lower-skilled individuals may maintain a sense of illusory superiority, which might suggest that users with the highest need for advice may be the least likely to defer judgment. Explanation interfaces - a method for persuading users to take a system's advice - are thought by many to be the solution for instilling trust, but do their effects hold for self-assured users? To address this knowledge gap, we conducted a quantitative study (N=529) wherein participants played a binary decision-making game with help from an intelligent assistant. Participants were profiled in terms of both actual (measured) expertise and reported familiarity with the task concept. The presence of explanations, level of automation, and number of errors made by the intelligent assistant were manipulated while observing changes in user acceptance of advice. An analysis of cognitive metrics lead to three findings for research in intelligent assistants: 1) higher reported familiarity with the task simultaneously predicted more reported trust but less adherence, 2) explanations only swayed people who reported very low task familiarity, and 3) showing explanations to people who reported more task familiarity led to automation bias.},
  isbn = {978-1-4503-6272-6},
  keywords = {cognitive modeling,decision support systems,human-computer interaction,information systems,intelligent assistants,read,user interfaces},
  file = {C\:\\Users\\Lenovo\\Documents\\schafferCanBetterYour2019 - Extracted Annotations (28112021, 185624)three findings for research in intelligent assistants 1) higher reported f.md;C\:\\Users\\Lenovo\\Zotero\\storage\\JKQC58M9\\Schaffer et al_2019_I can do better than your AI.pdf}
}

@article{SKITKA1999991,
  title = {Does Automation Bias Decision-Making?},
  author = {Skitka, Linda J. and Mosier, Kathleen L. and Burdick, Mark},
  date = {1999},
  journaltitle = {International Journal of Human-Computer Studies},
  volume = {51},
  number = {5},
  pages = {991--1006},
  issn = {1071-5819},
  doi = {10/bg5rb7},
  url = {https://www.sciencedirect.com/science/article/pii/S1071581999902525}
}

@article{tjoaSurveyExplainableArtificial2021,
  title = {A {{Survey}} on {{Explainable Artificial Intelligence}} ({{XAI}}): {{Toward Medical XAI}}},
  shorttitle = {A {{Survey}} on {{Explainable Artificial Intelligence}} ({{XAI}})},
  author = {Tjoa, Erico and Guan, Cuntai},
  date = {2021-11},
  journaltitle = {IEEE Transactions on Neural Networks and Learning Systems},
  volume = {32},
  number = {11},
  pages = {4793--4813},
  issn = {2162-2388},
  doi = {10.1109/TNNLS.2020.3027314}
}

@article{VANDERWAA2021103404,
  title = {Evaluating {{XAI}}: {{A}} Comparison of Rule-Based and Example-Based Explanations},
  author = {van der Waa, Jasper and Nieuwburg, Elisabeth and Cremers, Anita and Neerincx, Mark},
  options = {useprefix=true},
  date = {2021},
  journaltitle = {Artificial Intelligence},
  volume = {291},
  pages = {103404},
  issn = {0004-3702},
  doi = {10/gjm97h},
  url = {https://www.sciencedirect.com/science/article/pii/S0004370220301533}
}

@online{weertsHumanGroundedEvaluationSHAP2019,
  title = {A {{Human-Grounded Evaluation}} of {{SHAP}} for {{Alert Processing}}},
  author = {Weerts, Hilde J. P. and van Ipenburg, Werner and Pechenizkiy, Mykola},
  options = {useprefix=true},
  date = {2019-07-07},
  eprint = {1907.03324},
  eprinttype = {arxiv},
  primaryclass = {cs, stat},
  url = {http://arxiv.org/abs/1907.03324},
  %%urldate = {2021-11-15},
  abstract = {In the past years, many new explanation methods have been proposed to achieve interpretability of machine learning predictions. However, the utility of these methods in practical applications has not been researched extensively. In this paper we present the results of a human-grounded evaluation of SHAP, an explanation method that has been well-received in the XAI and related communities. In particular, we study whether this local model-agnostic explanation method can be useful for real human domain experts to assess the correctness of positive predictions, i.e. alerts generated by a classifier. We performed experimentation with three different groups of participants (159 in total), who had basic knowledge of explainable machine learning. We performed a qualitative analysis of recorded reflections of experiment participants performing alert processing with and without SHAP information. The results suggest that the SHAP explanations do impact the decision-making process, although the model's confidence score remains to be a leading source of evidence. We statistically test whether there is a significant difference in task utility metrics between tasks for which an explanation was available and tasks in which it was not provided. As opposed to common intuitions, we did not find a significant difference in alert processing performance when a SHAP explanation is available compared to when it is not.},
  archiveprefix = {arXiv},
  keywords = {⛔ No DOI found,archived,Computer Science - Human-Computer Interaction,Computer Science - Machine Learning,read,Statistics - Machine Learning},
  file = {C\:\\Users\\d07321ow\\Documents\\weertsHumanGroundedEvaluationSHAP2019 - Extracted Annotations (15112021, 204222)As opposed to common intuitions, we did not find a significant difference i.md;C\:\\Users\\Lenovo\\Zotero\\storage\\VEKDPUZA\\Weerts et al_2019_A Human-Grounded Evaluation of SHAP for Alert Processing.pdf;C\:\\Users\\Lenovo\\Zotero\\storage\\73PXGLKN\\1907.html}
}

@inproceedings{zhangEffectConfidenceExplanation2020,
  title = {Effect of Confidence and Explanation on Accuracy and Trust Calibration in {{AI-assisted}} Decision Making},
  booktitle = {Proceedings of the 2020 {{Conference}} on {{Fairness}}, {{Accountability}}, and {{Transparency}}},
  author = {Zhang, Yunfeng and Liao, Q. Vera and Bellamy, Rachel K. E.},
  date = {2020-01-27},
  series = {{{FAT}}* '20},
  pages = {295--305},
  publisher = {{Association for Computing Machinery}},
  location = {{New York, NY, USA}},
  doi = {10/ggjpcr},
  url = {https://doi.org/10.1145/3351095.3372852},
  %%urldate = {2021-12-01},
  abstract = {Today, AI is being increasingly used to help human experts make decisions in high-stakes scenarios. In these scenarios, full automation is often undesirable, not only due to the significance of the outcome, but also because human experts can draw on their domain knowledge complementary to the model's to ensure task success. We refer to these scenarios as AI-assisted decision making, where the individual strengths of the human and the AI come together to optimize the joint decision outcome. A key to their success is to appropriately calibrate human trust in the AI on a case-by-case basis; knowing when to trust or distrust the AI allows the human expert to appropriately apply their knowledge, improving decision outcomes in cases where the model is likely to perform poorly. This research conducts a case study of AI-assisted decision making in which humans and AI have comparable performance alone, and explores whether features that reveal case-specific model information can calibrate trust and improve the joint performance of the human and AI. Specifically, we study the effect of showing confidence score and local explanation for a particular prediction. Through two human experiments, we show that confidence score can help calibrate people's trust in an AI model, but trust calibration alone is not sufficient to improve AI-assisted decision making, which may also depend on whether the human can bring in enough unique knowledge to complement the AI's errors. We also highlight the problems in using local explanation for AI-assisted decision making scenarios and invite the research community to explore new approaches to explainability for calibrating human trust in AI.},
  isbn = {978-1-4503-6936-7},
  keywords = {archived,confidence,decision support,explainable AI,trust},
  annotation = {https://web.archive.org/web/20211202010405/https://dl.acm.org/doi/10.1145/3351095.3372852},
  file = {C\:\\Users\\Lenovo\\Zotero\\storage\\2S46AZA5\\Zhang et al. - 2020 - Effect of confidence and explanation on accuracy a.pdf}
}

@incollection{zhouEffectsUncertaintyCognitive2017,
  title = {Effects of {{Uncertainty}} and {{Cognitive Load}} on {{User Trust}} in {{Predictive Decision Making}}},
  booktitle = {Human-{{Computer Interaction}} – {{INTERACT}} 2017},
  author = {Zhou, Jianlong and Arshad, Syed Z. and Luo, Simon and Chen, Fang},
  editor = {Bernhaupt, Regina and Dalvi, Girish and Joshi, Anirudha and K. Balkrishan, Devanuj and O’Neill, Jacki and Winckler, Marco},
  date = {2017},
  volume = {10516},
  pages = {23--39},
  publisher = {{Springer International Publishing}},
  location = {{Cham}},
  doi = {10.1007/978-3-319-68059-0_2},
  url = {https://link.springer.com/10.1007/978-3-319-68059-0_2},
  %%urldate = {2021-11-02},
  abstract = {Rapid increase of data in different fields has been resulting in wide applications of Machine Learning (ML) based intelligent systems in predictive decision making scenarios. Unfortunately, these systems appear like a ‘black-box’ to users due to their complex working mechanisms and therefore significantly affect the user’s trust in human-machine interactions. This is partly due to the tightly coupled uncertainty inherent in the ML models that underlie the predictive decision making recommendations. Furthermore, when such analytics-driven intelligent systems are used in modern complex high-risk domains (such as aviation) - user decisions, in addition to trust, are also influenced by higher levels of cognitive load. This paper investigates effects of uncertainty and cognitive load on user trust in predictive decision making in order to design effective user interfaces for such ML-based intelligent systems. Our user study of 42 subjects in a repeated factorial design experiment found that both uncertainty types (risk and ambiguity) and cognitive workload levels affected user trust in predictive decision making. Uncertainty presentation leads to increased trust but only under low cognitive load conditions when users had sufficient cognitive resources to process the information. Presentation of uncertainty under high load conditions (when cognitive resources were short in supply) leads to a decrease of trust in the system and its recommendations.},
  isbn = {978-3-319-68058-3 978-3-319-68059-0},
  langid = {english},
  keywords = {read},
  file = {C\:\\Users\\d07321ow\\Documents\\zhouEffectsUncertaintyCognitive2017 - Extracted Annotations (03112021, 000728)42 subjects (Zhou et al 201723)Uncertainty presentation leads to increas.md;C\:\\Users\\Lenovo\\Zotero\\storage\\3WYAM9I2\\Zhou et al. - 2017 - Effects of Uncertainty and Cognitive Load on User .pdf}
}

@article{10.1145/3359152,
author = {Green, Ben and Chen, Yiling},
title = {The Principles and Limits of Algorithm-in-the-Loop Decision Making},
year = {2019},
issue_date = {November 2019},
publisher = {Association for Computing Machinery},
address = {New York, NY, USA},
volume = {3},
number = {CSCW},
url = {https://doi.org/10.1145/3359152},
doi = {10.1145/3359152},
journal = {Proc. ACM Hum.-Comput. Interact.},
month = {nov},
articleno = {50},
numpages = {24}
}

@article{bucincaTrustThinkCognitive2021,
  title = {To {{Trust}} or to {{Think}}: {{Cognitive Forcing Functions Can Reduce Overreliance}} on {{AI}} in {{AI-assisted Decision-making}}},
  shorttitle = {To {{Trust}} or to {{Think}}},
  author = {Buçinca, Zana and Malaya, Maja Barbara and Gajos, Krzysztof Z.},
  date = {2021-04-22},
  journaltitle = {Proceedings of the ACM on Human-Computer Interaction},
  shortjournal = {Proc. ACM Hum.-Comput. Interact.},
  volume = {5},
  pages = {188:1--188:21},
  doi = {10.1145/3449287},
  url = {https://doi.org/10.1145/3449287},
  %%urldate = {2022-03-03},
  abstract = {People supported by AI-powered decision support tools frequently overrely on the AI: they accept an AI's suggestion even when that suggestion is wrong. Adding explanations to the AI decisions does not appear to reduce the overreliance and some studies suggest that it might even increase it. Informed by the dual-process theory of cognition, we posit that people rarely engage analytically with each individual AI recommendation and explanation, and instead develop general heuristics about whether and when to follow the AI suggestions. Building on prior research on medical decision-making, we designed three cognitive forcing interventions to compel people to engage more thoughtfully with the AI-generated explanations. We conducted an experiment (N=199), in which we compared our three cognitive forcing designs to two simple explainable AI approaches and to a no-AI baseline. The results demonstrate that cognitive forcing significantly reduced overreliance compared to the simple explainable AI approaches. However, there was a trade-off: people assigned the least favorable subjective ratings to the designs that reduced the overreliance the most. To audit our work for intervention-generated inequalities, we investigated whether our interventions benefited equally people with different levels of Need for Cognition (i.e., motivation to engage in effortful mental activities). Our results show that, on average, cognitive forcing interventions benefited participants higher in Need for Cognition more. Our research suggests that human cognitive motivation moderates the effectiveness of explainable AI solutions.},
  issue = {CSCW1},
  keywords = {archived,artificial intelligence,cognition,explanations,read,trust},
  annotation = {https://web.archive.org/web/20220225182226/https://dl.acm.org/action/cookieAbsent},
  file = {C\:\\Users\\d07321ow\\Documents\\bucincaTrustThinkCognitive2021 - Extracted Annotations (03032022, 210659).md;C\:\\Users\\d07321ow\\Zotero\\storage\\S3J9PFZQ\\Buçinca et al_2021_To Trust or to Think.pdf}
}

@article{devarajBarriersFacilitatorsClinical2014,
  title = {Barriers and {{Facilitators}} to {{Clinical Decision Support Systems Adoption}}: {{A Systematic Review}}},
  shorttitle = {Barriers and {{Facilitators}} to {{Clinical Decision Support Systems Adoption}}},
  author = {Devaraj, Srikant and Sharma, Sushil K. and Fausto, Dyan J. and Viernes, Sara and Kharrazi, Hadi},
  date = {2014-07-24},
  journaltitle = {Journal of Business Administration Research},
  volume = {3},
  number = {2; https://web.archive.org/web/20220303231827/https://www.sciedu.ca/journal/index.php/jbar/article/view/5217},
  pages = {36},
  issn = {1927-9515},
  doi = {10.5430/jbar.v3n2p36},
  url = {https://www.sciedu.ca/journal/index.php/jbar/article/view/5217},
  %%urldate = {2022-03-03},
  abstract = {The objective of the study was to identify potential barriers and facilitators to improve clinical practice using computer-based Clinical Decision Support System (CDSS). Studies published since 2000 were found using PubMed database, PsychInfo, CINAHL, EBSCOhost database, and Google scholar. Twenty-six relevant publications were examined. Thirty-five unique barriers and twenty-five unique facilitators were identified in the literature as important determinants of CDSS’s adoption in clinical practice. The list of barriers and facilitators collected from each study were then organized under the four dimensions of The Unified Theory of Acceptance and Use of Technology (UTAUT) model: performance expectancy, effort expectancy, social influence, and facilitating conditions. Some of the important barriers to CDSS use include; lack of time or time constraints, economic constraints (e.g., finance and resources), lack of knowledge of system or content, reluctance to use system in front of patients, obscure workflow issues, less authenticity or reliability of information, lack of agreement with the system, and physician or user attitude toward the system. The study contributes immensely to the literature by identifying the important barriers and facilitators of CDSS.},
  issue = {2},
  langid = {english},
  keywords = {archived,read},
  file = {C\:\\Users\\d07321ow\\Documents\\devarajBarriersFacilitatorsClinical2014 - Extracted Annotations (04032022, 002410).md;C\:\\Users\\d07321ow\\Zotero\\storage\\2RFFG6ZP\\Devaraj et al_2014_Barriers and Facilitators to Clinical Decision Support Systems Adoption.pdf}
}

@article{dowdingNursesUseComputerised2009a,
  title = {Nurses’ Use of Computerised Clinical Decision Support Systems: A Case Site Analysis},
  shorttitle = {Nurses’ Use of Computerised Clinical Decision Support Systems},
  author = {Dowding, Dawn and Mitchell, Natasha and Randell, Rebecca and Foster, Rebecca and Lattimer, Valerie and Thompson, Carl},
  date = {2009},
  journaltitle = {Journal of Clinical Nursing},
  volume = {18},
  number = {8},
  pages = {1159--1167},
  issn = {1365-2702},
  doi = {10.1111/j.1365-2702.2008.02607.x},
  url = {https://onlinelibrary.wiley.com/doi/abs/10.1111/j.1365-2702.2008.02607.x},
  %%urldate = {2022-03-03},
  abstract = {Aims and objectives. To explore how nurses use computerised clinical decision support systems in clinical practice and the factors that influence use. Background. There is limited evidence for the benefits of computerised clinical decision support systems in nursing, with the majority of existing research focusing on nurses’ use of decision support for telephone triage. Research has suggested that several factors including nurses’ experience, features of the technology system and organisational factors may influence how decision support is used in practice. Design. A multiple case site study. Methods. Four case sites were purposively selected to provide variation in staff experience, technology used and decisions supported by the technology. Data were collected in each case site using non-participant observation of nurse/patient consultations (n = 115) and interviews with nurses (n = 55). Data were analysed using thematic content analysis. Results. Computerised decision support systems were used in a variety of ways by nurses, including recording information, monitoring patients’ progress and confirming decisions that had already been made. Nurses’ experience with the decision and the technology affected how they used a decision support system and whether or not they over-rode recommendations made by the system. The ability of nurses to adapt the technology also affected its use. Conclusions. How nurses use computerised decision support appears to be the result of an interaction between a nurses’ experience and their ability to adapt the technology to ‘fit’ with local clinical practice. Relevance to clinical practice. One of the stated aims of introducing computerised decision support systems to assist nursing practice is to reduce variation and/or the number of errors associated with clinical practice. The study found unanticipated uses in such systems such as the routine over-riding of recommendations which could lead to an increase rather than a decrease in variation or errors.},
  langid = {english},
  keywords = {case study research,clinical decision-making,computerised decision-making,nursing,read,technology},
  annotation = {\_eprint: https://onlinelibrary.wiley.com/doi/pdf/10.1111/j.1365-2702.2008.02607.x},
  file = {C\:\\Users\\d07321ow\\Documents\\dowdingNursesUseComputerised2009a - Extracted Annotations (04032022, 004839).md;C\:\\Users\\d07321ow\\Zotero\\storage\\3IQ2JCD8\\Dowding et al_2009_Nurses’ use of computerised clinical decision support systems.pdf;C\:\\Users\\d07321ow\\Zotero\\storage\\6FCXN2DH\\j.1365-2702.2008.02607.html}
}

@article{goddardAutomationBiasHidden2011,
  title = {Automation Bias - {{A}} Hidden Issue for Clinical Decision Support System Use},
  author = {Goddard, Kate and Roudsari, Abdul and Wyatt, Jeremy},
  date = {2011-01-01},
  journaltitle = {Studies in health technology and informatics},
  shortjournal = {Studies in health technology and informatics},
  volume = {164},
  pages = {17--22},
  doi = {10.3233/978-1-60750-709-3-17}
}

@incollection{hudonExplainableArtificialIntelligence2021,
  title = {Explainable {{Artificial Intelligence}} ({{XAI}}): {{How}} the {{Visualization}} of {{AI Predictions Affects User Cognitive Load}} and {{Confidence}}},
  shorttitle = {Explainable {{Artificial Intelligence}} ({{XAI}})},
  booktitle = {Information {{Systems}} and {{Neuroscience}}},
  author = {Hudon, Antoine and Demazure, Théophile and Karran, Alexander and Léger, Pierre-Majorique and Sénécal, Sylvain},
  editor = {Davis, Fred D. and Riedl, René and vom Brocke, Jan and Léger, Pierre-Majorique and Randolph, Adriane B. and Müller-Putz, Gernot},
  options = {useprefix=true},
  date = {2021},
  series = {Lecture {{Notes}} in {{Information Systems}} and {{Organisation}}},
  volume = {52},
  pages = {237--246},
  publisher = {{Springer International Publishing}},
  location = {{Cham}},
  doi = {10.1007/978-3-030-88900-5_27},
  url = {https://link.springer.com/10.1007/978-3-030-88900-5_27},
  %%urldate = {2022-03-03},
  abstract = {Explainable Artificial Intelligence (XAI) aims to bring transparency to AI systems by translating, simplifying, and visualizing its decisions. While society remains skeptical about AI systems, studies show that transparent and explainable AI systems result in improved confidence between humans and AI. We present preliminary results from a study designed to assess two presentation-order methods and three AI decision visualization attribution models to determine each visualization’s impact upon a user’s cognitive load and confidence in the system by asking participants to complete a visual decision-making task. The results show that both the presentation order and the morphological clarity impact cognitive load. Furthermore, a negative correlation was revealed between cognitive load and confidence in the AI system. Our findings have implications for future AI systems design, which may facilitate better collaboration between humans and AI.},
  isbn = {978-3-030-88899-2 978-3-030-88900-5},
  langid = {english},
  keywords = {read},
  file = {C\:\\Users\\d07321ow\\Documents\\hudonExplainableArtificialIntelligence2021 - Extracted Annotations (03032022, 210306).md;C\:\\Users\\d07321ow\\Zotero\\storage\\PXEMUNUT\\Hudon et al. - 2021 - Explainable Artificial Intelligence (XAI) How the.pdf}
}

@incollection{musenClinicalDecisionSupportSystems2021,
  title = {Clinical {{Decision-Support Systems}}},
  booktitle = {Biomedical {{Informatics}}: {{Computer Applications}} in {{Health Care}} and {{Biomedicine}}},
  author = {Musen, Mark A. and Middleton, Blackford and Greenes, Robert A.},
  editor = {Shortliffe, Edward H. and Cimino, James J.},
  date = {2021},
  pages = {795--840},
  publisher = {{Springer International Publishing}},
  location = {{Cham}},
  doi = {10.1007/978-3-030-58721-5_24},
  url = {https://doi.org/10.1007/978-3-030-58721-5_24},
  %%urldate = {2022-03-04},
  abstract = {This chapter discusses information technology that provides health-care workers and patients with situation-specific advice that can inform their decision making. The intricacies of the clinical environment, new legislative mandates, and the increasing complexity of medical practice all escalate the demand for clinical decision-support systems (CDSS) that can deliver tailored information at the right time and in the right context. The chapter describes methods for building CDSSs, which include context-specific information retrieval, grouping information within order sets, learning from data, and the use of declarative knowledge representations. The landscape for deploying CDSS technology is evolving rapidly, and the chapter discusses current standards and challenges for CDSS implementation. It concludes with a presentation of opportunities for future research.},
  isbn = {978-3-030-58721-5},
  langid = {english},
  keywords = {Belief network,Clinical decision support,Clinical Quality Language,Infobutton,Machine learning,Order set,Rule-based system,Situation-specific factors},
  file = {C\:\\Users\\d07321ow\\Zotero\\storage\\UPMFF8XI\\Musen et al_2021_Clinical Decision-Support Systems.pdf}
}

@article{suttonOverviewClinicalDecision2020,
  title = {An Overview of Clinical Decision Support Systems: Benefits, Risks, and Strategies for Success},
  shorttitle = {An Overview of Clinical Decision Support Systems},
  author = {Sutton, Reed T. and Pincock, David and Baumgart, Daniel C. and Sadowski, Daniel C. and Fedorak, Richard N. and Kroeker, Karen I.},
  date = {2020-12},
  journaltitle = {npj Digital Medicine},
  shortjournal = {npj Digit. Med.},
  volume = {3},
  number = {1},
  pages = {17},
  issn = {2398-6352},
  doi = {10.1038/s41746-020-0221-y},
  url = {http://www.nature.com/articles/s41746-020-0221-y},
  %%urldate = {2022-03-03},
  abstract = {Abstract             Computerized clinical decision support systems, or CDSS, represent a paradigm shift in healthcare today. CDSS are used to augment clinicians in their complex decision-making processes. Since their first use in the 1980s, CDSS have seen a rapid evolution. They are now commonly administered through electronic medical records and other computerized clinical workflows, which has been facilitated by increasing global adoption of electronic medical records with advanced capabilities. Despite these advances, there remain unknowns regarding the effect CDSS have on the providers who use them, patient outcomes, and costs. There have been numerous published examples in the past decade(s) of CDSS success stories, but notable setbacks have also shown us that CDSS are not without risks. In this paper, we provide a state-of-the-art overview on the use of clinical decision support systems in medicine, including the different types, current use cases with proven efficacy, common pitfalls, and potential harms. We conclude with evidence-based recommendations for minimizing risk in CDSS design, implementation, evaluation, and maintenance.},
  langid = {english},
  keywords = {archived,read},
  annotation = {https://web.archive.org/web/20220303222457/https://www.nature.com/articles/s41746-020-0221-y},
  file = {C\:\\Users\\d07321ow\\Documents\\suttonOverviewClinicalDecision2020 - Extracted Annotations (04032022, 005923).md;C\:\\Users\\d07321ow\\Zotero\\storage\\I6PAL6FY\\Sutton et al. - 2020 - An overview of clinical decision support systems .pdf}
}

@inproceedings{yangUnremarkableAIFitting2019,
  title = {Unremarkable {{AI}}: {{Fitting Intelligent Decision Support}} into {{Critical}}, {{Clinical Decision-Making Processes}}},
  shorttitle = {Unremarkable {{AI}}},
  booktitle = {Proceedings of the 2019 {{CHI Conference}} on {{Human Factors}} in {{Computing Systems}}},
  author = {Yang, Qian and Steinfeld, Aaron and Zimmerman, John},
  date = {2019-05-02},
  series = {{{CHI}} '19},
  pages = {1--11},
  publisher = {{Association for Computing Machinery}},
  location = {{New York, NY, USA}},
  doi = {10.1145/3290605.3300468},
  url = {https://doi.org/10.1145/3290605.3300468},
  %%urldate = {2022-03-03},
  abstract = {Clinical decision support tools (DST) promise improved healthcare outcomes by offering data-driven insights. While effective in lab settings, almost all DSTs have failed in practice. Empirical research diagnosed poor contextual fit as the cause. This paper describes the design and field evaluation of a radically new form of DST. It automatically generates slides for clinicians' decision meetings with subtly embedded machine prognostics. This design took inspiration from the notion of Unremarkable Computing, that by augmenting the users' routines technology/AI can have significant importance for the users yet remain unobtrusive. Our field evaluation suggests clinicians are more likely to encounter and embrace such a DST. Drawing on their responses, we discuss the importance and intricacies of finding the right level of unremarkableness in DST design, and share lessons learned in prototyping critical AI systems as a situated experience.},
  isbn = {978-1-4503-5970-2},
  keywords = {decision support systems,healthcare,read,user experience},
  file = {C\:\\Users\\d07321ow\\Documents\\yangUnremarkableAIFitting2019 - Extracted Annotations (03032022, 231452).md;C\:\\Users\\d07321ow\\Zotero\\storage\\KPDJVSCP\\Yang et al_2019_Unremarkable AI.pdf}
}

@book{raggiDissectingRepresentations2020,
  title = {Dissecting {{Representations}}},
  author = {Raggi, Daniel and Stockdill, Aaron and Jamnik, Mateja and Garcia Garcia, G. and Sutherland, H. E. A. and Cheng, P. C. H.},
  date = {2020-08-24},
  publisher = {{Springer International Publishing}},
  issn = {0302-9743},
  doi = {10.17863/CAM.57293; https://web.archive.org/web/20220314103546/https://www.repository.cam.ac.uk/handle/1810/310207},
  url = {https://www.repository.cam.ac.uk/handle/1810/310207},
  %%urldate = {2022-03-14},
  abstract = {Choosing effective representations for a problem and for the person solving it has benefits, including the ability or inability to solve it. We previously devised a novel framework consisting of a language to describe representations and computational methods to analyse them in terms of their formal and cognitive properties. In this paper we demonstrate the application of this framework to a variety of notations including natural languages, formal languages, and diagrams. We show how our framework, and the analysis of representations that it enables, gives us insight into how and why we can select representations which are appropriate for both the task and the user.},
  isbn = {978-3-030-54248-1},
  langid = {english},
  keywords = {archived},
  annotation = {Accepted: 2020-09-11T23:31:44Z},
  file = {C\:\\Users\\d07321ow\\Zotero\\storage\\NY432XX3\\Raggi et al_2020_Dissecting Representations.pdf;C\:\\Users\\d07321ow\\Zotero\\storage\\3GDV5SGZ\\310207.html}
}

@inproceedings{chengCognitivePropertiesRepresentations2021,
  title = {Cognitive {{Properties}} of {{Representations}}: {{A Framework}}},
  shorttitle = {Cognitive {{Properties}} of {{Representations}}},
  booktitle = {Diagrammatic {{Representation}} and {{Inference}}},
  author = {Cheng, Peter C.-H. and Garcia Garcia, Grecia and Raggi, Daniel and Stockdill, Aaron and Jamnik, Mateja},
  editor = {Basu, Amrita and Stapleton, Gem and Linker, Sven and Legg, Catherine and Manalo, Emmanuel and Viana, Petrucio},
  date = {2021},
  series = {Lecture {{Notes}} in {{Computer Science}}},
  pages = {415--430},
  publisher = {{Springer International Publishing}},
  location = {{Cham}},
  doi = {10.1007/978-3-030-86062-2_43},
  isbn = {978-3-030-86062-2},
  langid = {english}
}

@article{thayaparan2020survey,
  title={A survey on explainability in machine reading comprehension},
  author={Thayaparan, Mokanarangan and Valentino, Marco and Freitas, Andr{\'e}},
  journal={arXiv preprint arXiv:2010.00389},
  year={2020}
}

@article {Lee2020.11.30.20239095,
	author = {Lee, R.J. and Zhou, C. and Wysocki, O. and Shotton, R. and Tivey, A. and Lever, L. and Woodcock, J. and Angelakas, A. and Aung, T. and Banfill, K. and Baxter, M. and Bhogal, T. and Boyce, H. and Copson, E. and Dickens, E. and Eastlake, L. and Frost, H. and Gomes, F. and Graham, D.M and Hague, C. and Harrison, M. and Horsley, L. and Huddar, P. and Hudson, Z. and Khan, S. and Khan, U. T. and Maynard, A. and McKenzie, H. and Robinson, T. and Rowe, M. and Thomas, Anne and Turtle, Lance and Sheehan, R. and Stockdale, A. and Weaver, J. and Williams, S. and Wilson, C. and Hoskins, R. and Stevenson, J. and Fitzpatrick, P. and Palmieri, C. and Landers, D. and Cooksley, T and Dive, C. and Freitas, A. and Armstrong, A. C.},
	title = {Establishment of CORONET; COVID-19 Risk in Oncology Evaluation Tool to identify cancer patients at low versus high risk of severe complications of COVID-19 infection upon presentation to hospital},
	elocation-id = {2020.11.30.20239095},
	year = {2020},
	doi = {10.1101/2020.11.30.20239095},
	publisher = {Cold Spring Harbor Laboratory Press},
	abstract = {Background Cancer patients are at increased risk of severe COVID-19. As COVID-19 presentation and outcomes are heterogeneous in cancer patients, decision-making tools for hospital admission, severity prediction and increased monitoring for early intervention are critical.Objective To identify features of COVID-19 in cancer patients predicting severe disease and build a decision-support online tool; COVID-19 Risk in Oncology Evaluation Tool (CORONET)Method Data was obtained for consecutive patients with active cancer with laboratory confirmed COVID-19 presenting in 12 hospitals throughout the United Kingdom (UK). Univariable logistic regression was performed on pre-specified features to assess their association with admission (>=24 hours inpatient), oxygen requirement and death. Multivariable logistic regression and random forest models (RFM) were compared with patients randomly split into training and validation sets. Cost function determined cut-offs were defined for admission/death using RFM. Performance was assessed by sensitivity, specificity and Brier scores (BS). The CORONET model was then assessed in the entire cohort to build the online CORONET tool.Results Training and validation sets comprised 234 and 66 patients respectively with median age 69 (range 19-93), 54\% males, 46\% females, 71\% vs 29\% had solid and haematological cancers. The RFM, selected for further development, demonstrated superior performance over logistic regression with AUROC predicting admission (0.85 vs. 0.78) and death (0.76 vs. 0.72). C-reactive protein was the most important feature predicting COVID-19 severity. CORONET cut-offs for admission and mortality of 1.05 and 1.8 were established. In the training set, admission prediction sensitivity and specificity were 94.5\% and 44.3\% with BS 0.118; mortality sensitivity and specificity were 78.5\% and 57.2\% with BS 0.364. In the validation set, admission sensitivity and specificity were 90.7\% and 42.9\% with BS 0.148; mortality sensitivity and specificity were 92.3\% and 45.8\% with BS 0.442. In the entire cohort, the CORONET decision support tool recommended admission of 99\% of patients requiring oxygen and of 99\% of patients who died.Conclusions and Relevance CORONET, a decision support tool validated in hospitals throughout the UK showed promise in aiding decisions regarding admission and predicting COVID-19 severity in patients with cancer presenting to hospital. Future work will validate and refine the tool in further datasets.Competing Interest StatementR Lee speaker fees BMS and Astrazeneca, M Rowe honoraria from Astellas Pharma, speaker fees MSD and Servier. C. Wilson consultancy and speaker fees Pfizer, Amgen, Novartis, A Armstrong conference fee Merck, spouse shares in Astrazeneca. T Robinson financial support to attend educational workshops from Amgen and Daiichi-Sankyo. C Dive, outside of this scope of work, has received research funding from AstraZeneca, Astex Pharmaceuticals, Bioven, Amgen, Carrick Therapeutics, Merck AG, Taiho Oncology, Clearbridge Biomedics, Angle PLC, Menarini Diagnostics, GSK, Bayer, Boehringer Ingelheim, Roche, BMS, Novartis, Celgene, Thermofisher. C Dive is on advisory boards for, and has received consultancy fees/honoraria from, AstraZeneca, Biocartis and Merck KGaA.Funding StatementRebecca Lee and Tim Robinson are supported by the National Institute for Health Research as a Clinical Lecturer. Talvinder Bhogal is supported by the National Institute for Health Research as an academic clinical fellow. Umair Khan is an MRC Clinical Training Fellow based at the University of Liverpool supported by the North West England Medical Research Council Fellowship Scheme in Clinical Pharmacology and Therapeutics, which is funded by the Medical Research Council (Award Ref. MR/N025989/1). C Dive C5757/A27412), the CRUK Manchester Centre Award (C5759/A25254), and is supported by the NIHR Manchester Biomedical Research Centre. Funding for COVID-19 work has been provided by The Christie Charitable fund (1049751).Author DeclarationsI confirm all relevant ethical guidelines have been followed, and any necessary IRB and/or ethics committee approvals have been obtained.YesThe details of the IRB/oversight body that provided approval or exemption for the research described are given below:HRA and Health and Care Research Wales (HCRW) approval granted (reference 20/WA/0269)All necessary patient/participant consent has been obtained and the appropriate institutional forms have been archived.YesI understand that all clinical trials and any other prospective interventional studies must be registered with an ICMJE-approved registry, such as ClinicalTrials.gov. I confirm that any such study reported in the manuscript has been registered and the trial registration ID is provided (note: if posting a prospective study registered retrospectively, please provide a statement in the trial ID field explaining why the study was not registered in advance).YesI have followed all appropriate research reporting guidelines and uploaded the relevant EQUATOR Network research reporting checklist(s) and other pertinent material as supplementary files, if applicable.YesCode for the tool is available at Github (https://github.com/oskwys/CORONET). Raw data is available upon request to corresponding author, however may not include all details due to information governance regulations. https://github.com/oskwys/CORONET https://coronet.manchester.ac.uk/},
	URL = {https://www.medrxiv.org/content/early/2020/12/03/2020.11.30.20239095},
	%eprint = {https://www.medrxiv.org/content/early/2020/12/03/2020.11.30.20239095.full.pdf},
	journal = {medRxiv}
}

@article{coronetabstract,
author = {Lee, Rebecca and Wysocki, Oskar and Zhou, Cong and Calles, Antonio and Eastlake, Leonie and Ganatra, Sarju and Harrison, Michelle and Horsley, Laura and Huddar, Prerana and Khan, Khurum and Mckenzie, Hayley and Palmieri, Carlo and Rogado Revuelta, Jacobo and Thomas, Anne and Wilson, Caroline and Cooksley, Tim and Dive, Caroline and Freitas, Andre and Armstrong, Anne Caroline and CORONET Consortium},
title = {CORONET; COVID-19 in Oncology evaluatiON Tool: Use of machine learning to inform management of COVID-19 in patients with cancer.},
journal = {Journal of Clinical Oncology},
volume = {39},
number = {15\_suppl},
pages = {1505-1505},
year = {2021},
doi = {10.1200/JCO.2021.39.15\_suppl.1505},

URL = { 
        https://doi.org/10.1200/JCO.2021.39.15_suppl.1505
    
},
%eprint = { 
%        https://doi.org/10.1200/JCO.2021.39.15_suppl.1505
    
}
,
    abstract = { 1505Background: Patients (pts) with cancer are at increased risk of severe COVID-19 infection and death. Due to COVID-19 outcome heterogeneity, accurate assessment of pts is crucial. Early identification of pts who are likely to deteriorate allows timely discussions regarding escalation of care. Likewise, safe home management will reduce risk of nosocomial infection. To aid clinical decision-making, we developed a model to help determine which pts should be admitted vs. managed as an outpatient and which pts are likely to have severe COVID-19. Methods: Pts with active solid or haematological cancer presenting with symptoms/asymptomatic and testing positive for SARS-CoV-2 in Europe and USA were identified following institutional board approval. Clinical and laboratory data were extracted from pt records. Clinical outcome measures were discharge within 24 hours, requirement for oxygen at any stage during admission and death. Random Forest (RF) algorithm was used for model derivation as it compared favourably vs. lasso regression. Relevant clinical features were identified using recursive feature elimination based on SHAP. Internal validation (bootstrapping) with multiple imputations for missing data (maximum ≤2) were used for performance evaluation. Cost function determined cut-offs were defined for admission/death. The final CORONET model was trained on the entire cohort. Results: Model derivation set comprised 672 pts (393 male, 279 female, median age 71). 83\% had solid cancers, 17\% haematological. Predictive features were selected based on clinical relevance and data availability, supported by recursive feature elimination based on SHAP. RF model using haematological cancer, solid cancer stage, no of comorbidities, National Early Warning Score 2 (NEWS2), neutrophil:lymphocyte ratio, platelets, CRP and albumin achieved AUROC for admission 0.79 (+/-0.03) and death 0.75 (+/-0.02). RF explanation using SHAP revealed NEWS2 and C-reactive protein as the most important features predicting COVID-19 severity. In the entire cohort, CORONET recommended admission of 96\% of patients requiring oxygen and 99\% of patients who died. We then built a decision support tool using the model, which aids clinical decisions by presenting model predictions and explaining key contributing features. Conclusions: We have developed a model and tool available athttps://coronet.manchester.ac.uk/ to predict which pts with cancer and COVID-19 require hospital admission and are likely to have a severe disease course. CORONET is being continuously refined and validated over time. }
}

@article{LEE2021100005,
title = {Longitudinal characterisation of haematological and biochemical parameters in cancer patients prior to and during COVID-19 reveals features associated with outcome},
journal = {ESMO Open},
volume = {6},
number = {1},
pages = {100005},
year = {2021},
issn = {2059-7029},
doi = {https://doi.org/10.1016/j.esmoop.2020.100005},
url = {https://www.sciencedirect.com/science/article/pii/S2059702920328647},
author = {R.J. Lee and O. Wysocki and T. Bhogal and R. Shotton and A. Tivey and A. Angelakas and T. Aung and K. Banfill and M. Baxter and H. Boyce and G. Brearton and E. Copson and E. Dickens and L. Eastlake and F. Gomes and C. Hague and M. Harrison and L. Horsley and P. Huddar and Z. Hudson and S. Khan and U.T. Khan and A. Maynard and H. McKenzie and D. Palmer and T. Robinson and M. Rowe and A. Thomas and J. Tweedy and R. Sheehan and A. Stockdale and J. Weaver and S. Williams and C. Wilson and C. Zhou and C. Dive and T. Cooksley and C. Palmieri and A. Freitas and A.C. Armstrong}
}

@article{burkeBiomarkerIdentificationUsing2022,
  title = {Biomarker Identification Using Dynamic Time Warping Analysis: A Longitudinal Cohort Study of Patients with {{COVID-19}} in a {{UK}} Tertiary Hospital},
  shorttitle = {Biomarker Identification Using Dynamic Time Warping Analysis},
  author = {Burke, Hannah and Freeman, Anna and O'Regan, Paul and Wysocki, Oskar and Freitas, Andre and Dushianthan, Ahilanandan and Celinski, Michael and Batchelor, James and Phan, Hang and Borca, Florina and Sheard, Natasha and Williams, Sarah and Watson, Alastair and Fitzpatrick, Paul and Landers, D{\'o}nal and Wilkinson, Tom},
  year = {2022},
  month = feb,
  journal = {BMJ Open},
  volume = {12},
  number = {2},
  pages = {e050331},
  issn = {2044-6055, 2044-6055},
  doi = {10.1136/bmjopen-2021-050331},
  langid = {english},
  
 }

@article{freemanWaveComparisonsClinical2022,
  title = {Wave Comparisons of Clinical Characteristics and Outcomes of {{COVID-19}} Admissions - {{Exploring}} the Impact of Treatment and Strain Dynamics},
  author = {Freeman, Anna and Watson, Alastair and O'Regan, Paul and Wysocki, Oskar and Burke, Hannah and Freitas, Andre and Livingstone, Robert and Dushianthan, Ahilanadan and Celinski, Michael and Batchelor, James and Phan, Hang and Borca, Florina and Fitzpatrick, Paul and Landers, Donal and Wilkinson, Tom MA},
  year = {2022},
  month = jan,
  journal = {Journal of Clinical Virology},
  volume = {146},
  pages = {105031},
  issn = {13866532},
  doi = {10.1016/j.jcv.2021.105031},
  langid = {english},
  
  }

@article{wysockiInternationalComparisonPresentation2022a,
  title = {An {{International Comparison}} of {{Presentation}}, {{Outcomes}} and {{CORONET Predictive Score Performance}} in {{Patients}} with {{Cancer Presenting}} with {{COVID-19}} across {{Different Pandemic Waves}}},
  author = {Wysocki, Oskar and Zhou, Cong and Rogado, Jacobo and Huddar, Prerana and Shotton, Rohan and Tivey, Ann and Albiges, Laurence and Angelakas, Angelos and Arnold, Dirk and Aung, Theingi and Banfill, Kathryn and Baxter, Mark and Barlesi, Fabrice and Bayle, Arnaud and Besse, Benjamin and Bhogal, Talvinder and Boyce, Hayley and Britton, Fiona and Calles, Antonio and {Castelo-Branco}, Luis and Copson, Ellen and Croitoru, Adina and Dani, Sourbha S. and Dickens, Elena and Eastlake, Leonie and Fitzpatrick, Paul and Foulon, Stephanie and Frederiksen, Henrik and Ganatra, Sarju and Gennatas, Spyridon and Glenth{\o}j, Andreas and Gomes, Fabio and Graham, Donna M. and Hague, Christina and Harrington, Kevin and Harrison, Michelle and Horsley, Laura and Hoskins, Richard and Hudson, Zoe and Jakobsen, Lasse H. and {Joharatnam-Hogan}, Nalinie and Khan, Sam and Khan, Umair T. and Khan, Khurum and Lewis, Alexandra and Massard, Christophe and Maynard, Alec and McKenzie, Hayley and Michielin, Olivier and Mosenthal, Anne C. and Obispo, Berta and Palmieri, Carlo and Patel, Rushin and Pentheroudakis, George and Peters, Solange and {Rieger-Christ}, Kimberly and Robinson, Timothy and Romano, Emanuela and Rowe, Michael and Sekacheva, Marina and Sheehan, Roseleen and Stockdale, Alexander and Thomas, Anne and Turtle, Lance and Vi{\~n}al, David and Weaver, Jamie and Williams, Sophie and Wilson, Caroline and Dive, Caroline and Landers, Donal and Cooksley, Timothy and Freitas, Andr{\'e} and Armstrong, Anne C. and Lee, Rebecca J. and {on behalf of the ESMO Co-Care}},
  year = {2022},
  month = aug,
  journal = {Cancers},
  volume = {14},
  number = {16},
  pages = {3931},
  issn = {2072-6694},
  doi = {10.3390/cancers14163931},

  
}

@article{coronet_paper,

    title = {Establishment of CORONET, COVID-19 Risk in Oncology Evaluation Tool, to Identify Patients With Cancer at Low Versus High Risk of Severe Complications of COVID-19 Disease On Presentation to Hospital},
    author = {Lee, Rebecca J. and Wysocki, Oskar and Zhou, Cong and Shotton, Rohan and Tivey, Ann and Lever, Louise and Woodcock, Joshua and Albiges, Laurence and Angelakas, Angelos and Arnold, Dirk and Aung, Theingi and Banfill, Kathryn and Baxter, Mark and Barlesi, Fabrice and Bayle, Arnaud and Besse, Benjamin and Bhogal, Talvinder and Boyce, Hayley and Britton, Fiona and Calles, Antonio and Castelo-Branco, Luis and Copson, Ellen and Croitoru, Adina E. and Dani, Sourbha S. and Dickens, Elena and Eastlake, Leonie and Fitzpatrick, Paul and Foulon, Stephanie and Frederiksen, Henrik and Frost, Hannah and Ganatra, Sarju and Gennatas, Spyridon and Glenthoj, Andreas and Gomes, Fabio and Graham, Donna M. and Hague, Christina and Harrington, Kevin and Harrison, Michelle and Horsley, Laura and Hoskins, Richard and Huddar, Prerana and Hudson, Zoe and Jakobsen, Lasse H. and Joharatnam-Hogan, Nalinie and Khan, Sam and Khan, Umair T. and Khan, Khurum and Massard, Christophe and Maynard, Alec and McKenzie, Hayley and Michielin, Olivier and Mosenthal, Anne C. and Obispo, Berta and Patel, Rushin and Pentheroudakis, George and Peters, Solange and Rieger-Christ, Kimberly and Robinson, Timothy and Rogado, Jacobo and Romano, Emanuela and Rowe, Michael and Sekacheva, Marina and Sheehan, Roseleen and Stevenson, Julie and Stockdale, Alexander and Thomas, Anne and Turtle, Lance and Vinal, David and Weaver, Jamie and Williams, Sophie and Wilson, Caroline and Palmieri, Carlo and Landers, Donal and Cooksley, Timothy and Dive, Caroline and Freitas, Andre and Armstrong, Anne C.},
    
    journal = {JCO Clinical Cancer Informatics},
    volume = {},
    number = {6},
    pages = {e2100177},
    year = {2022},
    doi = {10.1200/CCI.21.00177},
        note ={PMID: 35609228},
    
    %URL = { https://doi.org/10.1200/CCI.21.00177},
    %eprint = { https://doi.org/10.1200/CCI.21.00177}
    }

@article{evansExplainabilityParadoxChallenges2022,
  title = {The Explainability Paradox: {{Challenges}} for {{xAI}} in Digital Pathology},
  shorttitle = {The Explainability Paradox},
  author = {Evans, Theodore and Retzlaff, Carl Orge and Gei{\ss}ler, Christian and Kargl, Michaela and Plass, Markus and M{\"u}ller, Heimo and Kiehl, Tim-Rasmus and Zerbe, Norman and Holzinger, Andreas},
  year = {2022},
  month = aug,
  journal = {Future Generation Computer Systems},
  volume = {133},
  pages = {281--296},
  issn = {0167739X},
  doi = {10.1016/j.future.2022.03.009},
  langid = {english}
}

@article{holzingerHumanAIInterfaces2021,
  title = {Toward {{Human}}\textendash{{AI Interfaces}} to {{Support Explainability}} and {{Causability}} in {{Medical AI}}},
  author = {Holzinger, Andreas and M{\"u}ller, Heimo},
  year = {2021},
  month = oct,
  journal = {Computer},
  volume = {54},
  number = {10},
  pages = {78--86},
  issn = {1558-0814},
  doi = {10.1109/MC.2021.3092610}
}

@article{holzingerInformationFusionIntegrative2022,
  title = {Information Fusion as an Integrative Cross-Cutting Enabler to Achieve Robust, Explainable, and Trustworthy Medical Artificial Intelligence},
  author = {Holzinger, Andreas and Dehmer, Matthias and {Emmert-Streib}, Frank and Cucchiara, Rita and Augenstein, Isabelle and Ser, Javier Del and Samek, Wojciech and Jurisica, Igor and {D{\'i}az-Rodr{\'i}guez}, Natalia},
  year = {2022},
  month = mar,
  journal = {Information Fusion},
  volume = {79},
  pages = {263--278},
  issn = {15662535},
  doi = {10.1016/j.inffus.2021.10.007}
}

@article{mullerExplainabilityCausabilityArtificial2022,
  title = {Explainability and Causability for Artificial Intelligence-Supported Medical Image Analysis in the Context of the {{European In Vitro Diagnostic Regulation}}},
  author = {M{\"u}ller, Heimo and Holzinger, Andreas and Plass, Markus and Brcic, Luka and Stumptner, Cornelia and Zatloukal, Kurt},
  year = {2022},
  month = sep,
  journal = {New Biotechnology},
  volume = {70},
  pages = {67--72},
  issn = {18716784},
  doi = {10.1016/j.nbt.2022.05.002},
  langid = {english}
}

@misc{apleyVisualizingEffectsPredictor2019,
  title = {Visualizing the {{Effects}} of {{Predictor Variables}} in {{Black Box Supervised Learning Models}}},
  author = {Apley, Daniel W. and Zhu, Jingyu},
  year = {2019},
  month = aug,
  number = {arXiv:1612.08468},
  eprint = {1612.08468},
  eprinttype = {arxiv},
  primaryclass = {stat},
  publisher = {{arXiv}},
  archiveprefix = {arXiv}
}

@misc{dhurandharExplanationsBasedMissing2018,
  title = {Explanations Based on the {{Missing}}: {{Towards Contrastive Explanations}} with {{Pertinent Negatives}}},
  shorttitle = {Explanations Based on the {{Missing}}},
  author = {Dhurandhar, Amit and Chen, Pin-Yu and Luss, Ronny and Tu, Chun-Chen and Ting, Paishun and Shanmugam, Karthikeyan and Das, Payel},
  year = {2018},
  month = oct,
  number = {arXiv:1802.07623},
  eprint = {1802.07623},
  eprinttype = {arxiv},
  primaryclass = {cs},
  publisher = {{arXiv}},

}

@misc{hanawaEvaluationSimilaritybasedExplanations2021,
  title = {Evaluation of {{Similarity-based Explanations}}},
  author = {Hanawa, Kazuaki and Yokoi, Sho and Hara, Satoshi and Inui, Kentaro},
  year = {2021},
  month = mar,
  number = {arXiv:2006.04528},
  eprint = {2006.04528},
  eprinttype = {arxiv},
  primaryclass = {cs, stat},
  publisher = {{arXiv}},
  archiveprefix = {arXiv},

}

@inproceedings{haraMakingTreeEnsembles2018,
  title = {Making {{Tree Ensembles Interpretable}}: {{A Bayesian Model Selection Approach}}},
  shorttitle = {Making {{Tree Ensembles Interpretable}}},
  booktitle = {Proceedings of the {{Twenty-First International Conference}} on {{Artificial Intelligence}} and {{Statistics}}},
  author = {Hara, Satoshi and Hayashi, Kohei},
  year = {2018},
  month = mar,
  pages = {77--85},
  publisher = {{PMLR}},
  issn = {2640-3498},
  langid = {english},

}

@misc{vanlooverenInterpretableCounterfactualExplanations2020,
  title = {Interpretable {{Counterfactual Explanations Guided}} by {{Prototypes}}},
  author = {Van Looveren, Arnaud and Klaise, Janis},
  year = {2020},
  month = feb,
  number = {arXiv:1907.02584},
  eprint = {1907.02584},
  eprinttype = {arxiv},
  primaryclass = {cs, stat},
  publisher = {{arXiv}},
  archiveprefix = {arXiv},

}

@inproceedings{jiangTrustNotTrust2018a,
  title = {To Trust or Not to Trust a Classifier},
  booktitle = {Proceedings of the 32nd {{International Conference}} on {{Neural Information Processing Systems}}},
  author = {Jiang, Heinrich and Kim, Been and Guan, Melody Y. and Gupta, Maya},
  year = {2018},
  month = dec,
  series = {{{NIPS}}'18},
  pages = {5546--5557},
  publisher = {{Curran Associates Inc.}},
  address = {{Red Hook, NY, USA}},

}

@article{friedman2001greedy,
  title={Greedy function approximation: a gradient boosting machine},
  author={Friedman, Jerome H},
  journal={Annals of statistics},
  pages={1189--1232},
  year={2001},
  publisher={JSTOR}
}

@article{goldsteinPeekingBlackBox2015,
  title = {Peeking {{Inside}} the {{Black Box}}: {{Visualizing Statistical Learning With Plots}} of {{Individual Conditional Expectation}}},
  shorttitle = {Peeking {{Inside}} the {{Black Box}}},
  author = {Goldstein, Alex and Kapelner, Adam and Bleich, Justin and Pitkin, Emil},
  year = {2015},
  month = jan,
  journal = {Journal of Computational and Graphical Statistics},
  volume = {24},
  number = {1},
  pages = {44--65},
  issn = {1061-8600, 1537-2715},
  doi = {10.1080/10618600.2014.907095},
  langid = {english},
}
\newpage

\clearpage

\setcounter{secnumdepth}{0}
\section{Supplementary Information}
\setcounter{secnumdepth}{0}
\label{sec:SuppMat}

\setcounter{table}{0}
\renewcommand\thetable{S.\arabic{table}}
\setcounter{figure}{0}
\renewcommand\thefigure{S.\arabic{figure}}

\subsection{Supplementary Methods}

\subsubsection{List of questions}
\label{supp_list_of_questions}

Below we list all questions from the experiment, that are relevant in the analysis aiming to answer broader research questions from the previous section. The questions were produced by both computer scientists and oncologists, same as involved in designing clinical cases.

Questions related to demographics:
\begin{itemize}
    \item D1. What is your job title?
    
\item D2. What is your age group?
\item D3. Which of the following tasks do you perform on computers at work?
\item D4. Rate your knowledge on the management of patients with cancer who have developed COVID-19
\end{itemize}

Questions related to expectations:

\begin{itemize}
\item Q1. I feel comfortable when using new technology
\item Q2. It is important for me to know the mathematics behind the model's recommendations
\item Q3. It is important for me to know the features of my patient contribute to the model's recommendation
\item Q4. It is important for me to know how the model makes its recommendation for my individual patient
\item Q5. It is important for me to know how uncertain (in \%) the model is about its recommendation
\end{itemize}

Questions related to the visual representation:
\begin{itemize}
\item Q6. The colour bar with the score is easy to interpret
\item Q7. The colour bar with the score convinces me to accept or reject the model's recommendation
\item Q8. The scatterplot with all patients is easy to interpret
\item Q9. The scatterplot with all patients convinces me to accept or reject the model's recommendation
\item Q10. The barplot with feature contribution is easy to interpret
\item Q11. The barplot with feature contribution convinces me to accept or reject the model's recommendation  
\end{itemize}

Questions related to the user's attitude towards the model:

\begin{itemize}
\item A.  I am satisfied with the output information that CORONET provides
\item B. CORONET helps me in making safe clinical decisions on patient management   
\item C. When my initial decision was the same as CORONET had recommended, I felt reassured    
\item D.  I understand when and why CORONET may provide the wrong recommendation in some cases    
\item E. CORONET helps in cases where I am less confident in the decision on how to proceed    
\item F. Even when my initial course of action was different to what CORONET recommended, I still had full confidence in my original decision.    
\item G. I was surprised when CORONET recommended an action different to my own
\end{itemize}

\subsubsection{Feedback from Healthcare Professionals}
\label{feedback_from_HCPs}
Below we quote received feedback regarding the usability of the tool from the Healthcare Professionals:

\begin{itemize}
\item \textit{This could be a very useful tool  to aid the clinician's decision. With every risk stratification score/tool - I'm used to using the MASCC score for febrile neutropenia, it helps your final decision. If my gut said admit, irrespective of any score, I would admit. Tools can be great to add confidence to a less obvious choice of admit/discharge.
\item With all the patients I would have discharged home, I still would have asked my team to safety-net call for the next 10-14 days.
\item I don’t think this can be proposed in clinical practice 
\item decision to admit a cancer patient is complex and multifactorial - often dependent on social support, patient engagement and both cancer and treatment related complications
\item Really good tool. Very easy to use. It would be great to get individual hospital trusts to consider adding it to hospital guidelines so doctors felt safer using it to base decisions off.
%\item I couldn't seem to use it on Chrome on PC - I couldn't click on the buttons for e.g. number of comorbidities (this was possibly an issue with poor internet in my trust).
\item Easy to use but differed a lot with my clinical assessment 
\item Having played with the app a little, it seems like the biggest impact on score comes from NEWS, so I think for many patients it won't add much, but can be a useful adjunct.
\item An interesting concept but not one that I think would be used in its current format in clinical care-too formulaic and not individualized to patient risk factors}
\end{itemize}

Additionally, at the end of the experiment Healthcare Professionals were asked about aspects in which the tool could be improved in the future. Below we quote received feedback:

\textit{\begin{itemize}
\item Inclusion in guidelines, clarify the legal implications 
\item CORONET is a standalone `App' - to be useful (and used) it has to integrate with patient EPRs (Electronic Patient Record) so HCPs are not copying electronic information from EPR to CORONET - this not only takes time but prone to errors - This would be a major factor in my decision to recommend CORONET
\item May be useful to separate high risk and low risk comorbidities. For example, some conditions such as COPD will cause more severe issues when combined with COVID than other conditions such as diabetes. 
\item It does not seem to take into account oxygen requirements independently of NEWS2; it is difficult to discharge someone who is needing 4L of oxygen!
\item Take into consideration if they are active chemotherapy/immunotherapy. As even with a covid positive test, they still require iv antibiotics and monitoring. Higher risk of deterioration, especially lung cancers, if discharged home. 
\item The wording on the red/yellow diagram seems to be inconsistent re categories. 1st and 2nd are actions e.g. `consider' admit/discharge, the 3rd is commenting on the disease severity. These are different categories, so the third should be more like `consider level 2+ care' or another type of action 
\item The labeling of the red/yellow score is a bit hard to follow, suggest removing the bottom legend
\end{itemize}}

\subsubsection{User interface design}
\label{user_interface_development}
The CORONET tool available at \url{https://coronet.manchester.ac.uk/} was developed following general principles of usability and universal design in order to meet level 2 (AA) of the Web Content Accessibility Guidelines (WCAG) 2.1 (\url{https://www.w3.org/TR/WCAG21/}). More details regarding the web accessibility are available at \url{https://www.manchester.ac.uk/web-accessibility/}.

\newpage
\subsection{Supplementary Tables}

% ----------------------------------------------------------------------------------------------------------------------------------------------------
% TASKS PERFORMED BY HCPS AT WORK

\begin{table}[htb!]
\centering
\caption{Tasks performed on computers at work by the healthcare professionals participating in the experiment.}
\label{tab:tasks_performed}
\resizebox{.99\textwidth}{!}{
\begin{tabular}{lcccccccccccccccccccccccc} 
\toprule
                                           & \multicolumn{1}{l}{\begin{sideways}Advanced Nurse Practitioner or
  higher\end{sideways}} & \multicolumn{1}{l}{\begin{sideways}Advanced Nurse Practitioner or
  higher\end{sideways}} & \multicolumn{1}{l}{\begin{sideways}Consultant / attending physician\end{sideways}} & \multicolumn{1}{l}{\begin{sideways}Consultant / attending physician\end{sideways}} & \multicolumn{1}{l}{\begin{sideways}Consultant / attending physician\end{sideways}} & \multicolumn{1}{l}{\begin{sideways}Consultant / attending physician\end{sideways}} & \multicolumn{1}{l}{\begin{sideways}Consultant / attending
  physician\end{sideways}} & \multicolumn{1}{l}{\begin{sideways}Doctor within first year of
  graduating (FY1) / intern\end{sideways}} & \multicolumn{1}{l}{\begin{sideways}Doctor (within 2-4 years of
  graduating / specialty trainee (ST) / fellow\end{sideways}} & \multicolumn{1}{l}{\begin{sideways}Doctor (within 2-4 years of
  graduating / specialty trainee (ST) / fellow\end{sideways}} & \multicolumn{1}{l}{\begin{sideways}General practitioner / GPST /
  community doctor\end{sideways}} & \multicolumn{1}{l}{\begin{sideways}General practitioner / GPST /
  community doctor\end{sideways}} & \multicolumn{1}{l}{\begin{sideways}General practitioner / GPST /
  community doctor\end{sideways}} & \multicolumn{1}{l}{\begin{sideways}Other (please enter text)\end{sideways}} & \multicolumn{1}{l}{\begin{sideways}Registered Nurse\end{sideways}} & \multicolumn{1}{l}{\begin{sideways}Specialist Nurse\end{sideways}} & \multicolumn{1}{l}{\begin{sideways}Specialist registrar / ST 3+ /
  senior resident/senior fellow\end{sideways}} & \multicolumn{1}{l}{\begin{sideways}Specialist registrar / ST 3+ /
  senior resident/senior fellow\end{sideways}} & \multicolumn{1}{l}{\begin{sideways}Specialist registrar / ST 3+ /
  senior resident/senior fellow\end{sideways}} & \multicolumn{1}{l}{\begin{sideways}Specialist registrar / ST 3+ /
  senior resident/senior fellow\end{sideways}} & \multicolumn{1}{l}{\begin{sideways}Specialist registrar / ST 3+ /
  senior resident/senior fellow\end{sideways}} & \multicolumn{1}{l}{\begin{sideways}Specialist registrar / ST 3+ /
  senior resident/senior fellow\end{sideways}} & \multicolumn{1}{l}{\begin{sideways}Specialist registrar / ST 3+ /
  senior resident/senior fellow\end{sideways}} & \multicolumn{1}{l}{\begin{sideways}\end{sideways}}  \\
Emailing                                   & x                                                                                         & x                                                                                         & x                                                                                  & x                                                                                  & x                                                                                  & x                                                                                  & x                                                                                    & x                                                                                                         & x                                                                                                                            & x                                                                                                                            &                                                                                                    & x                                                                                                  & x                                                                                                  & x                                                                           & x                                                                  & x                                                                  & x                                                                                                                & x                                                                                                                & x                                                                                                                & x                                                                                                                & x                                                                                                                & x                                                                                                                & x                                                                                                                & 96\%                                                \\
Accessing
  clinical guidelines            & x                                                                                         & x                                                                                         & x                                                                                  & x                                                                                  & x                                                                                  & x                                                                                  & x                                                                                    & x                                                                                                         & x                                                                                                                            & x                                                                                                                            &                                                                                                    & x                                                                                                  & x                                                                                                  & x                                                                           &                                                                    & x                                                                  & x                                                                                                                & x                                                                                                                & x                                                                                                                & x                                                                                                                & x                                                                                                                & x                                                                                                                & x                                                                                                                & 91\%                                                \\
Viewing
  test results                     & x                                                                                         & x                                                                                         & x                                                                                  & x                                                                                  & x                                                                                  & x                                                                                  & x                                                                                    & x                                                                                                         & x                                                                                                                            & x                                                                                                                            & x                                                                                                  & x                                                                                                  & x                                                                                                  & x                                                                           &                                                                    & x                                                                  & x                                                                                                                & x                                                                                                                & x                                                                                                                & x                                                                                                                & x                                                                                                                & x                                                                                                                & x                                                                                                                & 96\%                                                \\
Viewing
  drug charts                      & x                                                                                         & x                                                                                         & x                                                                                  & x                                                                                  & x                                                                                  & x                                                                                  & x                                                                                    & x                                                                                                         &                                                                                                                              & x                                                                                                                            &                                                                                                    &                                                                                                    & x                                                                                                  & x                                                                           &                                                                    & x                                                                  &                                                                                                                  & x                                                                                                                &                                                                                                                  & x                                                                                                                & x                                                                                                                & x                                                                                                                & x                                                                                                                & 74\%                                                \\
Online
  clinical calculators              &                                                                                           & x                                                                                         & x                                                                                  & x                                                                                  &                                                                                    &                                                                                    & x                                                                                    & x                                                                                                         & x                                                                                                                            & x                                                                                                                            & x                                                                                                  & x                                                                                                  & x                                                                                                  & x                                                                           & x                                                                  &                                                                    & x                                                                                                                & x                                                                                                                & x                                                                                                                & x                                                                                                                & x                                                                                                                & x                                                                                                                & x                                                                                                                & 83\%                                                \\
Online
  scoring systems and/or algorithms & x                                                                                         & x                                                                                         & x                                                                                  & x                                                                                  &                                                                                    &                                                                                    & x                                                                                    & x                                                                                                         &                                                                                                                              & x                                                                                                                            &                                                                                                    & x                                                                                                  & x                                                                                                  &                                                                             &                                                                    &                                                                    & x                                                                                                                & x                                                                                                                & x                                                                                                                & x                                                                                                                &                                                                                                                  & x                                                                                                                & x                                                                                                                & 65\%                                                \\
Statistical
  analysis                     & x                                                                                         &                                                                                           & x                                                                                  &                                                                                    &                                                                                    &                                                                                    & x                                                                                    &                                                                                                           &                                                                                                                              &                                                                                                                              &                                                                                                    &                                                                                                    &                                                                                                    &                                                                             &                                                                    & x                                                                  & x                                                                                                                &                                                                                                                  &                                                                                                                  &                                                                                                                  & x                                                                                                                & x                                                                                                                & x                                                                                                                & 35\%                                                \\
\bottomrule
\end{tabular}
}
\end{table}

% -------------------------------------------------------------------------------------------------------------------------------
% EASE OF INTERPRETATION MEDIANS

\begin{longtable}[c]{lc}
\caption{Responses related to ease of interpretation and convincing capacity of the tool’s output. No significant difference between answers.}
\label{tab:visual_stats}\\
\hline
 &
  \begin{tabular}[c]{@{}c@{}}median respons\\  in Likert scale 1-7 {[}Q1,Q3{]}\end{tabular} \\ \hline
\endfirsthead
\multicolumn{2}{c}%
{{\bfseries Table \thetable\ continued from previous page}} \\
\hline
 &
  \begin{tabular}[c]{@{}c@{}}median respons\\  in Likert scale 1-7 {[}Q1,Q3{]}\end{tabular} \\ \hline
\endhead
\begin{tabular}[c]{@{}l@{}}Q6. The colour bar with the score\\ is easy to interpret\end{tabular} &
  6 {[}5,6{]} \\
\begin{tabular}[c]{@{}l@{}}Q7. The colour bar with the score\\ convinces me to accept or reject\\ the model's   recommendation\end{tabular}           & 4 {[}4,6{]}   \\
\begin{tabular}[c]{@{}l@{}}Q8. The scatterplot with all patients\\ is easy to interpret\end{tabular} &
  5 {[}3.5,6{]} \\
\begin{tabular}[c]{@{}l@{}}Q9. The scatterplot with all patients\\ convinces me to accept or reject\\ the model's recommendation\end{tabular}       & 5 {[}4,5.5{]} \\
\begin{tabular}[c]{@{}l@{}}Q10. The barplot with feature contribution\\  is easy to interpret\end{tabular} &
  5 {[}5,6{]} \\
\begin{tabular}[c]{@{}l@{}}Q11. The barplot with feature contribution\\ convinces me to accept or reject\\  the model's recommendation\end{tabular} & 5 {[}4,6{]}   \\ \hline
\multicolumn{1}{c}{} &
  p\textgreater{}0.05 \\
 &
  Friedman test
\end{longtable}

% -------------------------------------------------------------------------------------------------------------------------------
% EASE OF INTERPRETATION

\begin{table}[hbt!]
\centering
\caption{Individual comparisons between reported ease of interpretation of the visuals and their convincing capacity. No significant differences; all p$>$0.05}
\label{tab:visual_stats_pairwise}
\resizebox{.8\textwidth}{!}{
\begin{tabular}{cccc} 
\toprule
                                                                                                                                 &     &                                                                                                                                  & p (Kruskal–Wallis)  \\
\begin{tabular}[c]{@{}c@{}}Q6. The diagram (colorbar)\\is easy to interpret\end{tabular}                                         & vs. & \begin{tabular}[c]{@{}c@{}}Q8. The diagram (dotplot)\\is easy to interpret\end{tabular}                                          & 0.131               \\
\begin{tabular}[c]{@{}c@{}}Q10. The diagram (barplot)\\is easy to interpret\end{tabular}                                         & vs. & \begin{tabular}[c]{@{}c@{}}Q6. The diagram (colour bar)\\is easy to interpret\end{tabular}                                         & 0.716               \\
\begin{tabular}[c]{@{}c@{}}Q10. The diagram (barplot)\\is easy to interpret\end{tabular}                                         & vs. & \begin{tabular}[c]{@{}c@{}}Q8. The diagram (dotplot)\\is easy to interpret\end{tabular}                                          & 0.244               \\
\begin{tabular}[c]{@{}c@{}}Q11. The diagram (barplot)\\convinces me to accept or reject\\the model's recommendation\end{tabular} & vs. & \begin{tabular}[c]{@{}c@{}}Q7. The diagram (colour bar)\\convinces me to accept or reject\\the model's recommendation\end{tabular} & 0.387               \\
\begin{tabular}[c]{@{}c@{}}Q11. The diagram (barplot)\\convinces me to accept or reject\\the model's recommendation\end{tabular} & vs. & \begin{tabular}[c]{@{}c@{}}Q9. The diagram (dotplot)\\convinces me to accept or reject\\the model's recommendation\end{tabular}  & 0.281               \\
\begin{tabular}[c]{@{}c@{}}Q7. The diagram (colour bar)\\convinces me to accept or reject\\the model's recommendation\end{tabular} & vs. & \begin{tabular}[c]{@{}c@{}}Q9. The diagram (dotplot)\\convinces me to accept or reject\\the model's recommendation\end{tabular}  & 0.740               \\
\bottomrule
\end{tabular}
}
\end{table}

% -------------------------------------------------------------------------------------------------------------------------------
% SPEARMAN INTERPRETABILITY OF VISUALS

\begin{table}[hbt!]
\centering
\caption{Spearman correlations between questions regarding ease of interpretation of the visuals or their convincing capacity (Q6-Q11) and questions regarding user competence in the task (D4) or expectations regarding the output (Q1-Q5). No significant correlations were found (Holm adjusted p\textgreater{}0.05).
}
\label{tab:visual_stats_corrs}
\resizebox{\textwidth}{!}{
\begin{tabular}{lcccccc} 
\toprule
Spearman correlation r                                                                                                                             & \multicolumn{1}{l}{\begin{tabular}[c]{@{}l@{}}D4. Rate your knowledge\\on the management of patients\\with cancer who have developed\\COVID-19\end{tabular}} & \multicolumn{1}{l}{\begin{tabular}[c]{@{}l@{}}Q1. I feel comfortable\\when using new technology\end{tabular}} & \multicolumn{1}{l}{\begin{tabular}[c]{@{}l@{}}Q2. It is important for me\\to know the mathematics\\behind the model's\\recommendations\end{tabular}} & \multicolumn{1}{l}{\begin{tabular}[c]{@{}l@{}}Q3. It is important for me\\to know the features\\of my patient contribute\\to the model's\\recommendation\end{tabular}} & \multicolumn{1}{l}{\begin{tabular}[c]{@{}l@{}}Q4. It is important for me\\to know how the model\\makes its recommendation\\for my individual patient\end{tabular}} & \multicolumn{1}{l}{\begin{tabular}[c]{@{}l@{}}Q5. It is important for me\\to know how uncertain (in \%)\\the model is about\\its recommendation\end{tabular}}  \\
\begin{tabular}[c]{@{}l@{}}Q6. The colour bar with the score\\is easy to interpret\end{tabular}                                                      & 0.026                                                                                                                                                        & -0.101                                                                                                        & -0.199                                                                                                                                               & -0.109                                                                                                                                                                 & -0.003                                                                                                                                                             & -0.187                                                                                                                                                         \\
\begin{tabular}[c]{@{}l@{}}Q7. The colour bar with the score\\convinces me to accept or reject\\the model's recommendation\end{tabular}              & -0.214                                                                                                                                                       & 0.062                                                                                                         & -0.195                                                                                                                                               & 0.031                                                                                                                                                                  & 0                                                                                                                                                                  & -0.011                                                                                                                                                         \\
\begin{tabular}[c]{@{}l@{}}Q8. The scatterplot with all patients\\is easy to interpret\end{tabular}                                                & -0.105                                                                                                                                                       & 0.333                                                                                                         & -0.122                                                                                                                                               & -0.309                                                                                                                                                                 & 0.085                                                                                                                                                              & -0.107                                                                                                                                                         \\
\begin{tabular}[c]{@{}l@{}}Q9. The scatterplot with all patients\\convinces me to accept or reject\\the model's recommendation\end{tabular}        & -0.156                                                                                                                                                       & 0.433                                                                                                         & -0.054                                                                                                                                               & -0.283                                                                                                                                                                 & 0.059                                                                                                                                                              & -0.167                                                                                                                                                         \\
\begin{tabular}[c]{@{}l@{}}Q10. The barplot with feature contribution\\is easy to interpret\end{tabular}                                           & -0.368                                                                                                                                                       & 0.077                                                                                                         & -0.288                                                                                                                                               & -0.06                                                                                                                                                                  & -0.164                                                                                                                                                             & -0.044                                                                                                                                                         \\
\begin{tabular}[c]{@{}l@{}}Q11. The barplot with feature contribution\\convinces me to accept or reject\\the model's recommendation~~\end{tabular} & -0.255                                                                                                                                                       & 0.388                                                                                                         & -0.268                                                                                                                                               & -0.124                                                                                                                                                                 & -0.038                                                                                                                                                             & -0.081                                                                                                                                                         \\
\bottomrule
\end{tabular}
}
\end{table}

% -------------------------------------------------------------------------------------------------------------------------------
% EXPECTATIONS

\begin{longtable}[c]{@{}ccc@{}}
\caption{Responses to questions Q2-Q5 related to user expectation regarding the output of the recommendation tool. Answers to Q2 are significantly lower than to other questions. No significant difference between Q3,Q4,Q5.}
\label{tab:expectation_stats}\\
\toprule
\multicolumn{1}{l}{} &
  \begin{tabular}[c]{@{}c@{}}median response\\ in Likert scale 1-7 {[}Q1,Q3{]}\end{tabular} &
  \multicolumn{1}{l}{} \\* \midrule
\endfirsthead
\multicolumn{3}{c}%
{{\bfseries Table \thetable\ continued from previous page}} \\
\toprule
\multicolumn{1}{l}{} &
  \begin{tabular}[c]{@{}c@{}}median response\\ in Likert scale 1-7 {[}Q1,Q3{]}\end{tabular} &
  \multicolumn{1}{l}{} \\* \midrule
\endhead
\bottomrule
\endfoot
\endlastfoot
\begin{tabular}[c]{@{}c@{}}Q2. It is important for me to know\\ the mathematics behind the model's recommendations\end{tabular} &
  4 {[}2,5{]} &
  b \\
\begin{tabular}[c]{@{}c@{}}Q3. It is important for me to know\\ the features of my patient contribute\\  to the model's recommendation\end{tabular} & 6 {[}5,6.5{]} & a \\
\begin{tabular}[c]{@{}c@{}}Q4. It is important for me to know\\ how the model makes its recommendation\\ for my individual patient\end{tabular} &
  6 {[}5,6{]} &
  a \\
\begin{tabular}[c]{@{}c@{}}Q5. It is important for me to know\\  how uncertain (in \%) the model is\\  about its recommendation\end{tabular} &
  6 {[}5.5, 7{]} &
  a \\
 &
  p\textless{}0.001 &
  p\textless{}0.001 \\
\multicolumn{1}{l}{} &
  Friedman test &
  posthoc Kruskal-Wallis \\* \bottomrule
\end{longtable}

\newpage

% -------------------------------------------------------------------------------------------------------------------------------
% TABLE WITH A-G QUESTIONS MEDIANS

\begin{table}
\centering
\caption{Responses to questions A-G after deciding on patient cases when supported by recommendation only (CS) and after decisions supported by recommendation and its explanation (CS+Exp). No statistically significant change observed in pairwise comparison (Wilcoxon signed-rank test).}
\label{tab:AG_medians}
\resizebox{\textwidth}{!}{
\begin{tabular}{llcccc} 
\toprule
\multicolumn{1}{c}{\multirow{2}{*}{Index}} & \multicolumn{1}{c}{\multirow{2}{*}{Question}}                                                                                                                                  & \multicolumn{2}{c}{\begin{tabular}[c]{@{}c@{}}Median response \\in Likert scale 1-7 [Q1,Q3]\end{tabular}}                                   & \multicolumn{2}{c}{\begin{tabular}[c]{@{}c@{}}Pairwise\\comparison\end{tabular}}  \\
\multicolumn{1}{c}{}                       & \multicolumn{1}{c}{}                                                                                                                                                           & \begin{tabular}[c]{@{}c@{}}CORONET Score\\(CS)\end{tabular} & \begin{tabular}[c]{@{}c@{}}CORONET Score\\+ Explanation\\(CS+Exp)\end{tabular} & p     & significance*                                                             \\
A                                          & \begin{tabular}[c]{@{}l@{}}I am satisfied with the output information\\that CORONET provides\end{tabular}                                                                      & 5 [4.5,6]                                                   & 5 [5,6]                                                                       & 0.976 & ns                                                                        \\
B                                          & \begin{tabular}[c]{@{}l@{}}CORONET helps me in making\\safe clinical decisions on patient management\end{tabular}                                                              & 5 [3.5,5]                                                   & 5 [4,5]                                                                       & 0.425 & ns                                                                        \\
C                                          & \begin{tabular}[c]{@{}l@{}}When my initial decision was the same\\as CORONET had recommended, I felt reassured\end{tabular}                                                    & 5 [4,6]                                                     & 5 [4,6]                                                                       & 0.746 & ns                                                                        \\
D                                          & \begin{tabular}[c]{@{}l@{}}I understand when and why CORONET\\may provide the wrong recommendation in some cases\end{tabular}                                                  & 5 [3,6]                                                     & 5 [4,5]                                                                       & 0.275 & ns                                                                        \\
E                                          & \begin{tabular}[c]{@{}l@{}}CORONET helps in cases where I am less confident\\in the decision on how to proceed\end{tabular}                                                    & 4 [3,5]                                                     & 5 [4,5.5]                                                                     & 0.056 & ns                                                                        \\
F                                          & \begin{tabular}[c]{@{}l@{}}Even when my initial course of action\\was different to what CORONET recommended,\\I still had full confidence in my original decision\end{tabular} & 5 [5,6]                                                     & 5 [5,6]                                                                       & 0.890 & ns                                                                        \\
G                                          & \begin{tabular}[c]{@{}l@{}}I was surprised when CORONET recommended\\an action different to my own\end{tabular}                                                                & 4 [3,5]                                                     & 4 [3,5]                                                                       & 0.821 & ns                                                                        \\
\bottomrule
\end{tabular}
}
\end{table}

% -------------------------------------------------------------------------------------------------------------------------------
% CORRELATIONS IN CHANGES

\begin{table}
\centering
\caption{Statistically significant correlations between changes in responses between CS and CS+Exp scenarios. Only one after adjustment for multiple comparison (using Sidak correction).}
\label{tab:correlations_changes}
\resizebox{\textwidth}{!}{
\begin{tabular}{llrrrl} 
\toprule
\multicolumn{1}{c}{Change
  in:}    & \multicolumn{1}{c}{Change in:}        & \begin{tabular}[c]{@{}c@{}}r\\(spearman)\end{tabular} & p     & \begin{tabular}[c]{@{}c@{}}p-adjusted\\(Sidak)\end{tabular} & significance  \\
Reassurance                         & Understanding
  wrong recommendation~ & 0.762                                                 & 0.000 & 0.003                                                       & **            \\
Satisfaction with the output        & Help in
  making safe decisions~   & 0.632                                                 & 0.001 & 0.135                                                       & ns            \\
Reassurance                         & Help when HCP is
  less confident      & 0.549                                                 & 0.007 & 0.553                                                       & ns            \\
Satisfaction
  with the output      & Reassurance                           & 0.513                                                 & 0.012 & 0.771                                                       & ns            \\
Help
  in making safe decisions~ & Help when HCP is
  less confident      & 0.418                                                 & 0.047 & 0.997                                                       & ns            \\
\bottomrule
\end{tabular}
}
\end{table}

\clearpage

\subsection{Supplementary Figures}

% LIKERT A-G
\begin{figure}[h!]
\centering
\includegraphics[width= .9\textwidth]{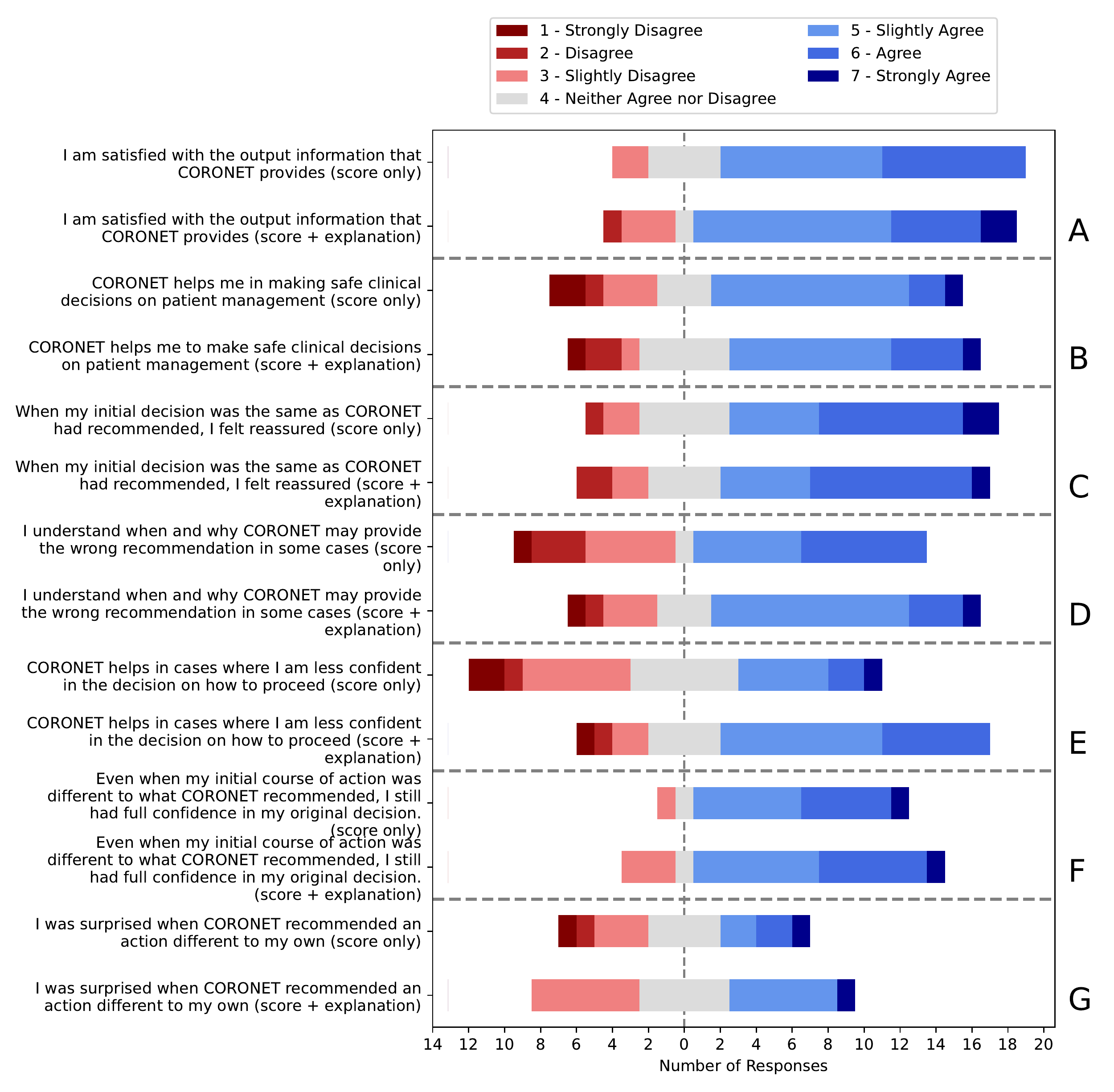}
\caption{Responses to questions A-G which were asked twice: after deciding on patients 1-5, where only the CORONET score (CS) was presented, and after deciding on the patients 6-10, when additional model’s explanation was delivered (CS+Exp).
}
\label{fig:likert_scale_change_in_feedback}
\end{figure}

\begin{figure}[htb!]
\centering
\includegraphics[width= .5\textwidth]{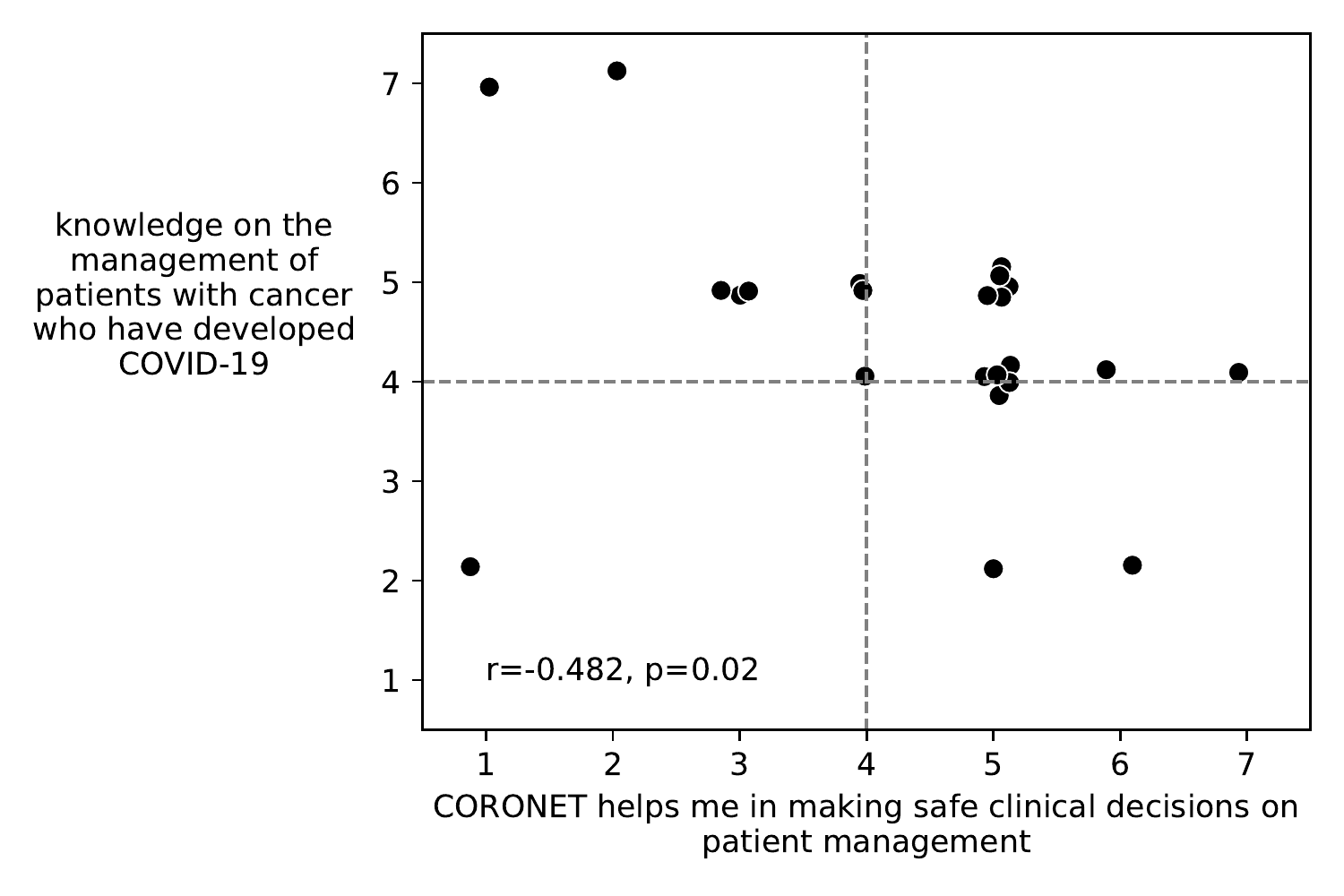}
\caption{Correlation indicating positive feedback to the model’s recommendation, which is not supported by an explanation (CS). The lower the expertise, the more helpful the tool appeared to be, even when no explanation is provided. Answers on X axis are from CS scenario; points are scattered for visibility.}
\label{fig:expertise_vs_feedback_CS}
\end{figure}

\begin{figure}[htbp]
\centering
\begin{subfigure}[t]{.45\textwidth}
  \centering
  \includegraphics[width=\linewidth]{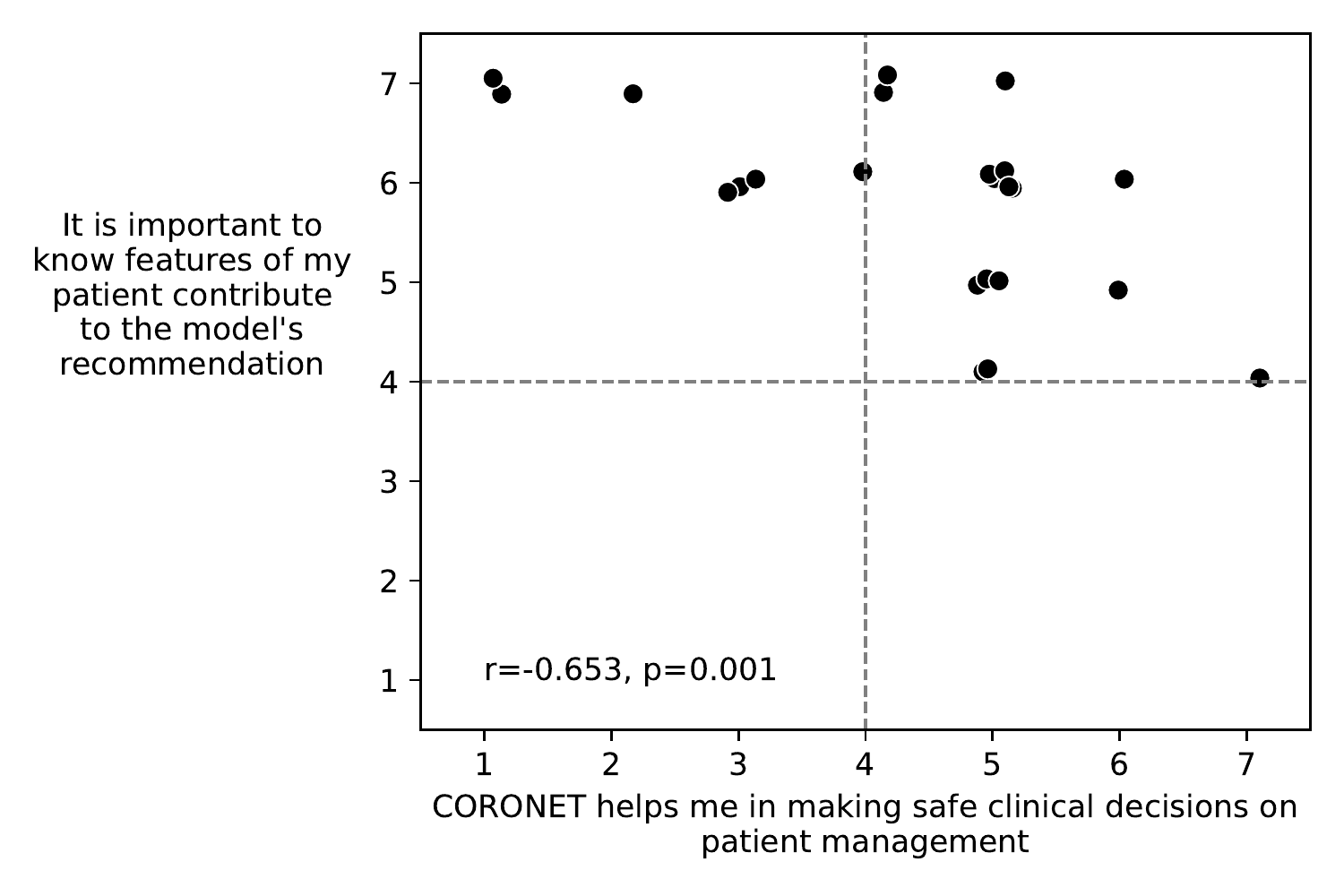}
  \caption{}
  
\end{subfigure}
%\hfill
\begin{subfigure}[t]{.45\textwidth}
  \centering
  \includegraphics[width=\linewidth]{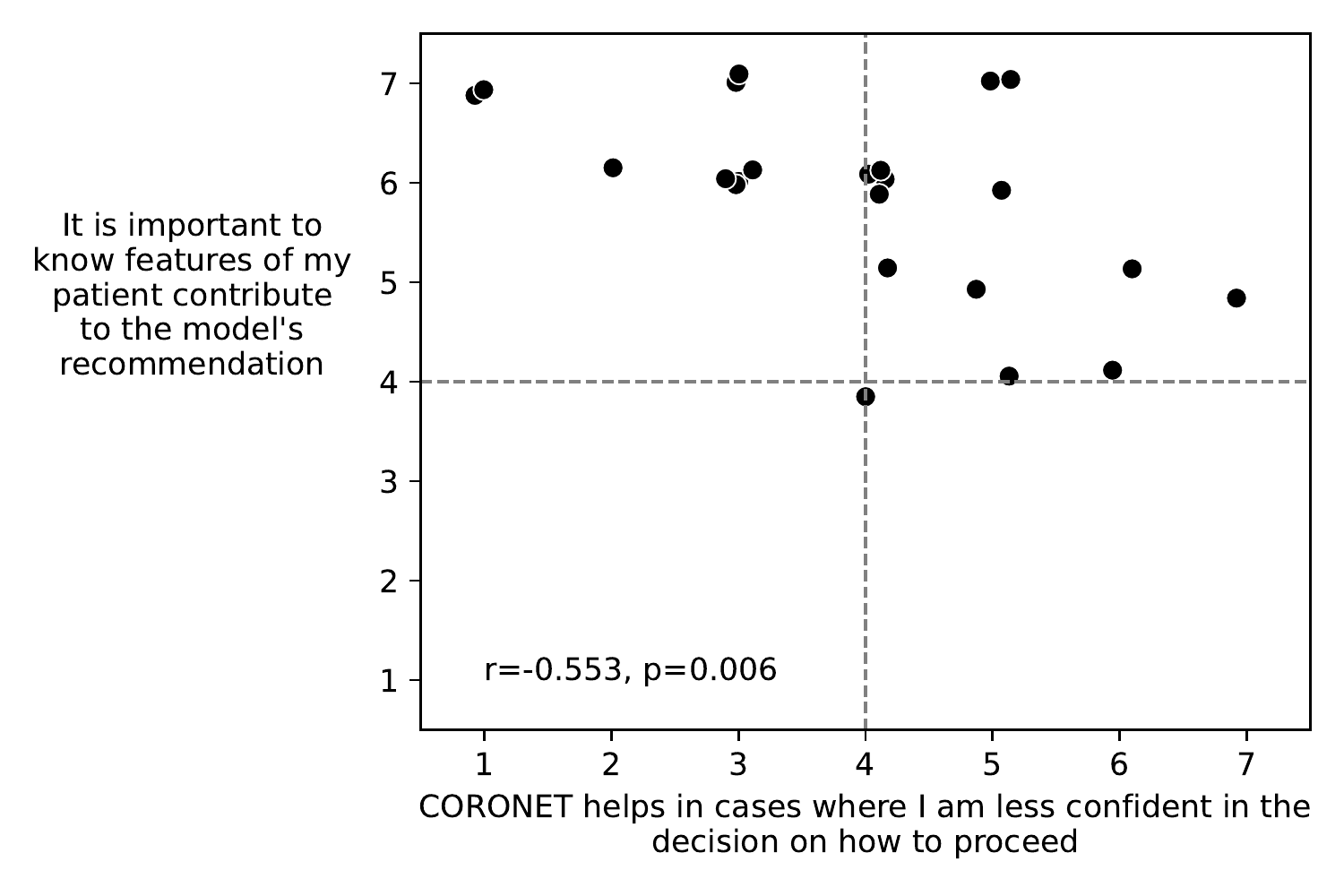}
  \caption{}
\end{subfigure}
%\hfill
\caption{Correlation indicating positive feedback to the model’s recommendation, which is not supported by an explanation (CS). The higher the need for knowing the contributing features, the less helpful CS output is likely to be. Answers on X axis are from CS scenario; points are scattered for visibility.}
\label{fig:knowing_features_vs_helpful_CS}
\end{figure}

%\begin{figure}[htb!]
%\centering
%\includegraphics[width= .99\textwidth]{figures/knowing_features_vs_helpful_CS.PNG}
%\caption{Correlation indicating positive feedback to the model’s recommendation, which is not supported by an explanation (CS). The higher the need for knowing the contributing features, the less helpful CS output is likely to be. Answers on X axis are from CS scenario; points are scattered for visibility.}
%\label{fig:knowing_features_vs_helpful_CS}
%\end{figure}

\begin{figure}[htbp]
\centering
\begin{subfigure}{.45\textwidth}
  \centering
  \includegraphics[width=\linewidth]{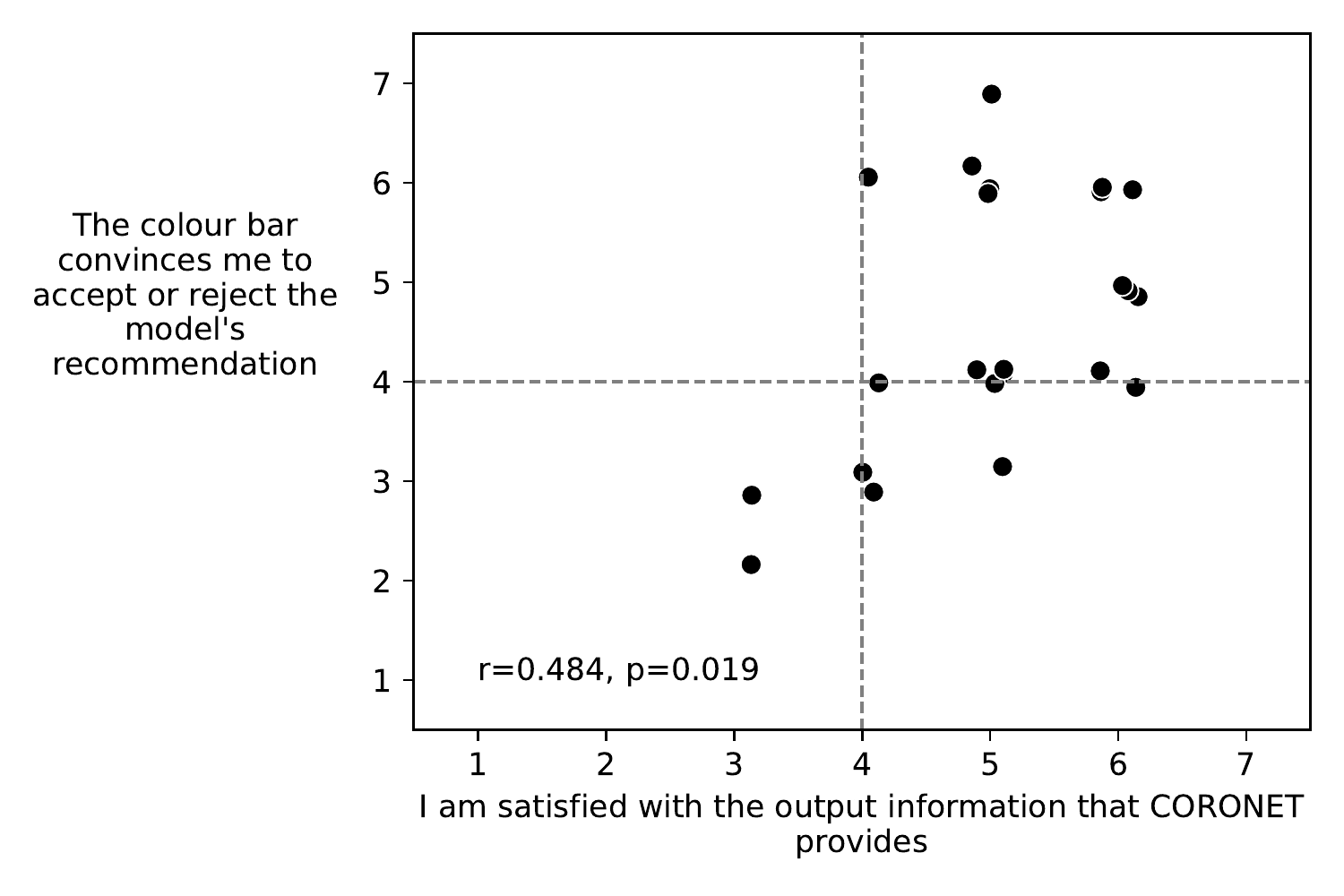}
  \caption{}
\end{subfigure}%
%\hfill
\begin{subfigure}{.45\textwidth}
  \centering
  \includegraphics[width=\linewidth]{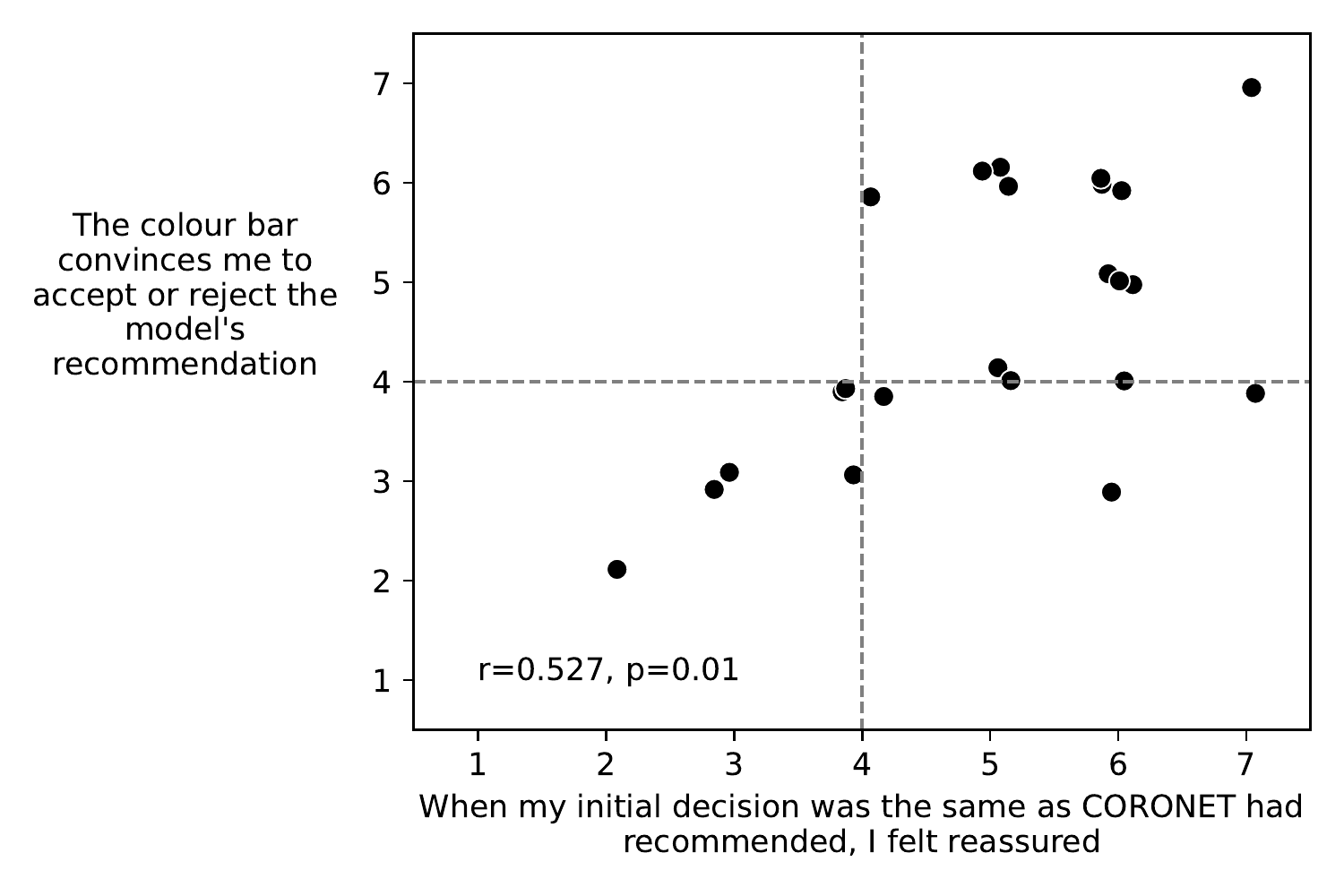}
  \caption{}
\end{subfigure}
%\hfill
\caption{Correlation indicating positive feedback to the model’s recommendation, which is not supported by an explanation (CS). Satisfaction with the CS output, and reassurance when the tool recommended the same action as their own were correlated with being convinced by the colour bar. Answers on X axis are from CS scenario; points are scattered for visibility.}
\label{fig:colorbar_vs_satisfaction}
\end{figure}

%\begin{figure}[htb!]
%\centering
%\includegraphics[width= .99\textwidth]{figures/colorbar_vs_satisfaction.PNG}
%\caption{Correlation indicating positive feedback to the model’s recommendation, which is not supported by an explanation (CS). Satisfaction with the CS output, and reassurance when the tool recommended the same action as their own were correlated with being convinced by the colour bar. Answers on X axis are from CS scenario; points are scattered for visibility.}
%\label{fig:colorbar_vs_satisfaction}
%\end{figure}

\begin{figure}[h!]
\centering
\includegraphics[width= .6\textwidth]{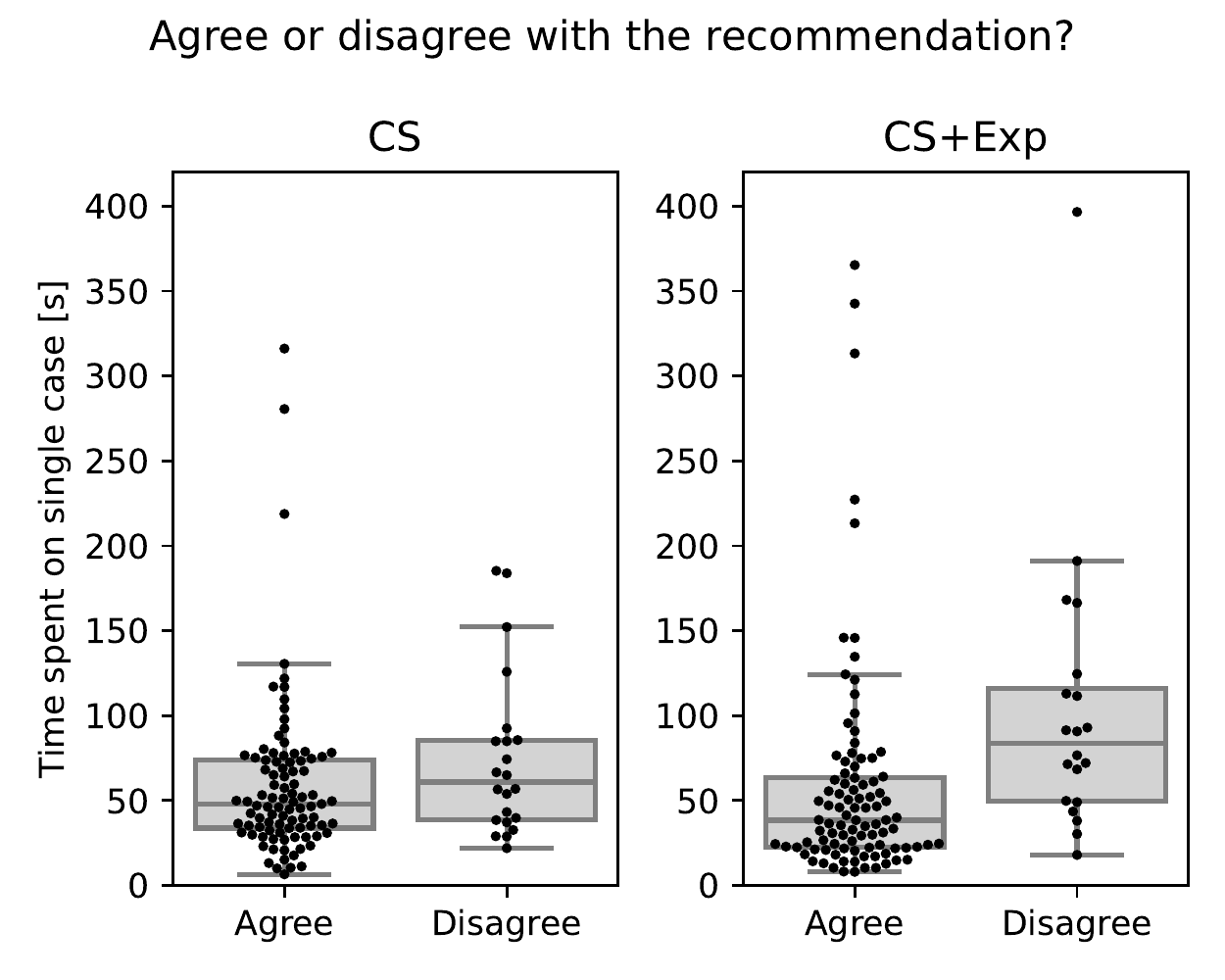}
\caption{Time spent on individual decisions for cases 1-5 (CS) and 6-10 (CS+Exp) stratified by the concordance between the user's final decision and the model’s recommendation.}
\label{fig:time_spent_on_decisions}
\end{figure}

\begin{figure}[h!]
\centering
\includegraphics[width= .4\textwidth]{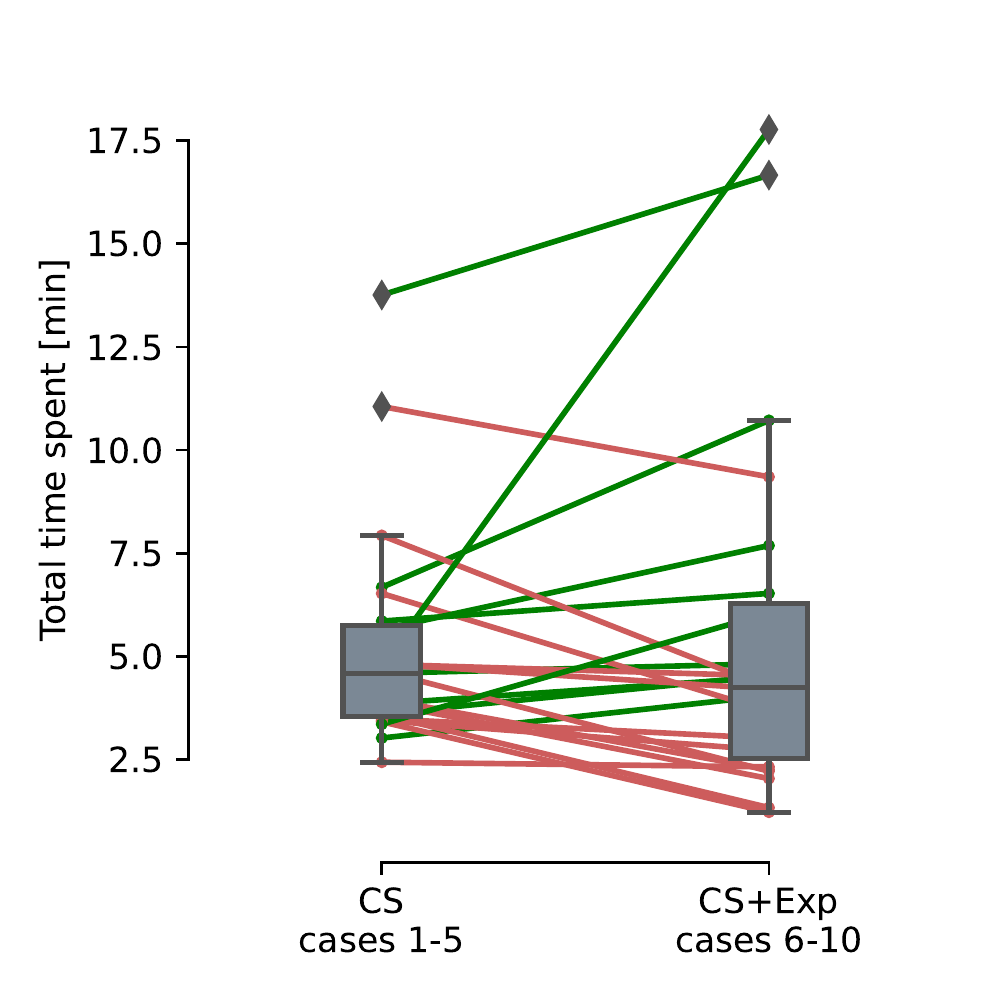}
\caption{Total time spent on making decisions about patients: cases 1-5 where only the score with colour bar were provided; cases 6-10 enhanced with two additional figures with explanation. Each line is an individual clinician. Green - time increased, red - decreased. No significant difference (p=0.846, Wilcoxon sign rank test).}
\label{fig:time_pairedplot}
\end{figure}

%%%%%%%%%%%%%%%%%%%%%%%%%%%%%%%%%%%%%%%%%%%%%%%%%%%%%%%%%%%%%%%%%%%%%%%%%%%%%%%%%%%%%%%%%%%%%%%%%%%%%%%%%%%%%%
% coronet interface figures
\begin{figure}[h!]
\centering
\includegraphics[width= .99\textwidth]{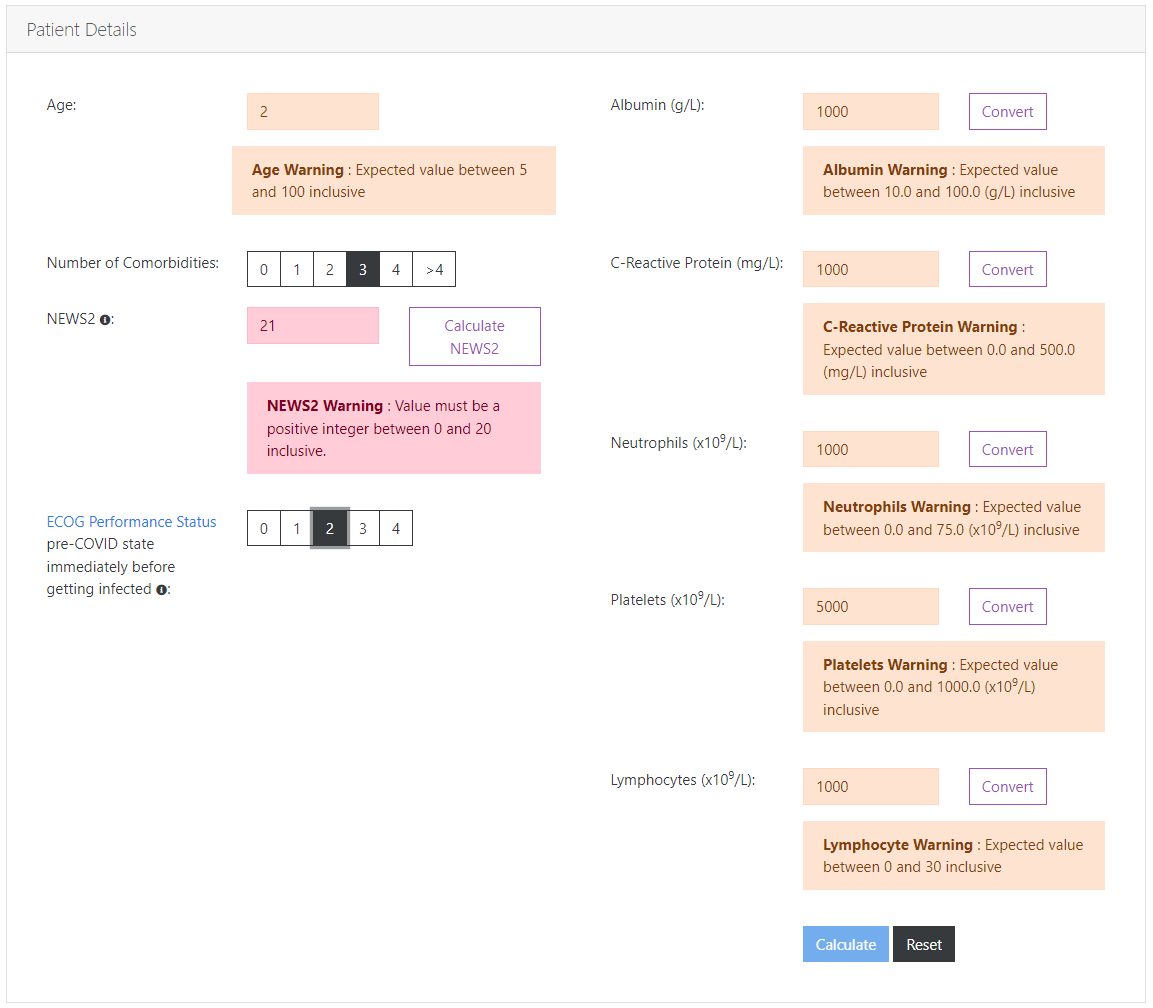}
\caption{Warnings displayed to the user when inputing the values of of expected range.
user interface of the CORONET available at \url{https://coronet.manchester.ac.uk}. }
\label{fig:out_ouf_range_warning}
\end{figure}

\begin{figure}[h!]
\centering
\includegraphics[width= .99\textwidth]{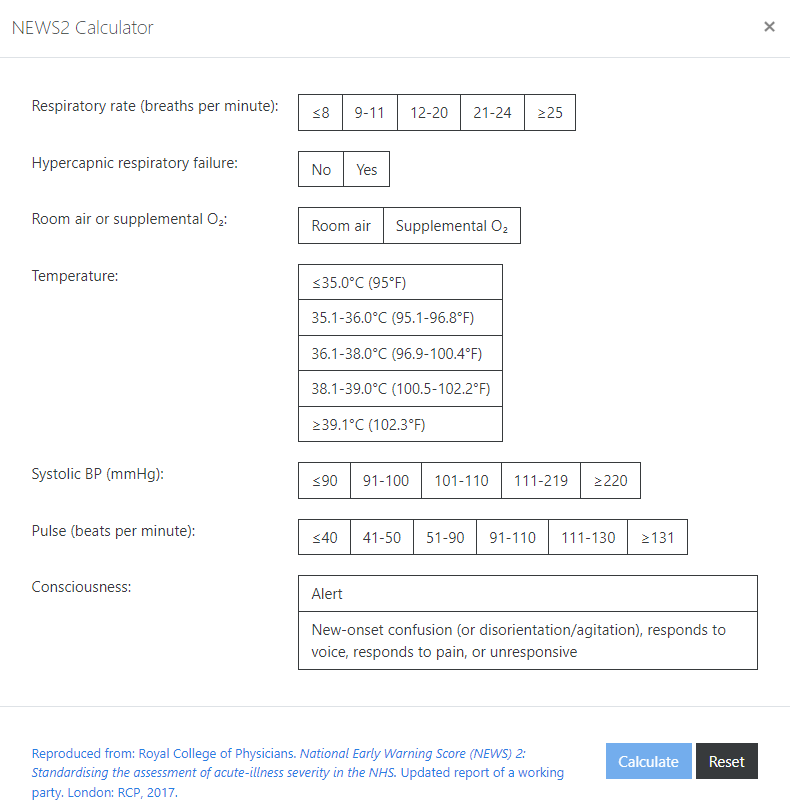}
\caption{A pop-window provided to the user when clicking on `Calculate NEWS2' buttom.
User interface of the CORONET available at \url{https://coronet.manchester.ac.uk}}
\label{fig:NEWS_pop_up}
\end{figure}
%%%%%%%%%%%%%%%%%%%%%%%%%%%%%%%%%%%%%%%%%%%%%%%%%%%%%%%%%%%%%%%%%%%%%%%%%%%%%%%%%%%%%%%%%%%%%%%%%%%%%%%%%%%%%%

\end{document}